\documentclass[final]{cvpr}

\usepackage{times}
\usepackage{epsfig}
\usepackage{graphicx}
\usepackage{amsmath}
\usepackage{amssymb}

\graphicspath{{./graphics/}}
\usepackage{animate}

\usepackage{arydshln}
\usepackage{amsmath}
\usepackage{amssymb}
\usepackage{color}
\usepackage{caption}
\usepackage[font=small]{caption}

\usepackage{tabulary}

\usepackage{microtype}
\usepackage{subfigure}
\usepackage{booktabs} % for professional tables

\usepackage[normalem]{ulem} %to strike the words

\usepackage{adjustbox}
\usepackage{tabularx}
\usepackage{multirow}
\makeatletter
\newcommand{\thickhline}{%
	\noalign {\ifnum 0=`}\fi \hrule height 1.3pt
	\futurelet \reserved@a \@xhline
}

\usepackage[pagebackref=true,breaklinks=true,colorlinks,bookmarks=false]{hyperref}

\def\cvprPaperID{2575} % *** Enter the CVPR Paper ID here
\def\confYear{CVPR 2021}

\begin{document}

\makeatletter
\g@addto@macro\@maketitle{
    \vspace{-30pt}
    \begin{center}\centering
        \setlength{\tabcolsep}{0.06cm}
        \setlength{\columnwidth}{2.81cm}
        \hspace*{-\tabcolsep}\begin{tabular}{cccc}
                \footnotesize CResMD ~\cite{jingwen2020interactive}
            &
                \footnotesize TSNet (Ours)
            &
                \footnotesize TA+TSNet (Ours)
            &
            \vspace{-0.1cm}
           \\
             \animategraphics[width=0.23\linewidth, poster=3, autoplay, palindrome, final, nomouse, method=widget]{1}{graphics/anne/CResMD/}{00}{10}
            &
             \animategraphics[width=0.23\linewidth, poster=3, autoplay, palindrome, final, nomouse, method=widget]{1.5}{graphics/anne/TSNet/}{00}{10}            
             &
             \animategraphics[width=0.23\linewidth, poster=3, autoplay, palindrome, final, nomouse, method=widget]{3}{graphics/anne/TA+TSNet/}{00}{10}     
             &
            \includegraphics[width=0.29\linewidth]{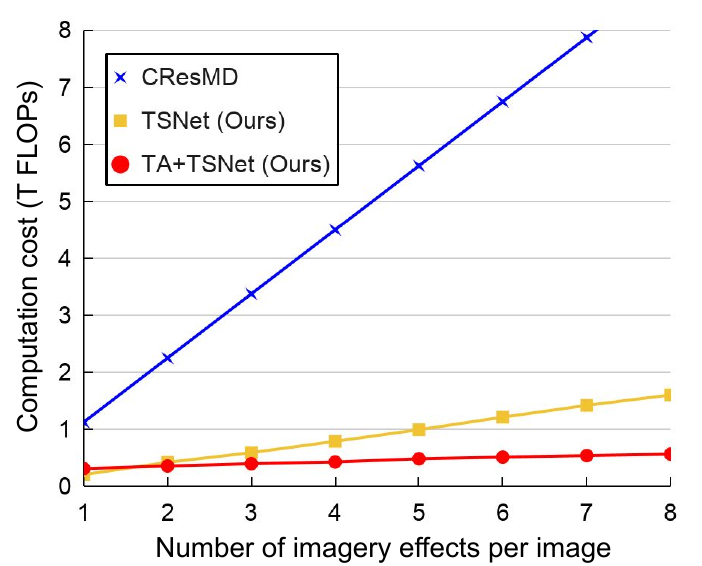} 
            \vspace{-0.1cm} \\
             \animategraphics[width=0.23\linewidth, poster=3, autoplay, palindrome, final, nomouse, method=widget]{1}{graphics/slidebar/CResMD/}{00}{10}
            &
            \animategraphics[width=0.23\linewidth, poster=3, autoplay, palindrome, final, nomouse, method=widget]{1.5}{graphics/slidebar/TSNet/}{00}{10}
            &
            \animategraphics[width=0.23\linewidth, poster=3, autoplay, palindrome, final, nomouse, method=widget]{3}{graphics/slidebar/TA+TSNet/}{00}{10}
        \end{tabular}\vspace{-0.3cm}
        \captionof{figure}{\textbf{A controllable image restoration example with network computation costs}. Our approach generates visually pleasing imagery effects during modulation with the low latency and FLOPs. \textit{This is a video figure that is best viewed using Adobe Reader.}}\vspace{-0.0cm}
        \label{fig:overview}
    \end{center}
    \vspace*{-2pt}
}

\makeatother

%%%%%%%%% TITLE
\title{Searching for Controllable Image Restoration Networks}

\author{
  Heewon Kim \hskip1.6em Sungyong Baik \hskip1.6em Myungsub Choi \hskip1.6em Janghoon Choi \hskip1.6em Kyoung Mu Lee \\
   ASRI, Department of ECE, Seoul National University \\
   {\tt\small \{ghimhw, dsybaik, cms6539, ultio791, kyoungmu\}@snu.ac.kr} 
}

\maketitle
\vspace*{15pt}
%%%%%%%%% ABSTRACT
\begin{abstract}
\vspace*{-20pt}
Diverse user preferences over images have recently led to a great amount of interest in controlling the imagery effects for image restoration tasks.
However, existing methods require separate inference through the entire network per each output, which hinders users from readily comparing multiple imagery effects due to long latency.
To this end, we propose a novel framework based on a neural architecture search technique that enables efficient generation of multiple imagery effects via two stages of pruning: task-agnostic and task-specific pruning.
Specifically, task-specific pruning learns to adaptively remove the irrelevant network parameters for each task, while task-agnostic pruning learns to find an efficient architecture by sharing the early layers of the network across different tasks.
Since the shared layers allow for feature reuse, only a single inference of the task-agnostic layers is needed to generate multiple imagery effects from the input image.
Using the proposed task-agnostic and task-specific pruning schemes together significantly reduces the FLOPs and the actual latency of inference compared to the baseline. 
%Furthermore, our training scheme facilitates the generation of visually pleasing imagery effects in diverse image modulation tasks.  
We reduce 95.7\% of the FLOPs when generating 27 imagery effects, and make the GPU latency 73.0\% faster on 4K-resolution images.
\end{abstract}

% !TEX root = main.tex
\vspace{0.2cm}
\section{Introduction}
\label{sec:introduction}
%{Task vector, architecture controller,}

%%%% Intro to Controllable Image Restoration - Motivation %%%%

Deep convolutional neural networks have achieved significant progress in many image restoration tasks, including image denoising, deblurring, and super-resolution.
Conventional approaches generate a fixed output from an input degraded image, restoring the image details in a deterministic manner.
However, user preferences often vary; thus the desired level of restoration may differ, depending on the content of an image and the user preferences.

%%%% Problems for existing methods %%%%
Photo editing tools alleviate this problem by enabling the modulation of imagery effects\footnote{In this paper, an imagery effect denotes an output image of algorithms with a pre-determined image restoration task.} with a sliding bar.
Recent deep-learning-based controllable image restoration (CIR) approaches~\cite{jingwen2019modulating,jingwen2020interactive,alon2019dynamic,wei2019cfsnet,xintao2019deep} greatly improved the quality of the imagery effects by making feature maps of their models adaptive to the input image and restoration task.
However, these approaches lack the consideration of the computation cost for multiple inference passes despite the  importance of the computational efficiency in practical applications.

%%%% Which aspects to consider when improving the efficiency %%%%
Controlling the imagery effects requires an infinite number of inference tasks for the purpose of continuously varying restoration levels.
Although we can discretize the control levels to enable feasible inference, existing models are still impractical when generating many different imagery effects, as illustrated in Figure~\ref{fig:overview} (we compare our models to the recent state-of-the-art model CResMD~\cite{jingwen2020interactive}).
To achieve seamless control over various imagery effects, the computation cost should be greatly reduced, especially when a large number of network inference is required to meet user preferences.

%%%% Our proposed method %%%%
In this paper, we present a novel controllable image restoration framework that utilizes neural architecture search to significantly improve the efficiency of models.
Specifically, we first introduce an architecture controller, which adaptively determines an efficient task-specific network (TSNet) architecture by pruning the redundant channels with respect to each restoration task.
Then, we search for a task-agnostic architecture that aims to effectively share the common low-level features.
The combination of task-agnostic and task-specific network (TA+TSNet) greatly removes the computation of redundant feature maps, which makes it especially more efficient for generating multiple imagery effects that may require a large number of inference pass.
The main concept of our search strategy is summarized in Figure~\ref{fig:search_strategy}.

Our search algorithm directly optimizes the trade-off between the restoration accuracy and FLOPs in multiple output generation.
When generating 27 imagery effects with different modulations, TSNet alone can reduce 79.1\% of the FLOPs, and TA+TSNet pushes it further to 95.7\%.
Moreover, our models also show faster actual latency, where TSNet runs 1.7 times faster and TA+TSNet runs 3.7 times faster than the state-of-the-art network when generating 4K resolution images on modern high-end GPUs.
We perform extensive experiments to analyze the latency of our models over different computing devices and image resolutions.

Overall, our contributions can be summarized as follows:
\begin{itemize}
\vspace{-0.1cm}
\item We present a novel neural architecture search strategy to remarkably improve the efficiency of controllable image restoration networks.
\vspace{-0.1cm}
\item The proposed models can effectively handle the generation of multiple imagery effects, and we analyze the accuracy-resource trade-off for the CIR problem.
\vspace{-0.1cm}
\item The experimental results show better image modulation performance with less computation costs in terms of FLOPs and CPU/GPU latency.
\end{itemize}

% !TEX root = main.tex
\section{Related Work}
\label{sec:related_work}

\subsection{Image restoration}
%\vspace{-0.2cm}

Image restoration, including denoising, deblurring, super-resolution, and compression artifact removal, is a classical topic in computer vision which aims at restoring the original image from its degraded versions.
Deep-learning-based image restoration networks~\cite{zhang2017beyond, dong2014image,kim2016accurate,ledig2016photo,bee2017enhanced,zhang2018residual,dong2015compression,dong2016accelerating} have achieved breakthroughs in restoring accurate image details.
While the conventional approaches specialize in dealing with a single degradation type, general image restoration aims to restore the corrupted image with multiple types of degradation.
Notable approaches include learning a policy to determine a specialized restoration network for the input image~\cite{yu2018crafting,yu2019path}, or using an operation-wise attention module to produce the specialized feature maps \textit{w.r.t.} each degradation type~\cite{suganuma2019attention}.
However, these approaches cannot control the diverse levels of restoration for the input images (which is already supported in many image editing tools), and sometimes generate overly smooth or sharpened outputs.

On the other hand, controllable image restoration is recently gaining interest from the computer vision research community, which aims to generate diverse imagery effects from an image for unseen degradation and user preferences.
Existing works learn to generate the intermediate imagery effects between two objectives~\cite{jingwen2019modulating,wei2019cfsnet,alon2019dynamic,xintao2019deep}.
AdaFM~\cite{jingwen2019modulating}, CFSNet~\cite{wei2019cfsnet}, and Dynamic-Net~\cite{alon2019dynamic} design their network architectures with tuning modules, which modulate the learned feature maps layer-wise~\cite{jingwen2019modulating} or block-wise~\cite{wei2019cfsnet,alon2019dynamic} with respect to the tasks of interest at test time.
Instead, DNI~\cite{xintao2019deep} interpolates all parameters of the differently trained networks for modulation.
For the general controllable image restoration task, CResMD~\cite{jingwen2020interactive} controls the intermediate images in restoring the multiple degradation types and levels with a block-wise tuning module.
While the prior works may have provided good performance and control over imagery effects, they have solely focused on the performance and do not consider computational efficiency.
In contrast, using CResMD as the baseline, the proposed task-specific and task-agnostic channel pruning schemes significantly reduce the computations and running time by a large margin.

\begin{figure*}[t]
	\centering
	%\framebox(200,100){}
	\includegraphics[width=1\linewidth, trim = 0cm 0cm 0cm 0cm]{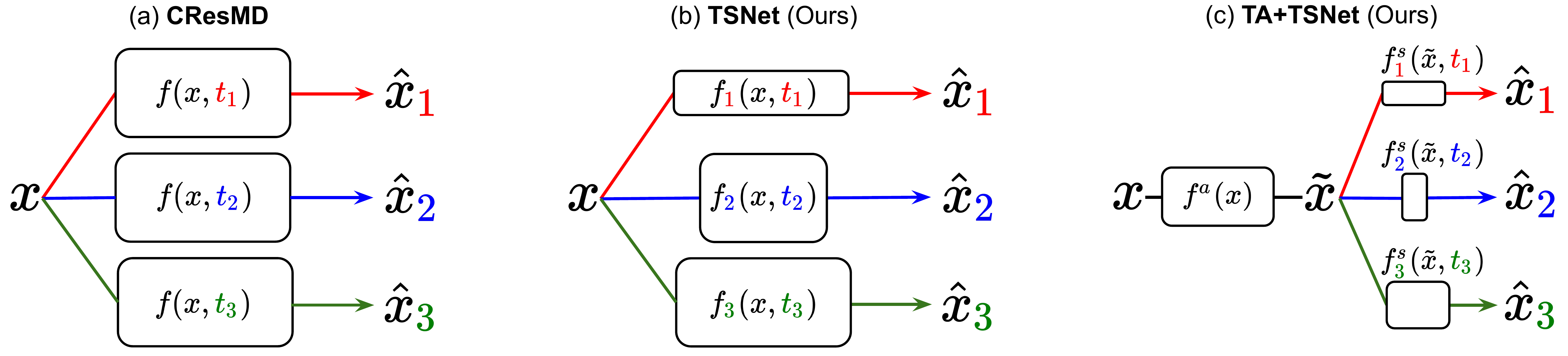} \\ \vspace{-0.2cm}
	\caption{The overview of our architecture search strategy for controllable image restoration. 
		The width and height of boxes represent the number of layers and channels of neural networks, respectively.
		Three tasks ($\boldsymbol{t}_1$, $\boldsymbol{t}_2$, and $\boldsymbol{t}_3$) are visualized in this example, where each task is illustrated with a different color. 
		(a) CResMD~\cite{jingwen2020interactive} modulates feature maps to produce multiple imagery effects $\hat{\boldsymbol{x}}_{m}$ from an input image $\boldsymbol{x}$ with respect to the task vector $\boldsymbol{t}_m$. The network $f$ in this approach has to compute all network parameters per task.
		(b) Task-specific network architecture (TSNet, denoted as $f_m$) adaptively removes channels and parameters that are irrelevant to a given task.
		(c) Task-agnostic and task-specific network architecture (TA+TSNet, denoted as $f_a$) shares the feature map of early layers to facilitate feature reuse and thus requires a single computation of the feature maps of early layers to generate multiple imagery effects.
	}
	\vspace{-0.2cm}
	\label{fig:search_strategy}
\end{figure*}

\subsection{Efficient CNNs for image restoration}
\vspace{-0.2cm}

To make the image restoration models efficient with less computation cost, many novel architectures have been developed for diverse restoration tasks.
The early works downscale the spatial resolution of the input image to reduce the computation costs of the convolution operations for denoising~\cite{zhang2018ffdnet} and super-resolution~\cite{dong2016accelerating}.
More recently, CARN~\cite{ahn2018fast} presents a cascading residual block with multiple skip connections, leading to a fast and light-weight super-resolution network.
For deblurring, a method using spatially variant deconvolution is proposed in \cite{yuan2020efficient} to achieve accurate performance with its efficient backbone network.
Meanwhile, FALSR~\cite{Chu2019FastAA}, ESRN~\cite{Song2019EfficientRD}, and FGNAS~\cite{kim2019fine} adopt neural architecture search (NAS) algorithms for efficient super-resolution networks.
Path-Restore~\cite{yu2019path} and AdaDSR~\cite{liu2020deep} save computation costs via adaptive inference for general image restoration and super-resolution, respectively.
Prior works also employ network quantization~\cite{jingwei2020binarized} or pruning~\cite{Hou2020efficient}, but these approaches are not adaptive to task variations.
On the other hand, we study the network acceleration approaches for controllable image restoration for the first time, especially when it requires a large number of inference per image.
The proposed method incorporates neural architecture search via channel pruning to \textit{adaptively} accelerate a general controllable image restoration network that handles multiple tasks.

% !TEX root = main.tex

\section{Method}
\label{sec:method}

%%%%%%%%%%%%%%%%%%%%%%%%%%
\begin{figure*}[t]
	\centering
	\includegraphics[width=1\linewidth]{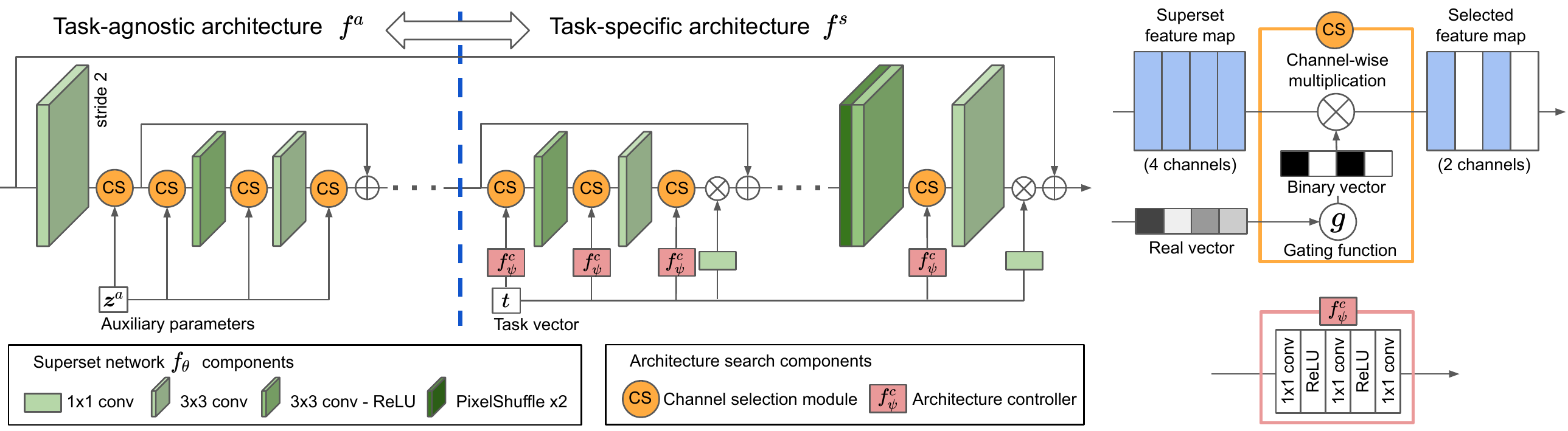} \\ \vspace{-0.1cm}
	\caption{Our search algorithm selects layers and channels from the super network $f_\theta$ (CResMD) as task-agnostic part $f^a$ and task-specific part $f^s$. The architecture controller $f^c_\psi(\boldsymbol{t}_m)$ determines the channels of $f^s$ by predicting $\boldsymbol{z}^s_m$ and updates $\boldsymbol{z}^a$ which learn task preferences on channels.
	When most tasks agree where to prune channels at the first task-specific layer, the layer becomes task-agnostic.
	The example of channel selection module describes selecting two channels from the four-channel feature map. Real vector denotes a channel dimension vector of $\boldsymbol{z}^a_{n}$ or $\boldsymbol{z}^s_{m,n}$. Gating function binarizes each element of the vector in forward pass. Note that the convolutions of the super network do not calculate the ``deactivate'' channels in the test time by selecting convolution weights and channels indexed by the binary vector.
	}
	\vspace{-1em}
	\label{fig:architecture}
\end{figure*}
%%%%%%%%%%%%%%%%%%%%%%%%%%

\subsection{Problem formulation}
\label{sec:problem}

\paragraph{Imagery effect with task vector.}
\textit{Controllable} image restoration (or image modulation) aims to produce imagery effects by restoring an image with different magnitudes of degradation for each degradation type.
Formally, given $D$ number of degradation types (\textit{e.g.} blur, noise, etc.), a task vector $\boldsymbol{t}_m \in \mathbb{R}^{D}$ encodes the $m$-th image restoration task (and thus the $m$-th imagery effect where $m \in \{1,2,...,M\}$), in which each $d$-th element of $\boldsymbol{t}_m$ ($t_{m,d} \in [0,1]$) determines the restoration level to the corresponding $d$-th degradation type.
During the training of deep neural networks, the task vector $\boldsymbol{t}_m$ are randomly sampled with its corresponding training pairs of an input image and a target image.
At inference time, the task vector is a control variable adjusted by users to determine an imagery effect.
We assume that the optimal task vector which generates the best imagery effect with respect to a pre-determined measure (\textit{e.g.} PSNR, LPIPS, user preferences, etc) is unknown at real-world degraded images.
Thus, to find such task vector, the controllable image restoration networks have to generate a large number of imagery effects per input image.
For convenience, we use $M$ to denote the arbitrary number of imagery effects generated for the given task until user preferences or needs are met.

\paragraph{Computation cost for imagery effects.}
Our goal is to design the accurate network architectures to generate multiple imagery effects per input image, while minimizing the computational costs of the entire process.
Figure~\ref{fig:search_strategy} summarizes our architecture search strategy, including comparisons to one of the representative previous works, CResMD~\cite{jingwen2020interactive}.
Architectures from CResMD and other previous works~\cite{jingwen2019modulating,wei2019cfsnet,alon2019dynamic,xintao2019deep} remain fixed and require the entire network inference per imagery effect, as outlined in Figure~\ref{fig:search_strategy}\textcolor{red}{(a)}.
Their average computation costs for generating $M$ imagery effects formally given by,
\begin{equation}\label{eq:cost_metric}
\mathcal{R}_\text{total}(f,\boldsymbol{x}) = {1\over M} \cdot \sum_{m=1}^M\mathcal{R}(f,\boldsymbol{x},\boldsymbol{t}_m),
\end{equation}
where $\mathcal{R}(f,\boldsymbol{x},\boldsymbol{t}_m)$ denotes FLOPs or latency to generate the $m$-th imagery effect with the network architecture $f$, the input image $\boldsymbol{x}$, and the task vector $\boldsymbol{t}_m$.
TSNet from our first strategy, which corresponds to Figure~\ref{fig:search_strategy}\textcolor{red}{(b)}, searches for an efficient network architecture specifically for each imagery effect.
This leads the average computation costs to
\begin{equation}
\label{eq:tsnet}
    \mathcal{R}_\text{total}^\text{TS}(f,\boldsymbol{x}) = \frac{1}{M} \cdot \sum_{m=1}^M \left[\mathcal{R}(f_m, \boldsymbol{x},\boldsymbol{t}_m) + \epsilon_m\right], 
\end{equation}
where a fixed architecture $f$ is replaced with a $m$-th-imagery-effect-specific efficient network $f_m$ with the auxiliary computation cost $\epsilon_m$ needed for the task-specific pruning process.
Then, TA+TSNet from our second strategy, described in Figure~\textcolor{red}{2(c)}, introduce a task-agnostic architecture $f^a$ which shares the feature map of early layers across tasks to facilitate feature reuse.
Formally, 
\begin{equation}
\label{eq:decomposed}
\begin{split}
    f_m(\boldsymbol{x},\boldsymbol{t}_m) &\approx f^s_m(f^a(\boldsymbol{x}),\boldsymbol{t}_m)\\
    & =f^s_m(\Tilde{\boldsymbol{x}},\boldsymbol{t}_m),
\end{split}
\end{equation}
where $f^s_m$ is the remaining task-specific layers of $f_m$ after $f^a$, and $\tilde{\boldsymbol{x}}$ is the feature map output of $f^a(\boldsymbol{x})$.
Consequently, only a single computation of $\tilde{\boldsymbol{x}}$ is required for all $M$ imagery effects, removing redundant $M-1$ number of computations of the feature maps of the shared early layers.
\begin{equation}\label{eq:cost_metric_feature_sharing}
\mathcal{R}_\text{total}^\text{TA+TS}(f,\boldsymbol{x}) 
= {1 \over M} \cdot \mathcal{R}(f^a,\boldsymbol{x}) + \mathcal{R}_\text{total}^\text{TS}(f^s,\Tilde{\boldsymbol{x}}) , 
\end{equation}
where $\mathcal{R}(f^a, \boldsymbol{x})$ is the computation costs of \textit{a single computation} for $f^a$.
As TSNet is a special instance of TA+TSNet that lacks a shared layer, the following section describes our neural architecture search algorithm for TA+TSNet.

%\vspace{-0.2cm}
\subsection{Searching for controllable architectures}
\paragraph{Search space.} 
Our search algorithm is a variant of super-network-based NAS approaches~\cite{liu2018darts,kim2019fine}, which aim to find an efficient or performance-wise optimal network from a large network called super network.
The search process is performed over a search space of operations or components, each combination of which provides a candidate network derived from a super network.
In this work, our search algorithm determines whether a layer should be shared across tasks and whether channels should be pruned from the super network CResMD~\cite{jingwen2020interactive} with the architecture controller in an end-to-end manner.

\vspace{-0.3cm}
\paragraph{Search strategy.}
As summarized in Figure~\ref{fig:architecture}, our search algorithm finds an efficient network by determining whether each channel is important for a given task, all tasks, or none.
For finding a task-specific architecture $f^s$, a channel is retained (removed) if it is deemed to be important (irrelevant) for a given task.
Similarly, for a task-agnostic architecture $f^a$, a channel is retained (removed) if it is deemed to be important (irrelevant) for most tasks.
The channel importance for tasks (or task preference on channels) is determined by $\boldsymbol{z}^a \in \mathbb{R}^{N \times C}$ and $\boldsymbol{z}^s_m \in \mathbb{R}^{N \times C}$, where $m$, $N$, and $C$ denote the task index, the channel selection module index, and the channel index, respectively.

\vspace{-0.3cm}
\paragraph{Channel selection module.}
To \emph{activate} or \emph{de-activate} each channel in the super network, channel selection (CS) module virtually multiplies a binary value in forward-pass from the differentiable gating function formally given by,
\begin{equation}\label{eq:gating_function}
g(z^*) = 
\begin{cases}
\mathbb{I}\left[z^*>0\right]& \mbox {if forward} \\
\text{sigmoid}(z^*) & \mbox{if backward},
\end{cases}
\end{equation}
where $*\in \{a,s\}$, $z^*$ denotes an element of $\boldsymbol{z}^*_{m}$, and $\mathbb{I}\left[\cdot\right]$ is an indicator function that returns 1 when its input is true (and 0 otherwise).
Thus, each parameter of $\boldsymbol{z}^s_m$ and $\boldsymbol{z}^a$ determines its corresponding channel in the super network to be \emph{active} or \emph{de-active} for $f^s$ and $f^a$, respectively.
Figure~\ref{fig:architecture} illustrates the locations of CS module, which are the input and output of convolution layers.

\vspace{-0.3cm}
\paragraph{Architecture controller.}
To adaptively modify network architectures of $f^s$, we introduce architecture controller $f^c(\cdot)$ consisting of fully connected layers with activation functions formally given by,
\begin{equation}\label{eq:architecture_controller}
\boldsymbol{z}^s_{m,n} \equiv f^c_n(\boldsymbol{t}_m),
\end{equation}
where $f^c_n$ denotes the architecture controller at the $n$-th CS.
$\boldsymbol{z}^s_m$ denotes the task preference on channels and is the function of $\boldsymbol{t}_m$ as each task vector adaptively activates channels in the super network.

\vspace{-0.3cm}
\paragraph{Searching for task-agnostic layer and channels.}
To find task-agnostic layers, $\boldsymbol{z}^a$ first gathers the preference on each channel by gathering the preference on each channel $z^s_{m,n,c}$ from tasks throughout training as follows:
\begin{equation}\label{eq:z_a}
z^a_{n,c} \leftarrow (1-\alpha) \cdot z^a_{n,c}+\alpha \cdot {1 \over M} \cdot \sum_{m=1}^M z^s_{m,n,c},
\end{equation}
where $\boldsymbol{z}^a$ is initialized with zero values, $c$ denotes the channel index at the $n$-th channel selection module, and $\alpha$ is a hyperparameter for exponential moving average.
$\boldsymbol{z}^a$ is then used to estimate the consensus of tasks in a mini-batch of size $M$ about the preference on each channel by computing an agreement criterion as follows:
\begin{equation}\label{eq:shared_layer_threshold}
{1 \over M} \cdot \sum_{m=1}^M \sum_{c=1}^C  g(z^s_{m,n,c}) \cdot g(z^a_{n,c}) >\gamma \cdot \sum_{c=1}^C g(z^a_{n,c}),
\end{equation}
where $\gamma$ is a threshold hyperparameter.
Whether Equation~\eqref{eq:shared_layer_threshold} holds is represented by a boolean variable $\eta$.
If the equation holds ($\eta=1$), most tasks agree where to prune channels whose layer is thus likely to be shared.
However, the condition represented by the equation may or may not hold depending on which task is in the current training mini-batch.
Thus, $\eta$ is accumulated via $s_n$ throughout training to obtain the consensus of tasks from the entire dataset, similar to Equation~\eqref{eq:z_a}, as follows:
\begin{equation}\label{eq:shared_layer_threshold_accummulate}
s_n \leftarrow (1-\alpha) \cdot s_n + \alpha \cdot \mathbb{I}\left[\eta\right],
\end{equation}
where $s_n$ is initialized with 0.
The higher $s_n$ is, the more tasks agree on the preference on a $n$-th channel, the more the proposed strategy prefers the $n$-th channel selection module to be task-agnostic.
To facilitate feature reuse across tasks, the task-agnostic layers need to be located together at the initial stage of the network.
To this end, the $n$-th channel selection module (CS) is task-agnostic if the $n$-th CS and all previous CSs have $s_i$ greater than a threshold, $\gamma$, formally given by,
\begin{equation}\label{eq:general}
\phi_n = 
\begin{cases}
1 & \mbox {if $s_i > \gamma, \forall i = 1,2,\sim,n$} \\
0 & \mbox{otherwise},
\end{cases}
\end{equation}
where $\boldsymbol{\phi}\in \mathbb{Z}^N_2$ denotes a decision variable, in which the $n$-th element $\phi_n$ is 1 if the $n$-th CS is task-agnostic.

\vspace{-0.3cm}
\paragraph{Objective function.}
To search for efficient architectures, our proposed algorithm introduces regularization terms.
Let $\mathcal{L}(\cdot,\cdot)$ denote a standard $\ell_1$ loss function for image restoration tasks.
The resource regularizer $\mathcal{R}_1(\cdot)$ computes the amount of resources for currently searched architectures by Equation~\eqref{eq:cost_metric_feature_sharing}.
We introduce a regularizer $\mathcal{R}_2$ to maximize the number of task-agnostic layers for more efficient generation of multiple imagery effects.
The overall objective function is formally given by
\begin{equation}\label{eq:objective}
\min_{\theta,\psi} \mathcal{L}(\theta,\psi) + \lambda_1\cdot\mathcal{R}_1(\psi) + \lambda_2\cdot\mathcal{R}_2(\psi), 
\end{equation}   
where $\theta$ and $\psi$ are learnable parameters in the super network $f$ and the architecture controller $f^c$, respectively, and $\lambda_1$ and $\lambda_2$ are hyperparameters for balancing them.
To make a network as task-agnostic as possible without sacrificing the performance, $\mathcal{R}_2$ penalizes the disagreement among tasks on the channel importance as follows:
\begin{equation}\label{eq:loss}
\mathcal{R}_2(\psi) = \sum_{n=1}^N \phi_{n-1} \cdot \sum_{c=1}^{C}\sum_{m=1}^{M}\left\lVert g(z^s_{m,n,c})-g(z^a_{n,c}) \right\rVert_1,
\end{equation}
where layer at $n = 0$ denotes an input image and $\phi_0 \equiv 1$ since the input image is shared for multiple imagery effects for the given task.

%%%%%%%%%%%%%%%%%%%%%%%%%%%%%%%%%%%%%%%%%%%%%%%%%%%%%%%%%%

\begin{figure}[t]
	\centering
	%\framebox(200,100){}

	\includegraphics[width=1\linewidth]{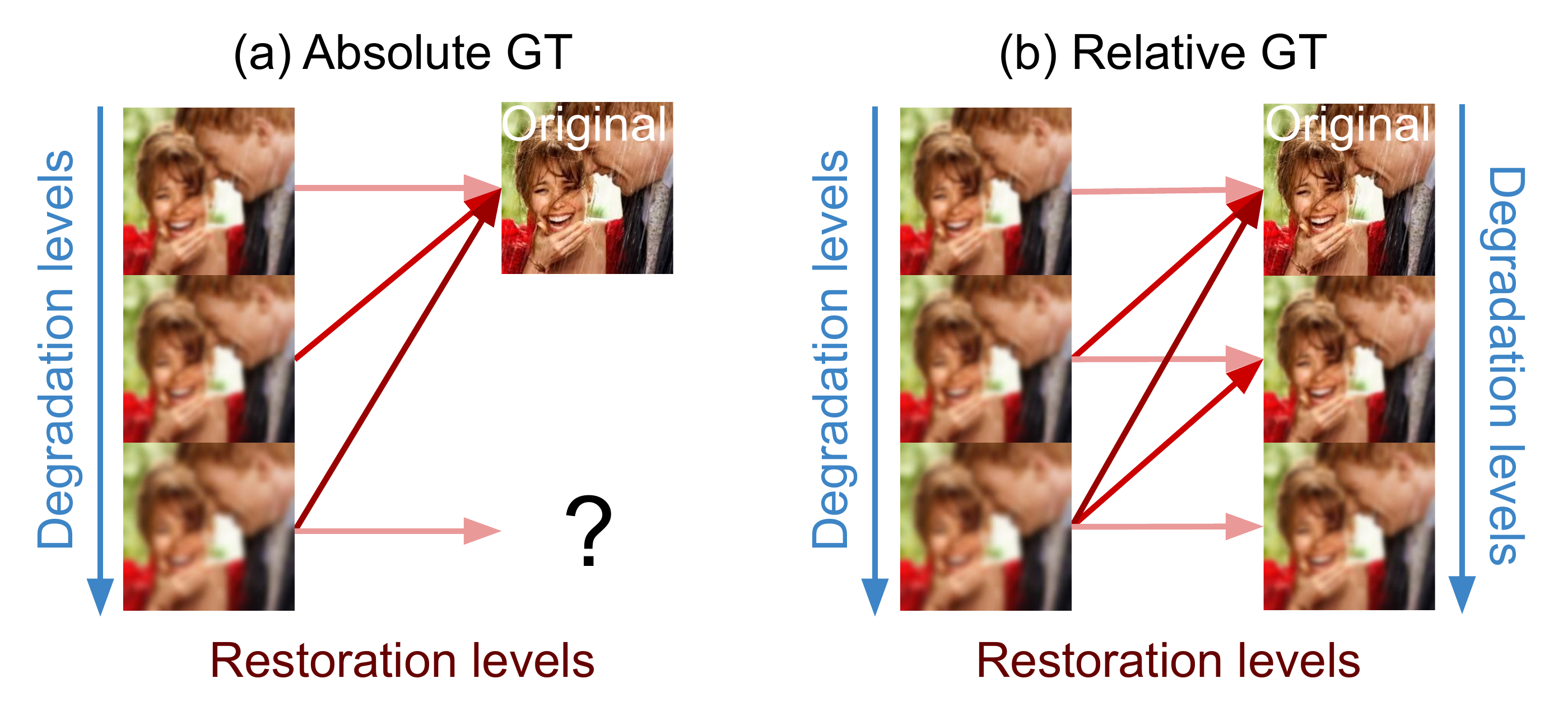}
	\vspace{-0.8cm}
	\caption{Absolute ground truth \textit{vs.} relative ground truth. (a) Mapping from all degraded versions to its original image. (b) Mapping from degraded versions to relatively higher-quality images.
	}
	\vspace{-0.5cm}
	\label{fig:relative_GT}
\end{figure}

%%%%%%%%%%%%%%%%%%%%%%%%%%%%%%%%%%%%%%%%%%%%%%%%%%%%%%%%%%

\subsection{Data sampling strategy}
\paragraph{Degradation level \textit{vs.} restoration level.}
Previous works define an image restoration task as restoring the degraded images with various degradation levels to its original image (see Figure~\ref{fig:relative_GT}\textcolor{red}{(a)}). 
However, CIR aims to produce diverse visually pleasing imagery effects, which may not be possible with the previous training scheme that only focuses on restoring a single original image.
In this work, we redefine a restoration task as restoring the degradation to some extent (see Figure~\ref{fig:relative_GT}\textcolor{red}{(b)}).
The degree of restoration is given as a restoration level, which describes a difference in degradation levels between an input and a ground truth.

\vspace{-0.3cm}
\paragraph{Task vector with relative GT.}
Our task vector $\boldsymbol{t}_{m}$ in training time samples the relative ground truth with the corresponding degrade image, formally given by, 
\begin{equation}\label{eq:relative_task_vector}
t_{m,d} \equiv l^{in}_d - l^{gt}_d,
\end{equation}
where $\boldsymbol{l}^{in}$, $\boldsymbol{l}^{gt} \in \mathbb{R}^D$ denote the degradation levels of the input and ground truth images, respectively, $l^{in}_d, l^{gt}_d \in [0,1]$ for $d$-th degradation type.
The larger degradation level represents the scenario $l^{in}_d \geq l^{gt}_d$, where the ground truth images are less degraded than the input.
For each mini-batch, the training image pairs are equally sampled from 1) uniform distribution for a single degradation type, 2) binary distribution for all types, and 3) uniform distribution for all types.

% !TEX root = main.tex

\begin{table*}[t]
\captionsetup{justification=centering}
\caption{Computation costs per imagery effect. 
TA+TSNet outperforms CResMD~\cite{jingwen2020interactive} and TSNet in all measures and resolutions. 
}
\centering
\vspace{-0.2cm}\hspace{-0.15cm}
\scalebox{1}{ 
\setlength{\tabcolsep}{3.5pt} % Default value: 6pt
\begin{tabular}{lccccccccccc}
	\toprule[1.5pt]
	    & \multicolumn{3}{c}{HD Image resolution} & & \multicolumn{3}{c}{2K Image resolution} & & \multicolumn{3}{c}{4K Image resolution}\\      \cline{2-4} \cline{6-8} \cline{10-12}\\[-10pt]
	    Computation cost & CResMD & TSNet & TA+TSNet && CResMD & TSNet & TA+TSNet&& CResMD & TSNet & TA+TSNet\\
     \midrule[1.0pt]
	    FLOPs (G)        &  1,124.3 &	274.4  &	\textbf{45.2}  &&	2,698.4  &	658.7  &	\textbf{108.4}  &&	10,119.2  &	2470.0  &	\textbf{406.7}   \\ \cdashline{1-12} \\[-10pt]
	    Latency-CPU-S (s)    &  22.8  &	13.5  &	\textbf{5.5}  &&	55.6  &	34.4  &	\textbf{13.5}  && 209.3  &	138.7 &	\textbf{55.5}   \\
	    Latency-CPU-M (s)  &  5.1 &	3.5 &	\textbf{1.7} &&	11.7 &	9.3 &	\textbf{4.2} &&	40.6 &	34.1 &	\textbf{13.1} \\ \cdashline{1-12} \\[-10pt]
	    Latency-GPU (ms)     & 144.4  &	127.2 &	\textbf{68.4} &&	280.8 &	203.1 &	\textbf{99.2} &&	930.0 &	563.4 &	\textbf{250.7} \\
	\toprule[1.5pt]
\end{tabular}}
\vspace{-0.3cm}
\label{table:computation_cost_comparison}
\end{table*}

\begin{figure*}[t]
	\centering
	%\framebox(500,100){}
	\includegraphics[width=1\linewidth]{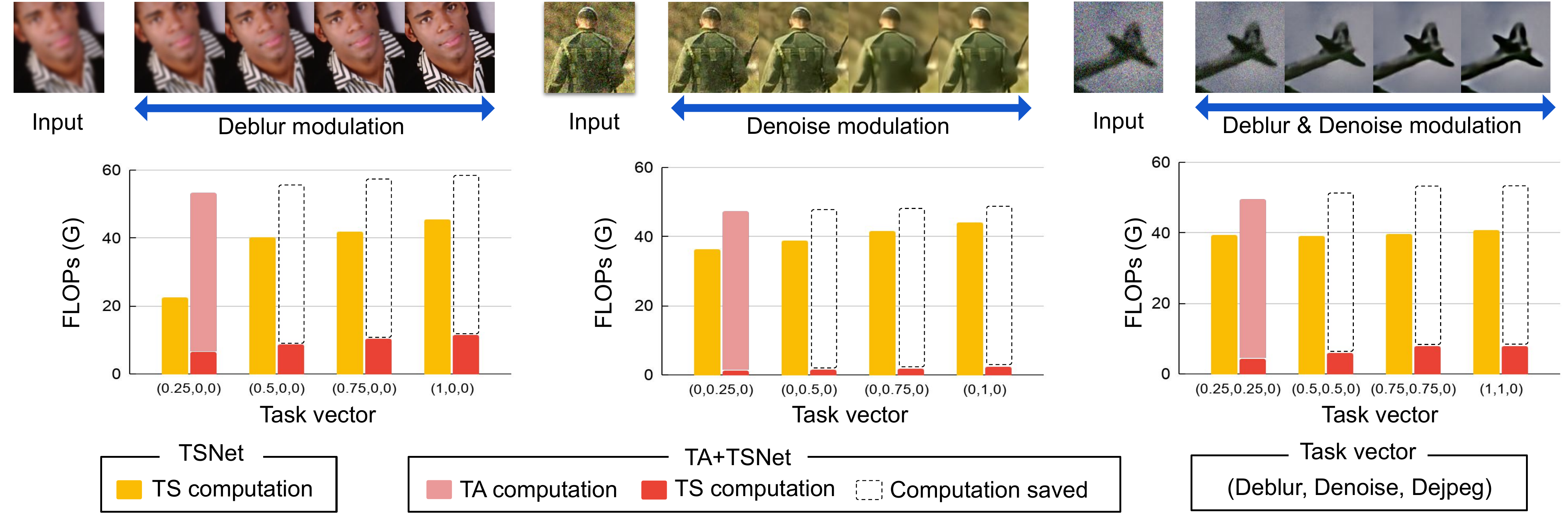}
	\vspace{-0.7cm}
	\caption{Computation cost comparisons in modulating task vectors. Each graph and a set of images present an image modulation example of removing Gaussian blur (left), Gaussian noise (middle), and joint degradation of Gaussian blur and noise (right). The images corresponding to task vectors with higher values present stronger restoration levels, which may require more computation costs for task-specific architectures. TA+TSNet is shown to be efficient for multiple inferences, owing to feature reuse of TS architecture across multiple inferences.
	}
	\vspace{-0.4cm}
	%\vspace{-1em}
	\label{fig:flops_wrt_task_vector}
\end{figure*}

\section{Experiments}
\label{sec:experiments}

This section presents the network efficiency comparison to the super network CResMD~\cite{jingwen2020interactive}, the state-of-the-art method in image modulation.
TSNet denotes our searched network only adopting task-specific architectures while TA+TSNet is the searched model in the joint space of task-agnostic and task-specific architectures.
TSNet-m and TA+TSNet-m respectively denote the variants of TSNet and TA+TSNet, which are searched using relative ground truth during training (see Figure~\ref{fig:relative_GT}).

\begin{figure*}[p]
        \begin{center}\centering
        \setlength{\tabcolsep}{0.02cm}
        \setlength{\columnwidth}{2.81cm}
        \hspace*{-\tabcolsep}\begin{tabular}{ccccccc}
            \multicolumn{1}{c}{\footnotesize Input}    
            &
            \footnotesize Task
            &
            \multicolumn{1}{c}{\footnotesize CResMD ~\cite{jingwen2020interactive}}
            &
            \multicolumn{1}{c}{ \footnotesize TSNet }
            &
            \multicolumn{1}{c}{\footnotesize TA+TSNet }
            &
            \multicolumn{1}{c}{\footnotesize TSNet-m }
            &
            \multicolumn{1}{c}{\footnotesize TA+TSNet-m }
            \\ \toprule[1.5pt]
            %\vspace{-0.1cm} \\
            &
             \begin{tabular}{c}
             \vspace{-2.8cm} \\  \footnotesize  Deblur 
             \end{tabular}
            &
             \includegraphics[width=0.15\linewidth]{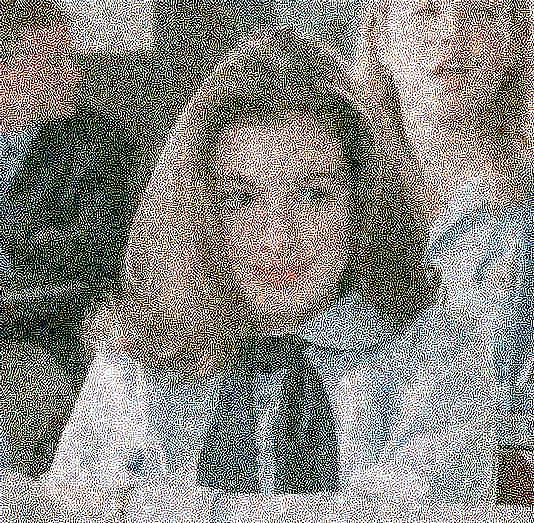}
            &
             \includegraphics[width=0.15\linewidth]{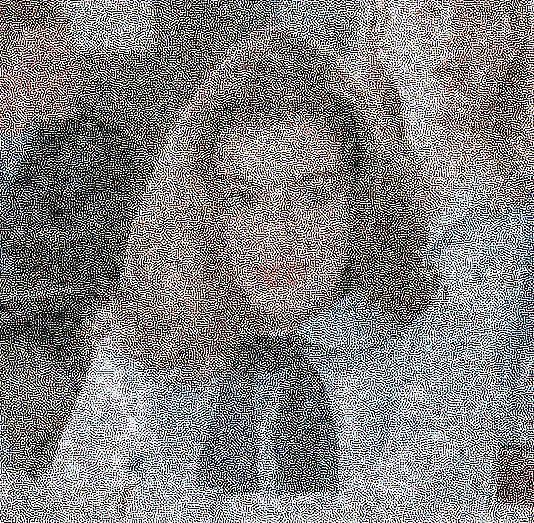}
            &
             \includegraphics[width=0.15\linewidth]{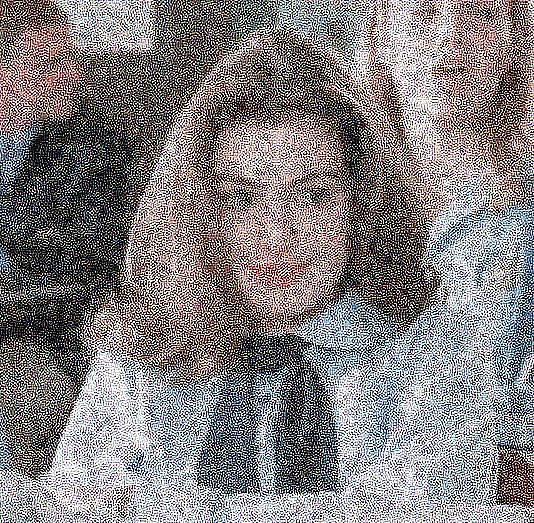}
            &
             \includegraphics[width=0.15\linewidth]{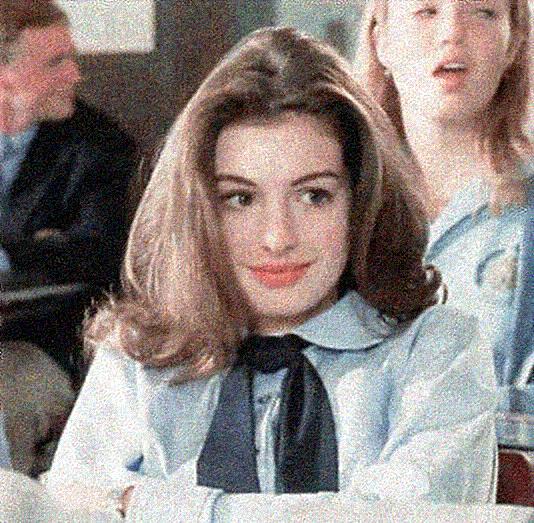}
            &
             \includegraphics[width=0.15\linewidth]{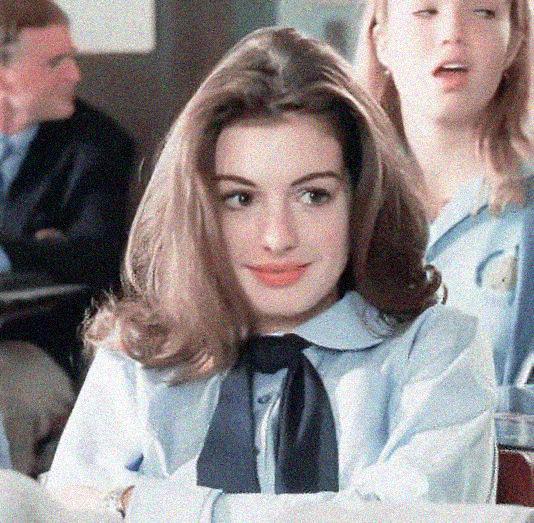}
             \vspace{-0.1cm} \\
            \includegraphics[width=0.15\linewidth]{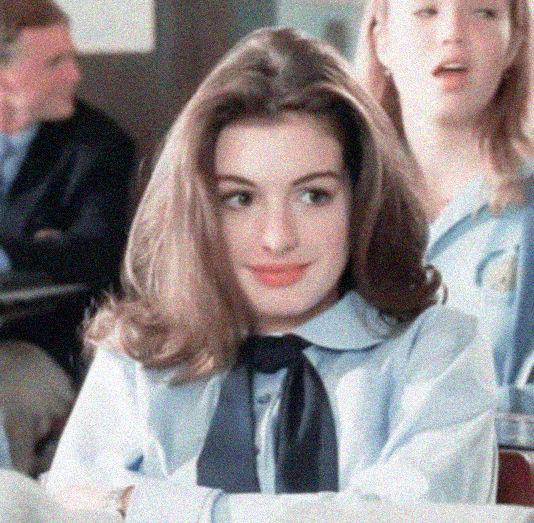}
            &
             \begin{tabular}{l}
             \vspace{-2.5cm} \\ \footnotesize ~~~Deblur   \vspace{-0.1cm}\\ \footnotesize + Dejpeg 
             \end{tabular}
            &
             \includegraphics[width=0.15\linewidth]{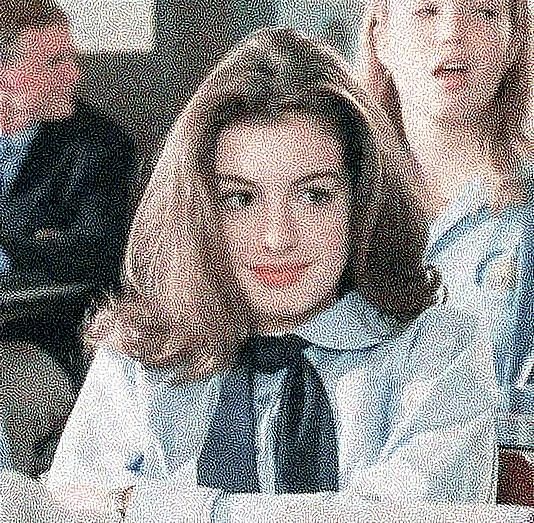}
            &
             \includegraphics[width=0.15\linewidth]{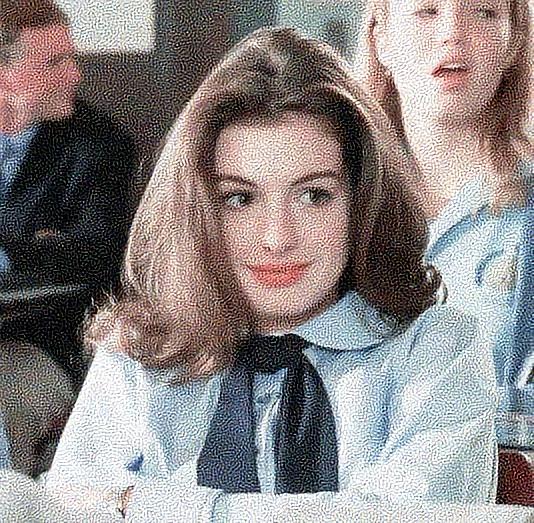}
            &
             \includegraphics[width=0.15\linewidth]{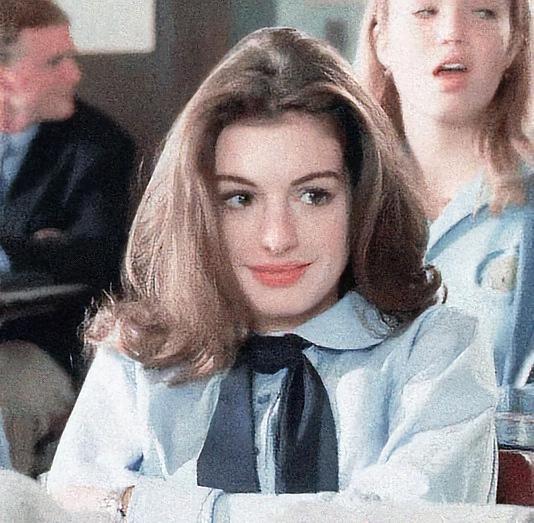}
            &
             \includegraphics[width=0.15\linewidth]{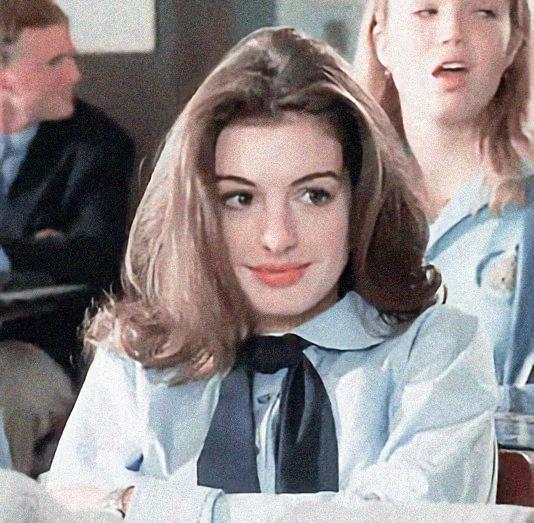}
            &
             \includegraphics[width=0.15\linewidth]{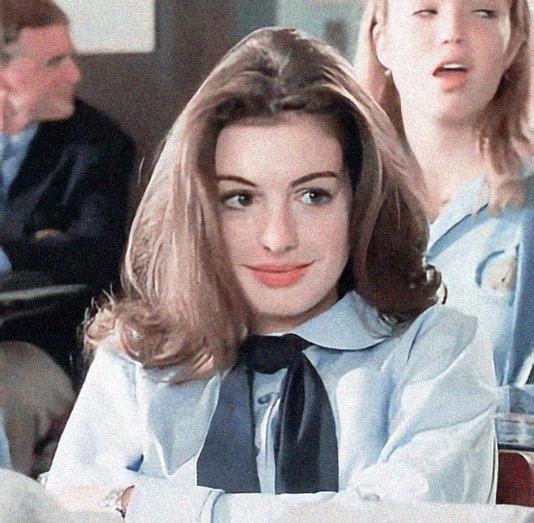}
             \vspace{-0.5cm} \\
             {\color{black} Synthetic }
             \\
            &
             \begin{tabular}{l}
             \vspace{-2.8cm} \\ \footnotesize ~~~Deblur  \vspace{-0.1cm} \\ \footnotesize + Denoise  \vspace{-0.1cm}\\ \footnotesize + Dejpeg 
             \end{tabular}
            &
             \includegraphics[width=0.15\linewidth]{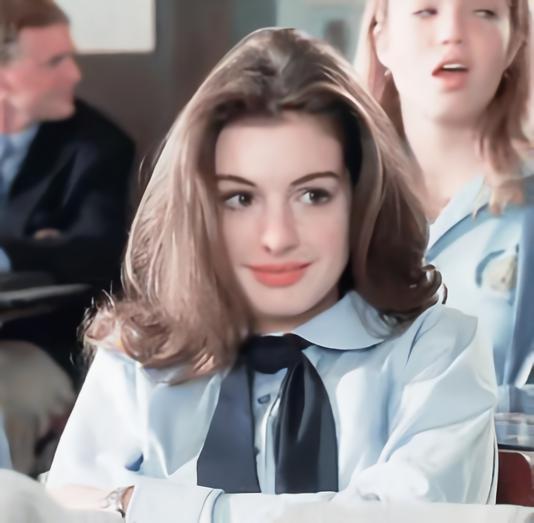}
            &
             \includegraphics[width=0.15\linewidth]{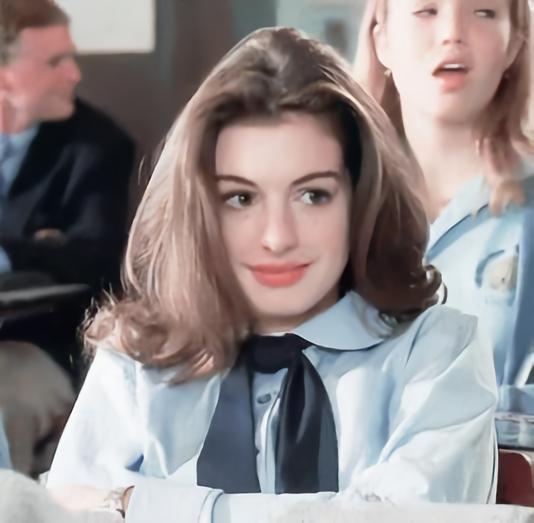}
            &
             \includegraphics[width=0.15\linewidth]{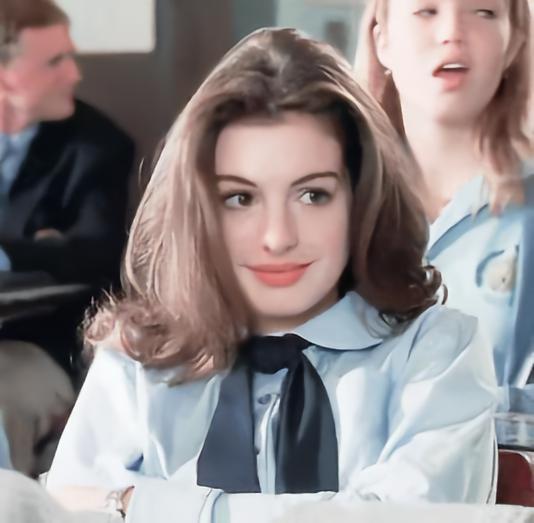}
            &
             \includegraphics[width=0.15\linewidth]{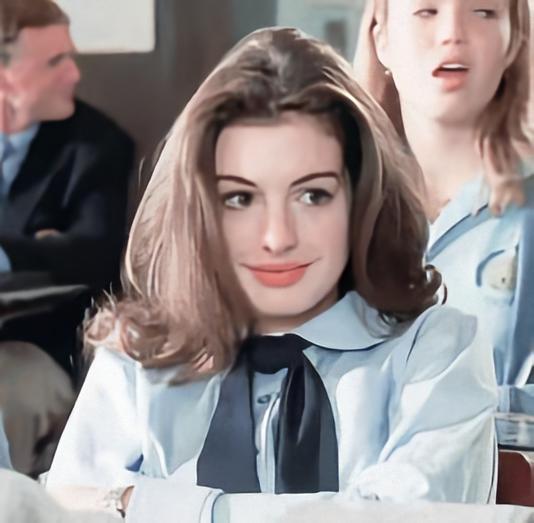}
            &
             \includegraphics[width=0.15\linewidth]{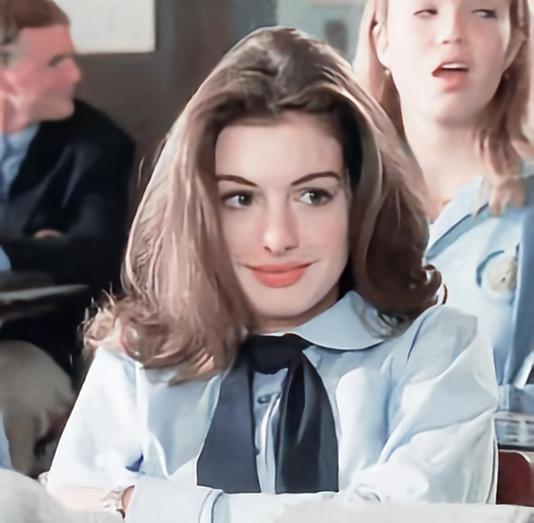}
            \\\cdashline{1-7} \\[-10pt]
            \includegraphics[width=0.15\linewidth]{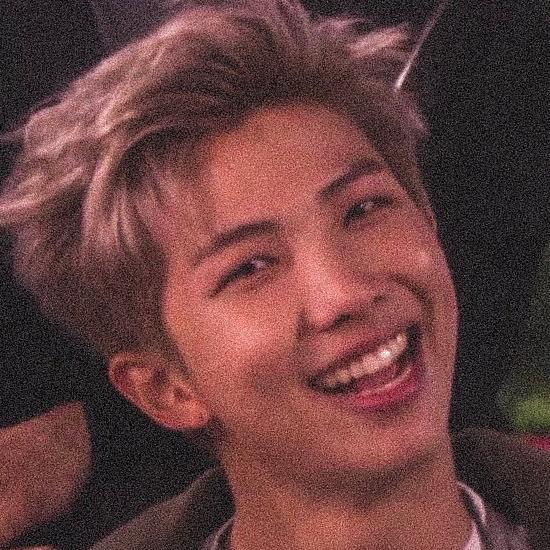} 
            &
             \begin{tabular}{l}
             \vspace{-2.8cm} \\ \footnotesize ~~~Deblur  \vspace{-0.1cm} \\ \footnotesize + Denoise  \vspace{-0.1cm}\\ \footnotesize + Dejpeg 
             \end{tabular}
            &
             \includegraphics[width=0.15\linewidth]{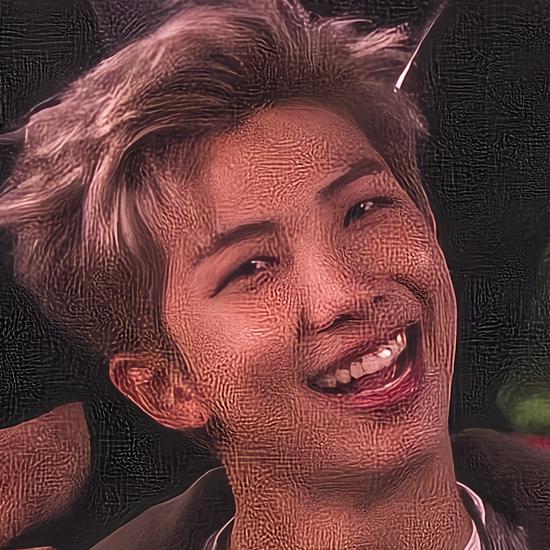}
            &
             \includegraphics[width=0.15\linewidth]{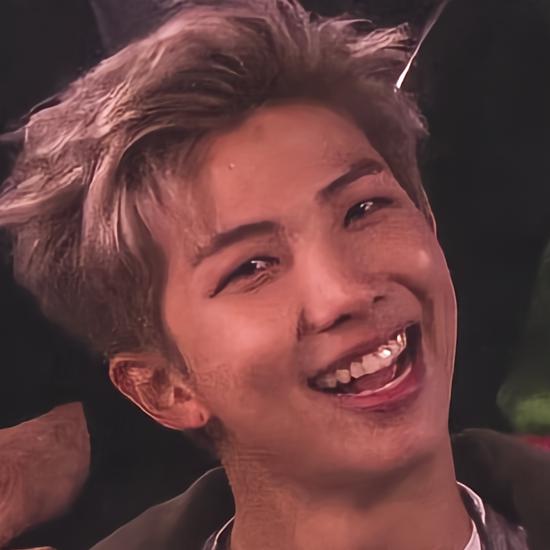}
            &
             \includegraphics[width=0.15\linewidth]{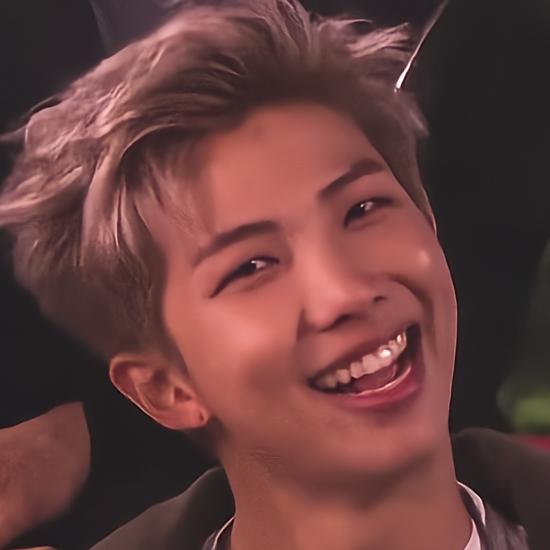}
            &
             \includegraphics[width=0.15\linewidth]{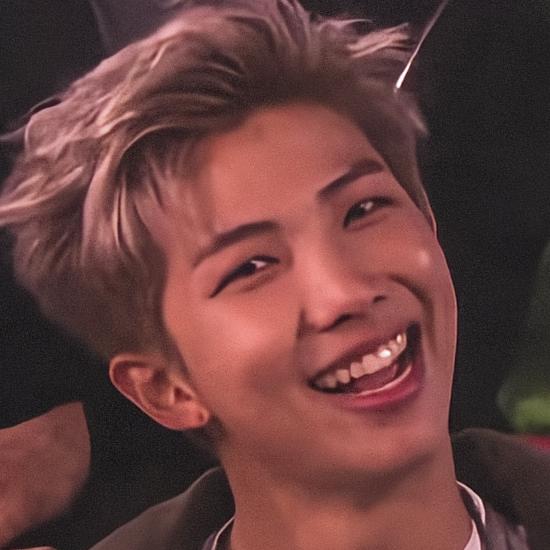}
            &
             \includegraphics[width=0.15\linewidth]{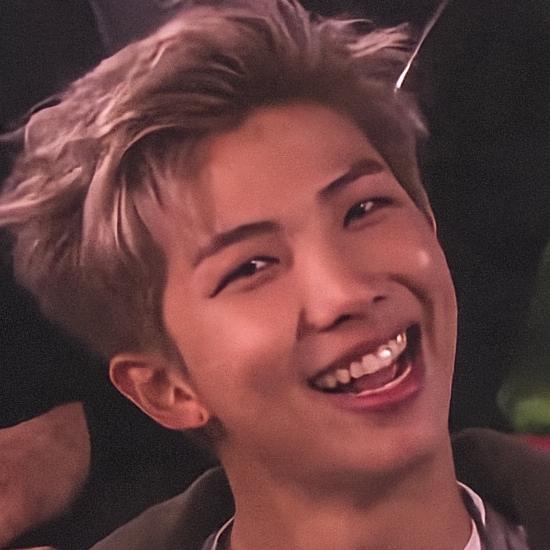}
             \vspace{-0.5cm} \\
             {\color{white} Real}
             \\
            \includegraphics[width=0.15\linewidth]{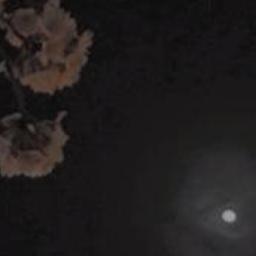}
            &
             \begin{tabular}{c}
             \vspace{-2.8cm} \\  \footnotesize Denoise 
             \end{tabular}
            &
             \includegraphics[width=0.15\linewidth]{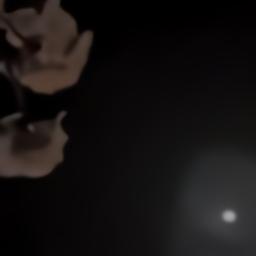}
            &
             \includegraphics[width=0.15\linewidth]{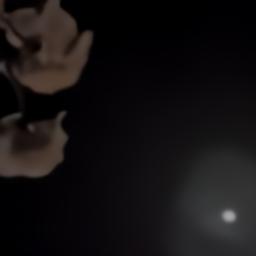}            
             &
             \includegraphics[width=0.15\linewidth]{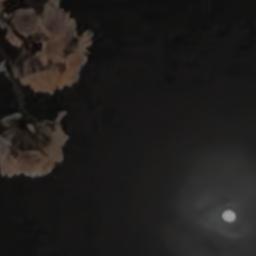}
             &
             \includegraphics[width=0.15\linewidth]{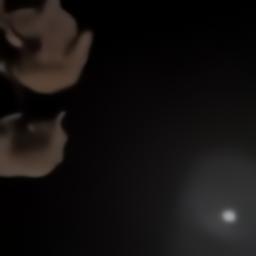}
             &
             \includegraphics[width=0.15\linewidth]{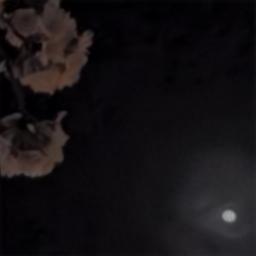}
             \vspace{-0.5cm} \\
             {\color{white} Real}
             \\
            \includegraphics[width=0.15\linewidth]{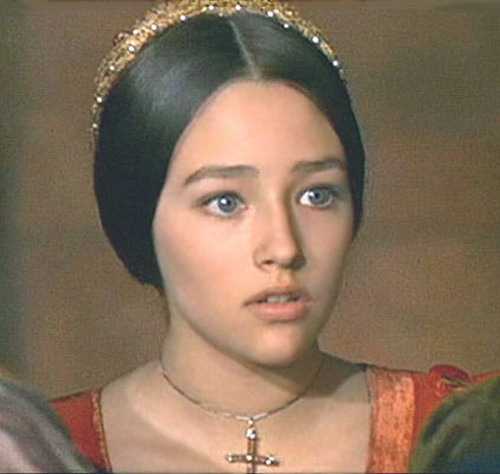}
            &
             \begin{tabular}{c}
             \vspace{-2.8cm} \\  \footnotesize Dejpeg 
             \end{tabular}
            &
             \includegraphics[width=0.15\linewidth]{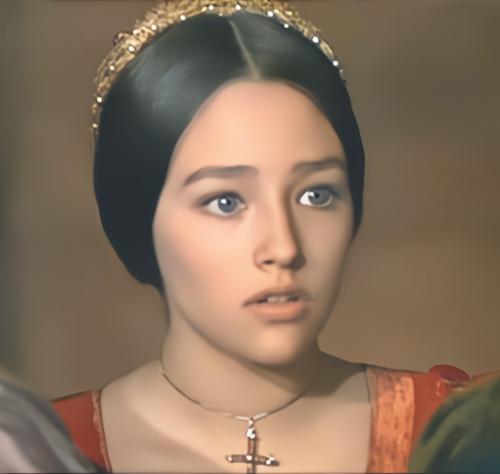}
            &
             \includegraphics[width=0.15\linewidth]{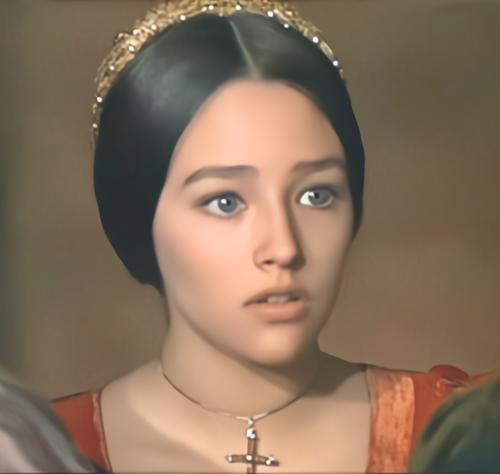}            
             &
             \includegraphics[width=0.15\linewidth]{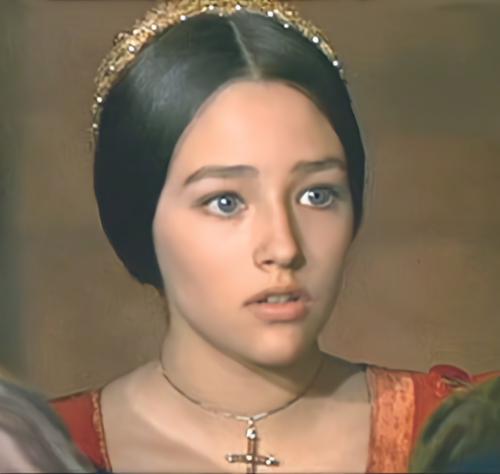}
             &
             \includegraphics[width=0.15\linewidth]{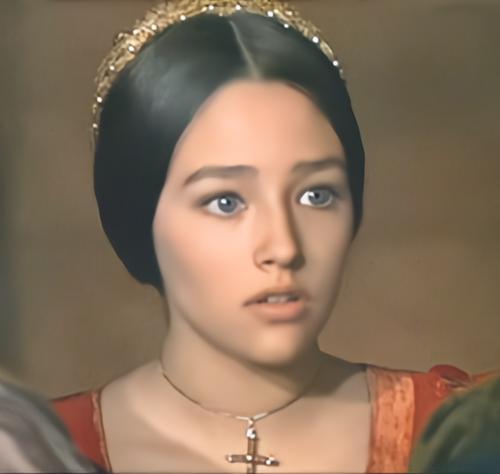}
             &
             \includegraphics[width=0.15\linewidth]{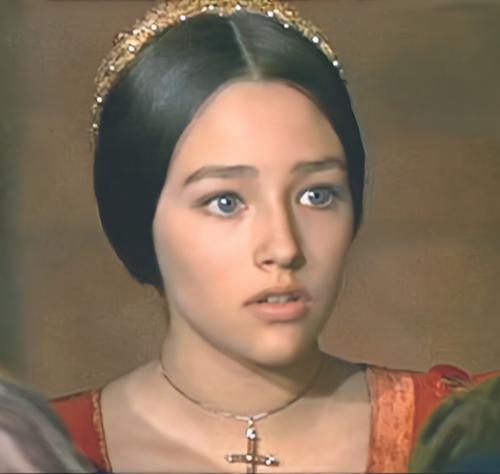}
             \vspace{-0.5cm} \\
             {\color{white} Real}
            \\\cdashline{1-7} \\[-10pt]

            \includegraphics[width=0.15\linewidth]{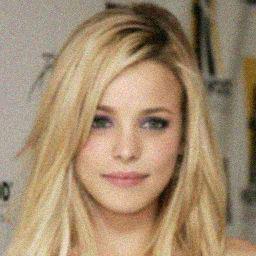}
            &
             \begin{tabular}{l}
             \vspace{-2.5cm} \\ \footnotesize ~~~Denoise   \vspace{-0.1cm}\\ \footnotesize + Dejpeg 
             \end{tabular}
            &
             \animategraphics[width=0.15\linewidth, poster=4, autoplay, palindrome, final, nomouse, method=widget]{1}{graphics/rachel1/CResMD/}{00}{10}
            &
             \animategraphics[width=0.15\linewidth, poster=4, autoplay, palindrome, final, nomouse, method=widget]{1}{graphics/rachel1/TSNet/}{00}{10}            
             &
             \animategraphics[width=0.15\linewidth, poster=4, autoplay, palindrome, final, nomouse, method=widget]{1}{graphics/rachel1/TA+TSNet/}{00}{10}     
             &
             \animategraphics[width=0.15\linewidth, poster=4, autoplay, palindrome, final, nomouse, method=widget]{1}{graphics/rachel1/TSNet-m/}{00}{10} 
             &
             \animategraphics[width=0.15\linewidth, poster=4, autoplay, palindrome, final, nomouse, method=widget]{1}{graphics/rachel1/TA+TSNet-m/}{00}{10}
             \vspace{-0.5cm} \\
             {\color{white}  \textit{Video}}
             \\
	            \includegraphics[width=0.15\linewidth]{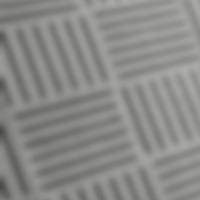}
            &
             \begin{tabular}{c}
             \vspace{-2.8cm} \\ \footnotesize   \vspace{-0.1cm} \\ \footnotesize Deblur  \vspace{-0.1cm}\\ \footnotesize  
             \end{tabular}
            &
             \animategraphics[width=0.15\linewidth, poster=7, autoplay, palindrome, final, nomouse, method=widget]{1}{graphics/urban/CResMD/blur/}{00}{10}
            &
             \animategraphics[width=0.15\linewidth, poster=7, autoplay, palindrome, final, nomouse, method=widget]{1}{graphics/urban/TSNet/blur/}{00}{10}            
             &
             \animategraphics[width=0.15\linewidth, poster=7, autoplay, palindrome, final, nomouse, method=widget]{1}{graphics/urban/TA+TSNet/blur/}{00}{10}     
             &
             \animategraphics[width=0.15\linewidth, poster=7, autoplay, palindrome, final, nomouse, method=widget]{1}{graphics/urban/TSNet-m/blur/}{00}{10} 
             &
             \animategraphics[width=0.15\linewidth, poster=7, autoplay, palindrome, final, nomouse, method=widget]{1}{graphics/urban/TA+TSNet-m/blur/}{00}{10}
             \vspace{-0.5cm} \\
             {\color{white} \textit{Video}}
             \\
	\toprule[1.5pt]
        \end{tabular}\vspace{-0.3cm}
        \captionof{figure}{\textbf{Controllable image restoration examples}. Our TA+TSNet-m modulates diverse imagery effects with respect to the given restoration tasks. It generates less auxiliary visual artifacts and less over-smoothed textures in both synthetic and real images.  \textit{This is a video figure that is best viewed using Adobe Reader.}}\vspace{-0.0cm}
        \label{fig:image_results}
    \end{center}
	\vspace{-0.0cm}
\end{figure*}

%%%%%%%%%%%%%%%%%%%%%%%%%%%%%%%%%%%%%%

\subsection{Computation cost comparison}
\paragraph{Latency and FLOPs reduction.}
Table~\ref{table:computation_cost_comparison} presents the computation cost comparisons to the super network CResMD~\cite{jingwen2020interactive} across diverse image resolutions and devices.
All models demonstrate comparable image restoration performances, which will be described in the next section.
FLOPs are dramatically reduced in both TSNet (75.6\%) and TA+TSNet (95.7\%), while CResMD requires 10 T FLOPs per imagery effect.
Our models boast faster speed than the baseline on all reported devices.
TA+TSNet is 3.8 times on a single-core CPU, 3.1 times on a multi-core CPU, and 3.7 times on a GPU faster than CResMD when generating 4K (3840$\times$2160) imagery effects.
TA+TSNet only requires 0.07s to generate a HD (1280$\times$720) imagery effect.

\vspace{-0.3cm}
\paragraph{Code optimization.}
For fair comparisons, we optimize the official CResMD codes~\footnote{https://github.com/hejingwenhejingwen/CResMD} for fast tensor computing, which results in 2 times faster speed in GPU latency. 
We also observe an interesting phenomenon, where the latency reduction ratio from CResMD becomes small in low resolution images.
This phenomenon emanates from memory allocation of the feature map and weight tensor slicing in task-specific architectures.
These additional computation costs become more noticeable when the image resolution is small.
Since FLOPs of convolution operations increase quadratically, the latency of the proposed method is also much faster than the baseline in 4K image resolution.
We apply tensor slicing only for the middle channel selection module in residual blocks to measure GPU latency, which expects extra image quality improvement with more parameters.

\vspace{-0.3cm}
\paragraph{Computation costs visualization.}
Figure~\ref{fig:flops_wrt_task_vector} presents FLOPs of TSNet and TA+TSNet in various task vectors to generate HVGA (481$\times$321) resolution imagery effects from CBSD68 dataset.
Task-specific parts in TSNet and TA+TSNet tend to require more channels for higher restoration levels.
Task-agnostic part in TA+TSNet is computed only once for each input image and requires substantially small amount of computation from the second effect.
Note that TA+TSNet requires higher computation costs than TSNet for the first inference.

\begin{table}[t]
\centering
%\vspace{-0.3cm}
\caption{Image restoration performances on CBSD68. 
%(Up)
}
\vspace{-0.2cm}\hspace{-0.3cm}
\scalebox{0.9}{ 
\setlength{\tabcolsep}{7.5pt} % Default value: 6pt
\begin{tabular}{rcccc}
	\toprule[1.5pt]
      Method & PSNR$_\uparrow$ & SSIM$_\uparrow$ & NIQE$_\downarrow$ & LPIPS$_\downarrow$ \\       
     \midrule[1.0pt]
     CResMD &  25.86 dB & 0.8195 & 5.21  & 0.3737  \\ \cdashline{1-5} \\[-10pt]
    TSNet  &   25.76 dB & 0.8184 & 4.79 & 0.3797  \\
    TA+TSNet & 25.75 dB & 0.8179 & 4.51  & 0.3548 \\ \cdashline{1-5} \\[-10pt]
    TSNet-m  &   25.62 dB & 0.8144 & 4.86 & 0.3830  \\
    TA+TSNet-m & 25.64 dB & 0.8137 & 5.11  & 0.3752 \\
	\toprule[1.5pt]
\end{tabular}}
%\vspace{-0.2cm}
\vspace{-1em}
\label{table:restoration_performance}
\end{table}

\subsection{Image modulation comparison}

\paragraph{Image restoration performance.}
Table~\ref{table:restoration_performance} presents the average best-score among 27 imagery effects per input image over dataset. 
PSNR scores of TSNet and TA+TSNet drop from upper bound performances of CResMD by $0.11$ dB whereas they exhibit better performances IN NIQE measure. 
TSNet-m and TA+TSNet-m may seem to perform worse than their counterparts trained with absolute ground truth, according to Table~\ref{table:restoration_performance}.
However, the metrics presented in the table are limited in that they do not accurately quantify the image modulation quality.
We support argument by providing generated images from diverse task vectors in Figure~\ref{fig:image_results}.
%To this end, Figure~\ref{} presents PSNR scores with respect to relative ground truth images.

\vspace{-0.3cm}
\paragraph{Qualitative results.}
Figure~\ref{fig:image_results} presents controllable image restoration examples of CResMD and our models.
All results per imagery effect are generated by models with a given task vector.
The synthetic examples have blur, noise, and/or jpeg compression.
CResMD and TSNet are shown to generate critical artifacts when the task vector does not match the level of degradation present in the image. 
By constrast, TSNet-m and TA+TSNet-m generate successful image modulation results, in which degradation is removed to a certain extent specified by a task vector. 
The contributions of our proposed method to image modulation performances become clearer in a real-world scenario, where real examples with degradation are downloaded from the internet.
Interestingly, TA+TSNet and TA+TSNet-m, which learn task-agnostic feature maps, generate visually pleasing imagery effects without over-smoothing. 
Moreover, TSNet-m and TA+TSNet-m presents dynamic image modulation results on video examples with diverse task vectors.

 \begin{figure}[t]
	\centering
	%\framebox(200,100){}
	\includegraphics[trim=0.5cm 0cm 0cm 0cm, width=0.75\linewidth]{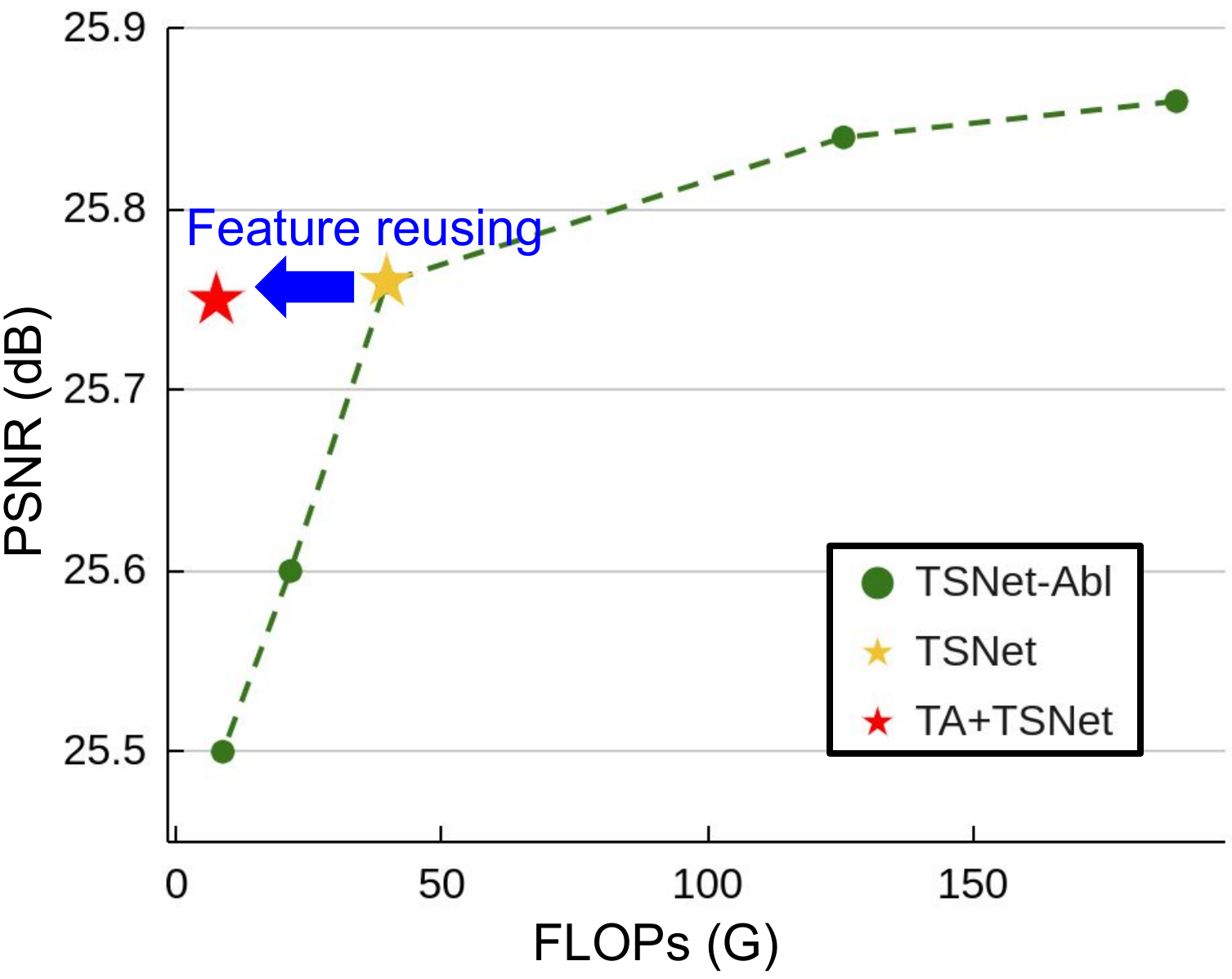}\\\vspace{-0.2cm}
	\caption{Ablation study of TSNet on CBSD68. Models of TSNet-Abl are searched by different values of $\lambda_1$. 
	}
	\vspace{-0.4cm}
	\label{fig:TSNet_ablation}
\end{figure}

\section{Ablation study}
\label{sec:ablation_study}

\paragraph{Extreme FLOPs reduction in TSNet.}
Figure~\ref{fig:TSNet_ablation} presnets the accuracy-FLOPs trade-off of TSNet, where TSNet-Abl denotes the variations of TSNet trained with different values of the hyperparameter $\lambda_1$ in Equation~\eqref{eq:objective}.
%\eqref{eq:objective}. 
We gradually increase the value of the hyperparameter to search extremely fast TSNet by task-specific channel pruning.
%$\lambda_1$ from 0 to $5\times10^-10$ while $\lambda_2$ set to 0.
%The PSNR scores of TSNet-Abl reaches to CResMD when $\lambda_1$ of 0 and gradually decrease with higher values of $\lambda_1$.
TSNet-Abl presents quite noticeable efficiency improvement.  
However, comparing with TA+TSNet, TSNet-Abl presents significant PSNR drop under comparable FLOPs.

\vspace{-0.3cm}
\paragraph{Searching without pruning.}
Table~\ref{table:TANet_ablation} describes the variations of TA+TSNet (TA+TSNet-Abl) which manually determine the number of shared layers without channel pruning. 
TA+TSNet-Abl presents comparable network efficiency to TSNet when generating multiple imagery effects ($M=27$).
However, TA+TSNet-Abl can not reduce the computation costs of the single inference from the baseline ($M=1$), which are crucial in the real-world scenario, while TA+TSNet reduces it by 3.6 times on CBSD68 dataset.

\vspace{-0.3cm}
\paragraph{Balancing hyperparameters.}
Table~\ref{table:TA+TSNet_ablation} presents the ablation study of hyperparameters $\lambda_1$ and $\lambda_2$ which balance the trade-off between the network computation cost $\mathcal{R}_\text{total}^\text{TA+TS}(f,\boldsymbol{x})$ and the number of shared layers while minimizing Equation~\eqref{eq:objective}.
The models trained with small $\lambda_1$ and large $\lambda_2$ have large portions of shared layers, and thus they are efficient in generating multiple imagery effects (\textcircled{\footnotesize{2}} {\it vs.} \textcircled{\footnotesize{3}} and \textcircled{\footnotesize{5}} {\it vs.} \textcircled{\footnotesize{4}}).
In contrast, the models trained with the opposite balance between $\lambda_1$ and $\lambda_2$ are efficient for a single inference (\textcircled{\footnotesize{2}} {\it vs.} \textcircled{\footnotesize{1}} and \textcircled{\footnotesize{5}} {\it vs.} \textcircled{\footnotesize{6}}).

\begin{table}[t]
    \centering
    %\vspace{-0.2cm}
    \caption{Ablation study for TA+TSNet. Models of TA+TSNet-Abl are trained with the different number of shared layers (SL).}
    \vspace{-0.2cm}\hspace{-0.3cm}
    \scalebox{0.9}{ 
    \setlength{\tabcolsep}{6.5pt} % Default value: 6pt
    \begin{tabular}{rcccc}
    	\toprule[1.5pt]
         Method & \multirow{2}{*}{SL}  & \multirow{2}{*}{PSNR} & \multicolumn{2}{c}{FLOPs (G)} \\
        (on CBSD68)    &  &   & $M=1$ & $M=27$ \\
         \midrule[1.0pt]
        CResMD & 0 \% & 25.86 dB  & 189.1  & 189.1\\
        \cdashline{1-5} \\[-10pt]
        \multirow{5}{*}{TA+TSNet-Abl } & 0 \%  &  25.86 dB  & 189.1 & 189.1  \\
         & 31 \% & 25.83 dB   & 189.1 & 132.0 \\
         & 62 \% & 25.81 dB  & 189.1 & 69.5 \\
         & 85 \% & 25.75 dB  & 189.1 & 32.5\\
         & 99 \% & 25.34 dB  & 189.1 & 7.0\\
        \cdashline{1-5} \\[-10pt]
        TSNet & 0 \% & 25.76 dB  & 40.8 & 39.6 \\
        TA+TSNet & 62 \% & 25.75 dB  & 52.9 & 7.5 \\
         \toprule[1.5pt]
    \end{tabular}}
    \vspace{-0.2cm}
    \label{table:TANet_ablation}
\end{table}

\begin{table}[t]
    \centering
    %\vspace{-0.2cm}
    \caption{Ablation study of $\lambda_1$ and $\lambda_2$ on CBSD68. SL denotes the percentage of the shared layers in the super network.
    }
    %\vspace{0.2cm}
    \vspace{-0.2cm}\hspace{-0.3cm}
    \scalebox{0.9}{ 
    \setlength{\tabcolsep}{4pt} % Default value: 6pt
    \begin{tabular}{ccccccc}
    	\toprule[1.5pt]
         \multirow{2}{*}{Ex.\#}&\multirow{2}{*}{$\lambda_1$} & \multirow{2}{*}{$\lambda_2$} & \multirow{2}{*}{SL}  & PSNR & \multicolumn{2}{c}{FLOPs (G)} \\
         &  &   &  &  (dB) & $M=1$ & $M=27$ \\
         \midrule[1.0pt]
       \textcircled{\footnotesize{1}}   & & $1\times 10^{-3}$ & 18 \%  &  25.67  & 35.2 & 23.1  \\
       \textcircled{\footnotesize{2}} & $5\times 10^{-11}$ & $1\times 10^{-2}$ & 62 \% & 25.75  & 52.9 & 7.5 \\
       \textcircled{\footnotesize{3}} &   &$1\times 10^{-1}$ & 99 \% & 25.48  & 125.5 & 4.8 \\
        \cdashline{1-7} \\[-10pt]
      \textcircled{\footnotesize{4}}  &  $5\times 10^{-12}$ &  & 99 \% & 25.46 & 154.6 &  6.0  \\
         %& 65 & 25.69 & 1.12  \\
       \textcircled{\footnotesize{5}} &  $5\times 10^{-11}$ & $1\times 10^{-2}$ & 62 \% &  25.75 & 52.9 & 7.5\\
      \textcircled{\footnotesize{6}} &    $5\times 10^{-10}$ &  & 16 \% & 25.50  &  15.4 & 1.9 \\
        %TA+TSNet & 62 \% & 25.75 dB  & 52.9 & 7.5 \\
         \toprule[1.5pt]
    \end{tabular}}
    \vspace{-0.4cm}
    \label{table:TA+TSNet_ablation}
\end{table}

\section{Conclusion}

We propose a novel neural architecture search algorithm to find efficient networks for controllable image restoration (or image modulation).
In particular, the proposed algorithm performs task-agnostic and task-specific channel pruning.
The resulting pruned networks not only reduce network computation costs of the state-of-the-art network greatly but also provide better image quality during modulation.
The feature reuse strategy across tasks pushes the network efficiency further when comparing multiple imagery effects in practice.

\clearpage
\newpage
\newpage
{\small
\bibliographystyle{ieee_fullname}
\bibliography{egbib}

\begin{thebibliography}{10}\itemsep=-1pt

\bibitem{agustsson2017ntire}
Eirikur Agustsson and Radu Timofte.
\newblock Ntire 2017 challenge on single image super-resolution: Dataset and
  study.
\newblock In {\em CVPRW}, 2017.

\bibitem{ahn2018fast}
Namhyuk Ahn, Byungkon Kang, and Kyung-Ah Sohn.
\newblock Fast, accurate, and lightweight super-resolution with cascading
  residual network.
\newblock In {\em ECCV}, 2018.

\bibitem{Chu2019FastAA}
Xiangxiang Chu, Bo Zhang, Hailong Ma, Ruijun Xu, Jixiang Li, and Qingyuan Li.
\newblock Fast, accurate and lightweight super-resolution with neural
  architecture search.
\newblock In {\em ICPR}, 2020.

\bibitem{dong2015compression}
Chao Dong, Yubin Deng, Chen~Chang Loy, and Xiaoou Tang.
\newblock Compression artifacts reduction by a deep convolutional network.
\newblock In {\em ICCV}, 2015.

\bibitem{dong2014image}
Chao Dong, Chen~Change Loy, Kaiming He, and Xiaoou Tang.
\newblock Image super-resolution using deep convolutional networks.
\newblock {\em TPAMI}, 2014.

\bibitem{dong2016accelerating}
Chao Dong, Chen~Change Loy, and Xiaoou Tang.
\newblock Accelerating the super-resolution convolutional neural network.
\newblock In {\em ECCV}, 2016.

\bibitem{jingwen2019modulating}
Jingwen He, Chao Dong, and Yu Qiao.
\newblock Modulating image restoration with continual levels via adaptive
  feature modification layers.
\newblock In {\em CVPR}, 2019.

\bibitem{jingwen2020interactive}
Jingwen He, Chao Dong, and Yu Qiao.
\newblock Interactive multi-dimension modulation with dynamic controllable
  residual learning for image restoration.
\newblock In {\em ECCV}, 2020.

\bibitem{Hou2020efficient}
Z. {Hou} and S. {Kung}.
\newblock Efficient image super resolution via channel discriminative deep
  neural network pruning.
\newblock In {\em ICASSP}, 2020.

\bibitem{kim2019fine}
Heewon Kim, Seokil Hong, Bohyung Han, Heesoo Myeong, and Kyoung~Mu Lee.
\newblock Fine-grained neural architecture search.
\newblock {\em arXiv preprint arXiv:1911.07478}, 2019.

\bibitem{kim2016accurate}
Jiwon Kim, Jungkwon Lee, and Kyoung~Mu Lee.
\newblock Accurate image super-resolution using very deep convolutional
  networks.
\newblock In {\em CVPR}, 2016.

\bibitem{ledig2016photo}
Christian Ledig, Lucas Theis, Ferenc Huszar, Jose Caballero, Andrew Cunningham,
  Alejandro Acosta, Andrew Aitken, Alykhan Tejani, Johannes Totz, Zehan Wang,
  and Wenzhe Shi.
\newblock Photo-realistic single image super-resolution using a generative
  adversarial network.
\newblock In {\em CVPR}, 2017.

\bibitem{bee2017enhanced}
Bee Lim, Sanghyun Son, Heewon Kim, Seungjun Nah, and Kyoung~Mu Lee.
\newblock Enhanced deep residual networks for single image super-resolution.
\newblock In {\em CVPRW}, 2017.

\bibitem{liu2018darts}
Hanxiao Liu, Karen Simonyan, and Yiming Yang.
\newblock {DARTS}: Differentiable architecture search.
\newblock In {\em ICLR}, 2019.

\bibitem{liu2020deep}
Ming Liu, Zhilu Zhang, Liya Hou, Wangmeng Zuo, and Lei Zhang.
\newblock Deep adaptive inference networks for single image super-resolution.
\newblock In {\em ECCVW}, 2020.

\bibitem{MartinFTM01}
D. Martin, C. Fowlkes, D. Tal, and J. Malik.
\newblock A database of human segmented natural images and its application to
  evaluating segmentation algorithms and measuring ecological statistics.
\newblock In {\em ICCV}, 2001.

\bibitem{mittal2013niqe}
A. Mittal, Soundararajan R., and Bovik. A., C.
\newblock Making a completely blind image quality analyzer.
\newblock {\em IEEE Signal Processing Letters}, 2013.

\bibitem{alon2019dynamic}
Alon Shoshan, Roey Mechrez, and Lihi Zelnik-Manor.
\newblock Dynamic-net: Tuning the objective without re-training for synthesis
  tasks.
\newblock In {\em ICCV}, 2019.

\bibitem{Song2019EfficientRD}
Dehua Song, Chang Xu, Xu Jia, Yiyi Chen, Chunjing Xu, and Yunhe Wang.
\newblock Efficient residual dense block search for image super-resolution.
\newblock In {\em AAAI}, 2020.

\bibitem{suganuma2019attention}
Masanori Suganuma, Xing Liu, and Takayuki Okatani.
\newblock Attention-based adaptive selection of operations for image
  restoration in the presence of unknown combined distortions.
\newblock In {\em CVPR}, 2019.

\bibitem{wei2019cfsnet}
Wei Wang, Ruiming Guo, Yapeng Tian3, and Wenming Yang.
\newblock Cfsnet: Toward a controllable feature space for image restoration.
\newblock In {\em ICCV}, 2019.

\bibitem{xintao2019deep}
Xintao Wang, Ke Yu, Chao Dong, Xiaoou Tang, and Chen~Change Loy.
\newblock Deep network interpolation for continuous imagery effect transition.
\newblock In {\em CVPR}, 2019.

\bibitem{jingwei2020binarized}
Jingwei Xin, Nannan Wang, Xinrui~JiangJie Li, Heng Huang, and Xinbo Gao.
\newblock Binarized neural network for single image super resolution.
\newblock In {\em ECCV}, 2020.

\bibitem{yu2018crafting}
Ke Yu, Chao Dong, Liang Lin, and Chen~Change Loy.
\newblock Crafting a toolchain for image restoration by deep reinforcement
  learning.
\newblock In {\em CVPR}, 2018.

\bibitem{yu2019path}
Ke Yu, Xintao Wang, Chao Dong, Xiaoou Tang, and Chen~Change Loy.
\newblock Path-restore: Learning network path selection for image restoration.
\newblock {\em arXiv preprint arXiv:1904.10343}, 2019.

\bibitem{yuan2020efficient}
Yuan Yuan, Wei Su, and Dandan Ma.
\newblock Efficient dynamic scene deblurring using spatially variant
  deconvolution network with optical flow guided training.
\newblock In {\em CVPR}, 2020.

\bibitem{zhang2017beyond}
Kai Zhang, Wangmeng Zuo, Yunjin Chen, Deyu Meng, and Lei Zhang.
\newblock Beyond a {Gaussian} denoiser: Residual learning of deep {CNN} for
  image denoising.
\newblock {\em TIP}, 2017.

\bibitem{zhang2018ffdnet}
Kai Zhang, Wangmeng Zuo, and Lei Zhang.
\newblock Ffdnet: Toward a fast and flexible solution for {CNN} based image
  denoising.
\newblock {\em TIP}, 2018.

\bibitem{zhang2018perceptual}
Richard Zhang, Phillip Isola, Alexei~A Efros, Eli Shechtman, and Oliver Wang.
\newblock The unreasonable effectiveness of deep features as a perceptual
  metric.
\newblock In {\em CVPR}, 2018.

\bibitem{zhang2018residual}
Yulun Zhang, Yapeng Tian, Yu Kong, Bineng Zhong, and Yun Fu.
\newblock Residual dense network for image super-resolution.
\newblock In {\em CVPR}, 2018.

\end{thebibliography}
}

\clearpage
% !TEX root = main.tex

\section{Supplementary Material}

This section first presents an analysis of image modulation performance in Section~\ref{sec:modulation_analysis} and additional qualitative results of image modulation in Section~\ref{sec:additional_qualitative_results}.
Then, Section~\ref{sec:implementation_details} presents our implementation details and Section~\ref{sec:evaluation_metric} presents our evaluation metrics.

\subsection{Image modulation performance analysis}
\label{sec:modulation_analysis}
\begin{figure}[t]
	\centering
	%\framebox(200,100){}
	\includegraphics[width=0.8\linewidth]{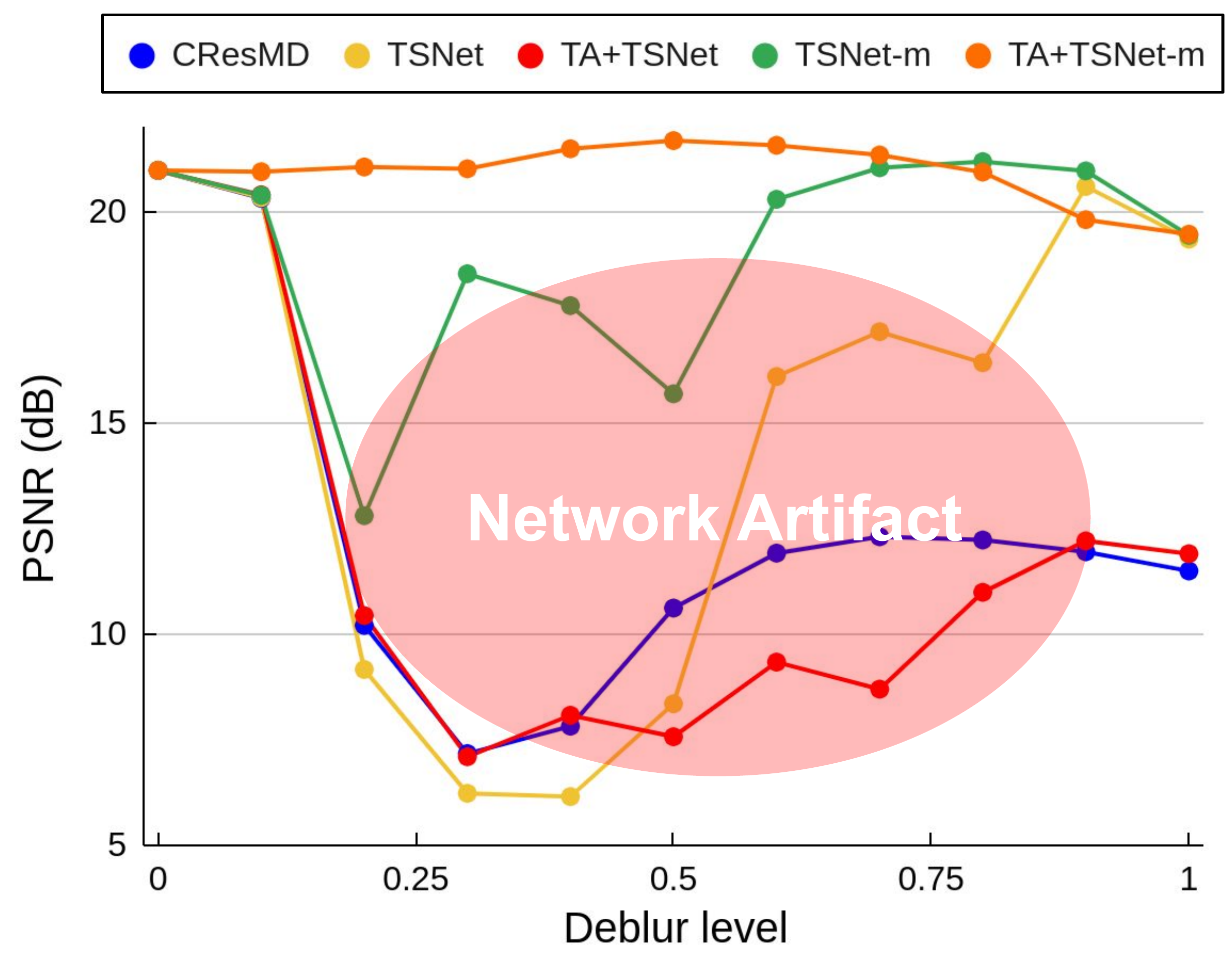}\\
	\vspace{-0.1cm}
	\footnotesize{(a) Deblur modulation to blur, noise, and jpeg images (mismatched case)}\\ \vspace{0.2cm}
	\includegraphics[width=0.8\linewidth]{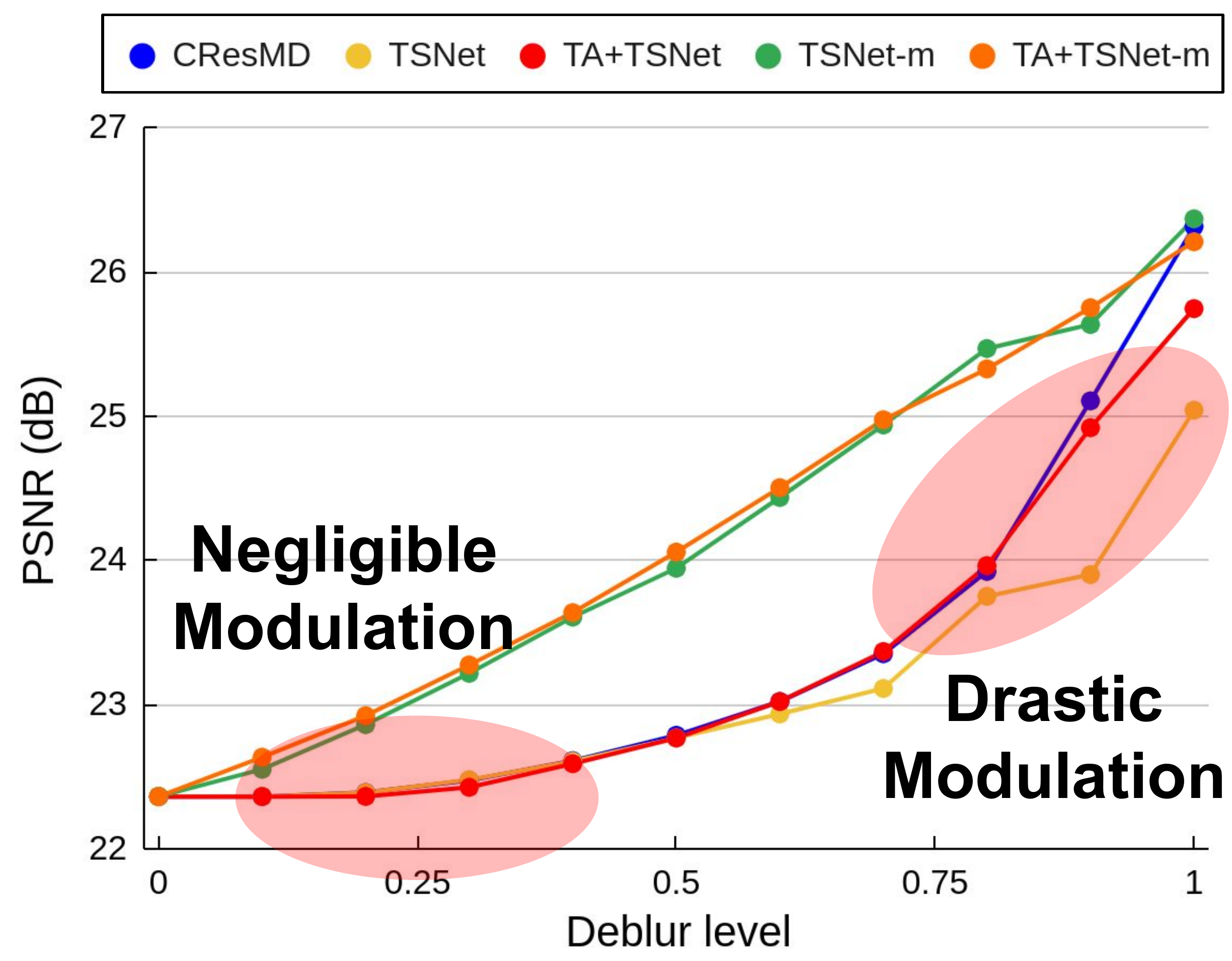}\\
	\vspace{-0.1cm}
	\footnotesize{(b) Deblur modulation to blur images (matched case)}\\
	%\vspace{-0.2cm}
	\caption{Two image modulation problems in CResMD. (a) Network artifact generation when modulating unmatched restoration tasks to the input images. (b) Uneven image modulations across task levels: Negligible modulation at lower deblur levels while drastic image modulation at higher deblur levels.
	}
	\vspace{-0.2cm}
	\label{fig:downgrade}
\end{figure}

We analyze the image modulation performances by the image quality of imagery effects from the models.
Figure~\ref{fig:downgrade} summarizes two problems with CResMD during modulation, namely (a) artifact in generated images and (b) uneven modulation across task vectors, which downgrade user experience in controlling imagery effects. 
The first problem with CResMD is the susceptibility to artifact in the generated images during the modulation process when the restoration types of task vectors are mismatched to the degradation types of input images. 
The modulation process to the real-world images includes such mismatch scenarios since the degradation types of the real-world images are hard to determine. 
As such, users have to sweep through various task vectors to find the output image that best meets user preferences or to find the task vector that successfully restores the degraded input image. 
During the process of modulating with various task vectors, the degradation type and restoration type are bound to mismatch.
Another problem with CResMD is the uneven modulation across task levels, as exemplified in Figure~\ref{fig:downgrade}(b).
Negligible modulation or drastic modulation, as shown in the figure, can result in discrete restoration levels, restricting the range of output images users can explore. 
Appendix presents qualitative analysis of these problems.

\subsection{Additional qualitative results}
\label{sec:additional_qualitative_results}
\begin{figure*}[p]
        \begin{center}\centering
        \setlength{\tabcolsep}{0.02cm}
        \setlength{\columnwidth}{2.81cm}
        \hspace*{-\tabcolsep}\begin{tabular}{ccccccc}
            \multicolumn{1}{c}{\footnotesize Input}    
            &
            \footnotesize Task vector
            &
            \multicolumn{1}{c}{\footnotesize CResMD~\cite{jingwen2020interactive}}
            &
            \multicolumn{1}{c}{ \footnotesize TSNet }
            &
            \multicolumn{1}{c}{\footnotesize TA+TSNet }
            &
            \multicolumn{1}{c}{\footnotesize TSNet-m }
            &
            \multicolumn{1}{c}{\footnotesize TA+TSNet-m }
            \\ \toprule[1.5pt]
            %\vspace{-0.1cm} \\
            &
             \begin{tabular}{c}
             \vspace{-2.8cm} \\  \footnotesize  (0,0,0) 
             \end{tabular}
            &
             \includegraphics[width=0.15\linewidth]{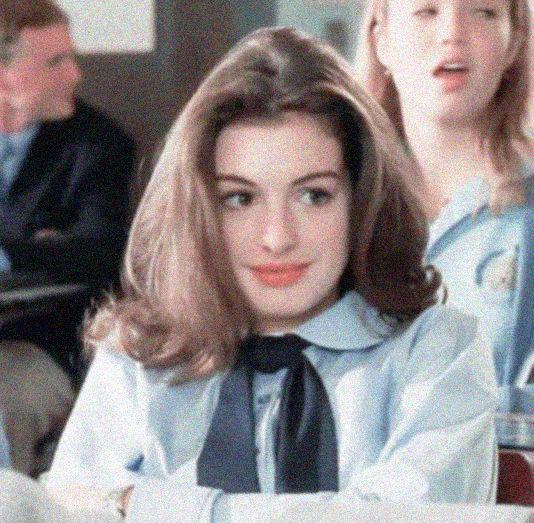}
            &
             \includegraphics[width=0.15\linewidth]{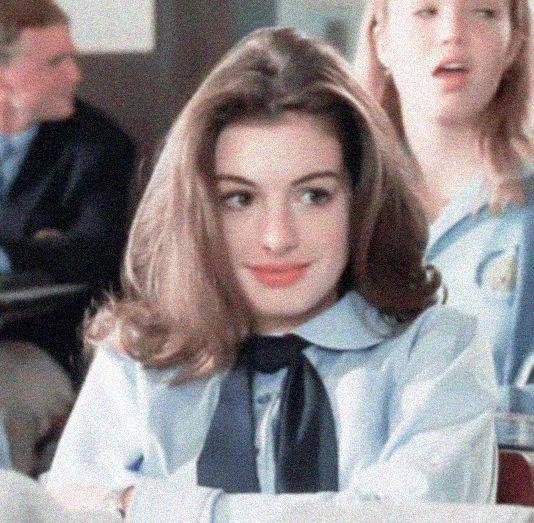}
            &
             \includegraphics[width=0.15\linewidth]{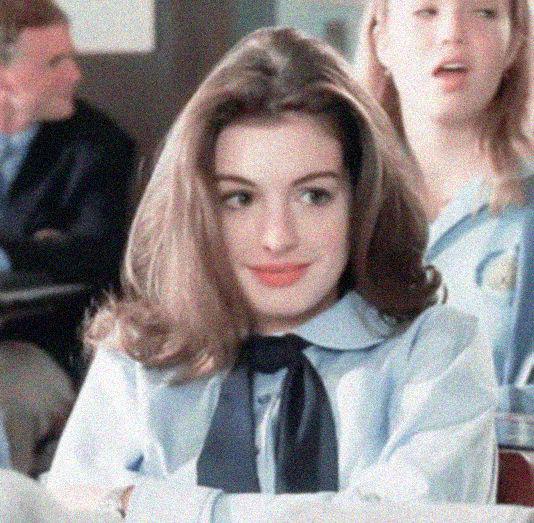}
            &
             \includegraphics[width=0.15\linewidth]{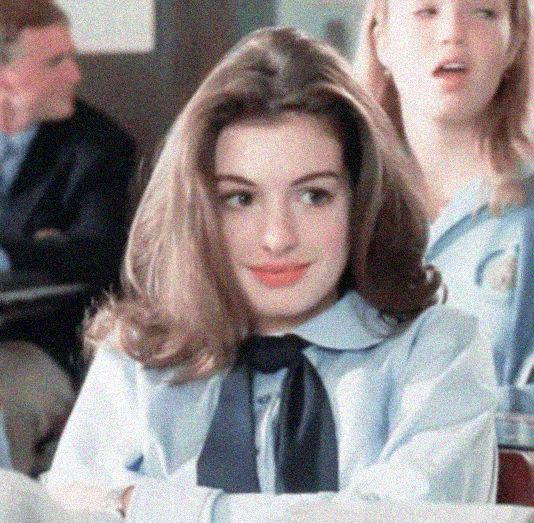}
            &
             \includegraphics[width=0.15\linewidth]{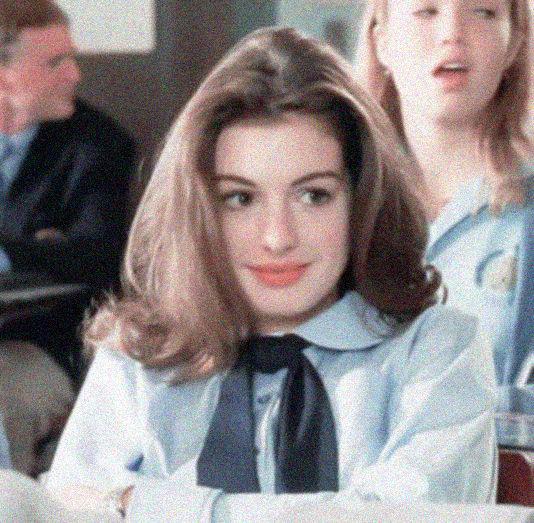}
             \vspace{-0.1cm} \\
             
            &
             \begin{tabular}{l}
             \vspace{-2.8cm} \\  \footnotesize  (0.2,0,0) 
             \end{tabular}
            &
             \includegraphics[width=0.15\linewidth]{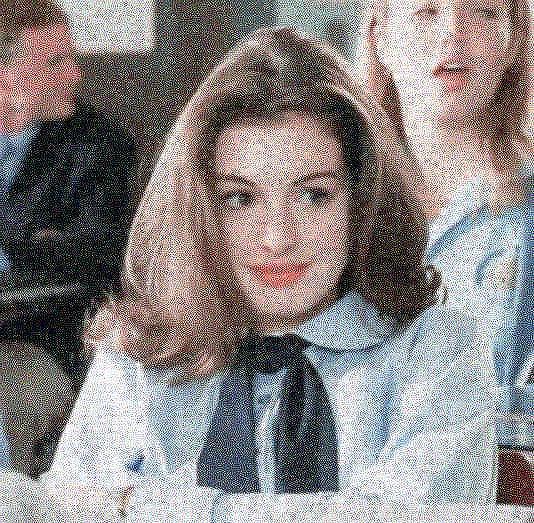}
            &
             \includegraphics[width=0.15\linewidth]{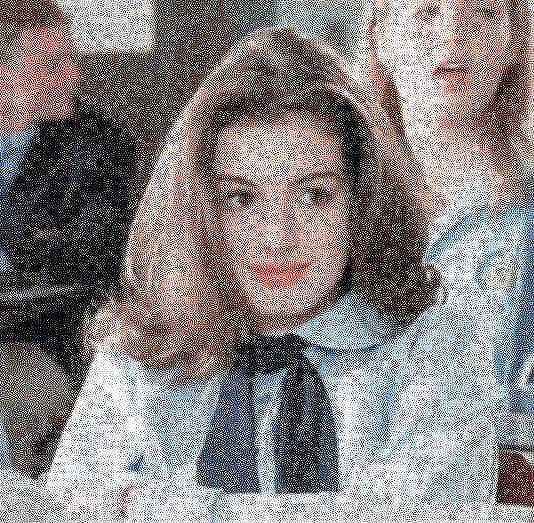}
            &
             \includegraphics[width=0.15\linewidth]{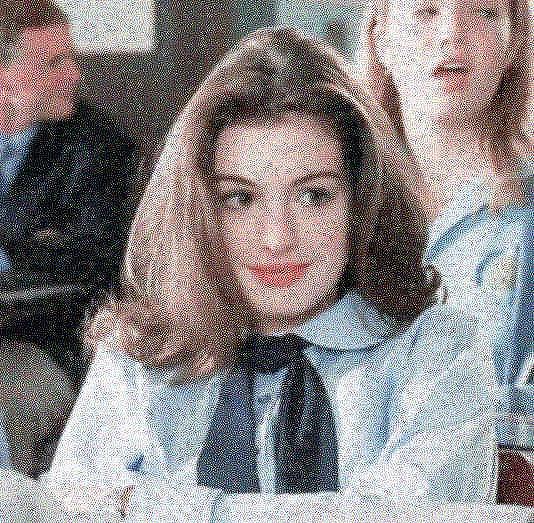}
            &
             \includegraphics[width=0.15\linewidth]{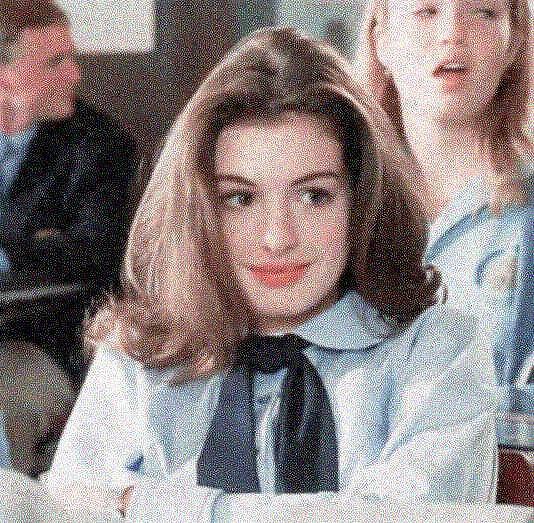}
            &
             \includegraphics[width=0.15\linewidth]{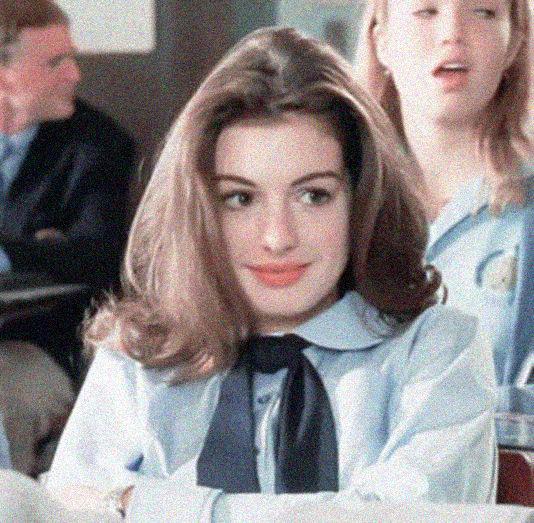}
             \vspace{-0.1cm} \\
            &
             \begin{tabular}{l}
             \vspace{-2.8cm} \\  \footnotesize  (0.4,0,0) 
             \end{tabular}
            &
             \includegraphics[width=0.15\linewidth]{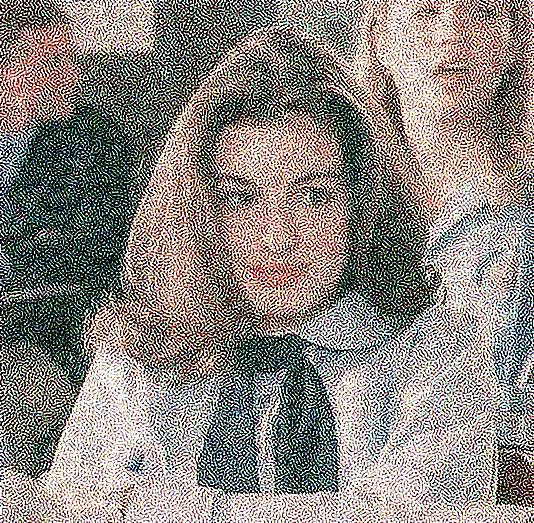}
            &
             \includegraphics[width=0.15\linewidth]{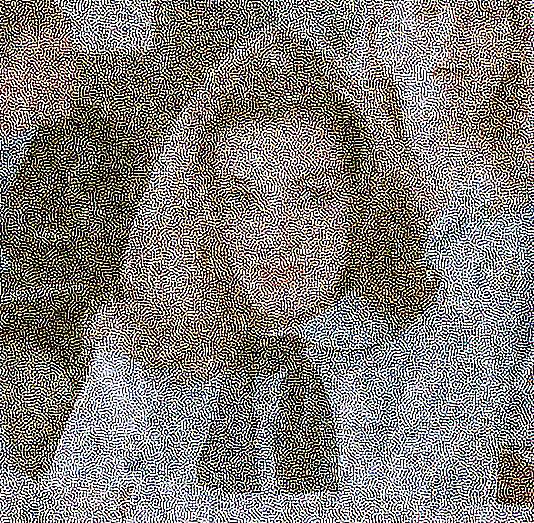}
            &
             \includegraphics[width=0.15\linewidth]{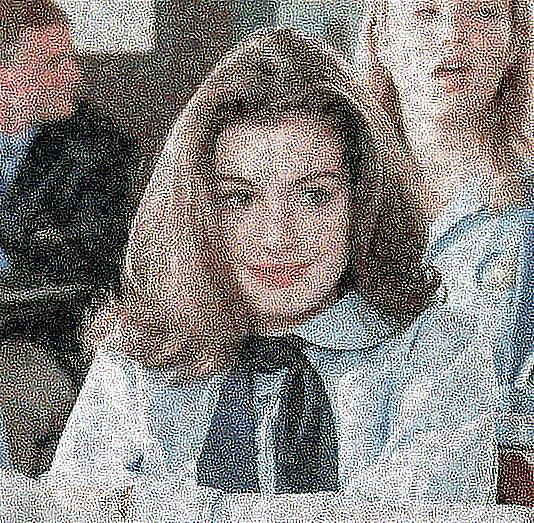}
            &
             \includegraphics[width=0.15\linewidth]{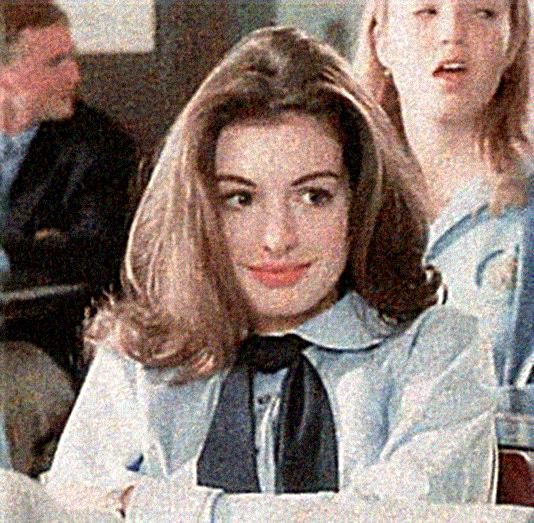}
            &
             \includegraphics[width=0.15\linewidth]{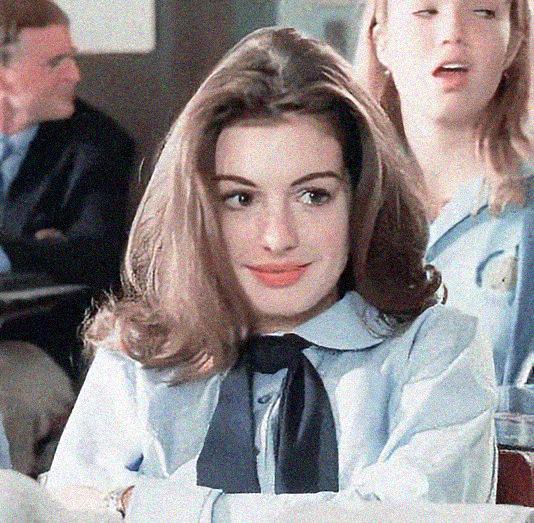}
             \vspace{-0.1cm} \\
             \begin{tabular}{c}
             \vspace{-2.8cm}\\
             \includegraphics[width=0.15\linewidth]{graphics/anne_able/anne_able.jpg}\\
             \vspace{-1cm} \\ Synthetic 
             \end{tabular}
            &
             \begin{tabular}{l}
             \vspace{-2.8cm} \\  \footnotesize  (0.5,0,0) 
             \end{tabular}
            &
             \includegraphics[width=0.15\linewidth]{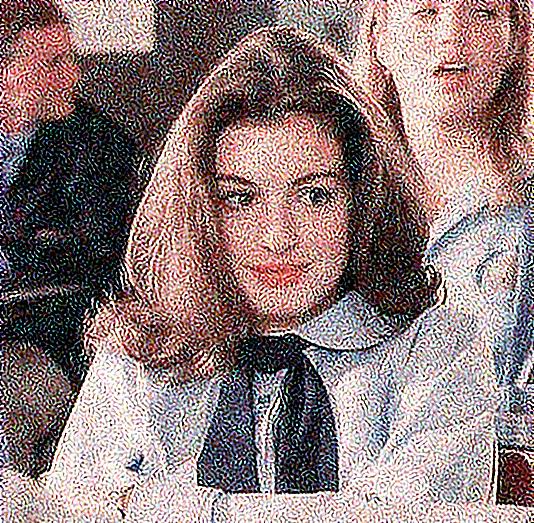}
            &
             \includegraphics[width=0.15\linewidth]{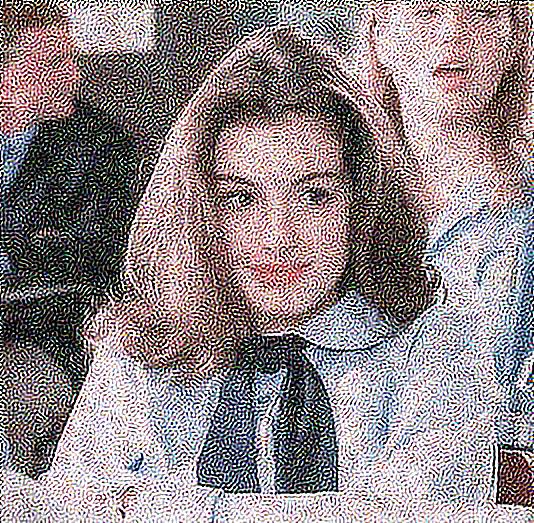}
            &
             \includegraphics[width=0.15\linewidth]{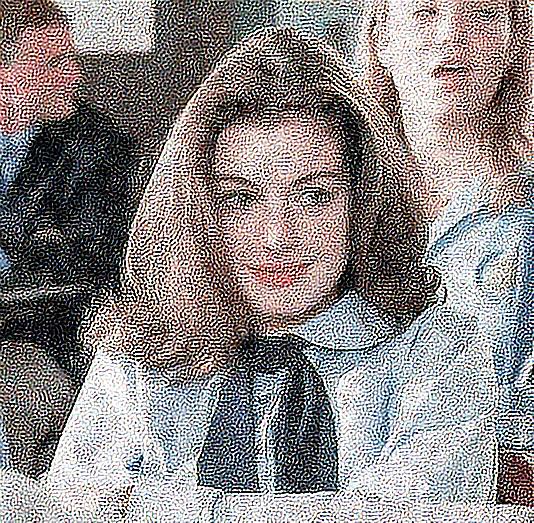}
            &
             \includegraphics[width=0.15\linewidth]{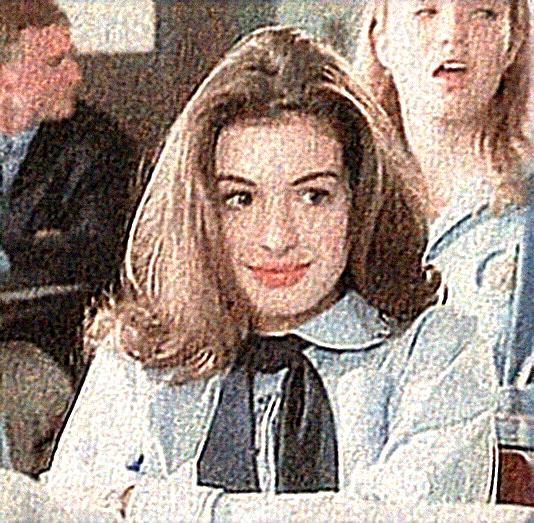}
            &
             \includegraphics[width=0.15\linewidth]{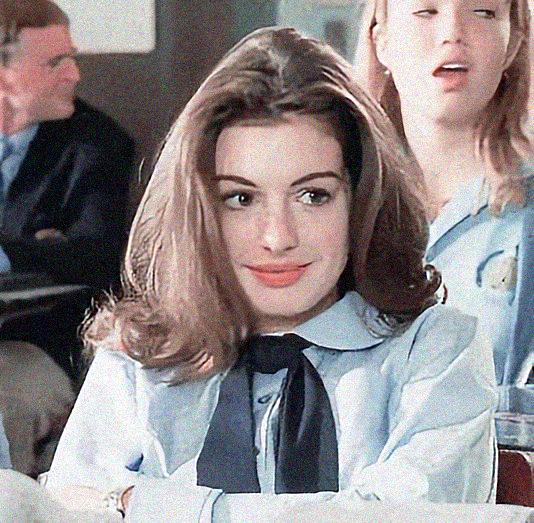}
             \vspace{-0.1cm} \\
             \begin{tabular}{c}
             \vspace{-4.2cm} \\ \footnotesize Optimal task vector  \vspace{-0.1cm} \\ \footnotesize =  \vspace{-0.1cm}\\ \footnotesize (0.3,0.3,0.3) 
             \end{tabular}
             &
             \begin{tabular}{l}
             \vspace{-2.8cm} \\  \footnotesize  (0.6,0,0) 
             \end{tabular}
            &
             \includegraphics[width=0.15\linewidth]{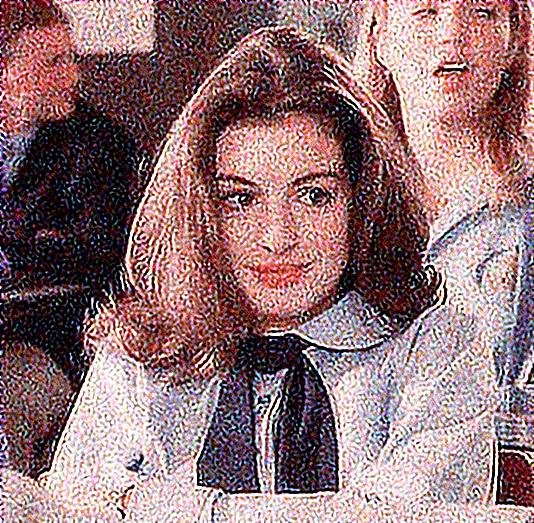}
            &
             \includegraphics[width=0.15\linewidth]{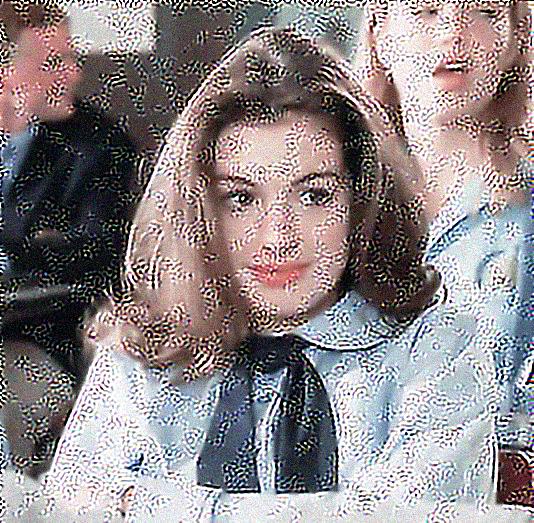}
            &
             \includegraphics[width=0.15\linewidth]{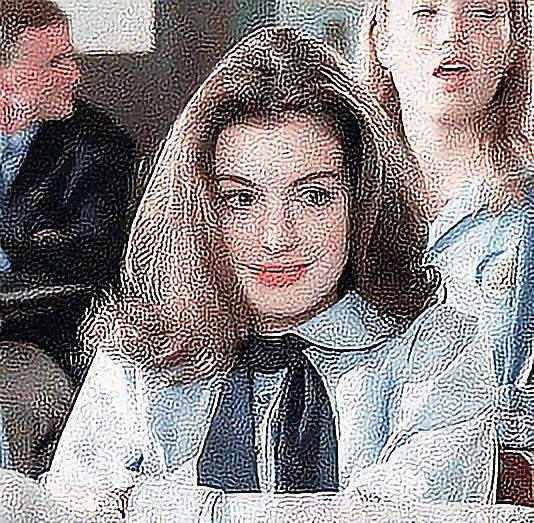}
            &
             \includegraphics[width=0.15\linewidth]{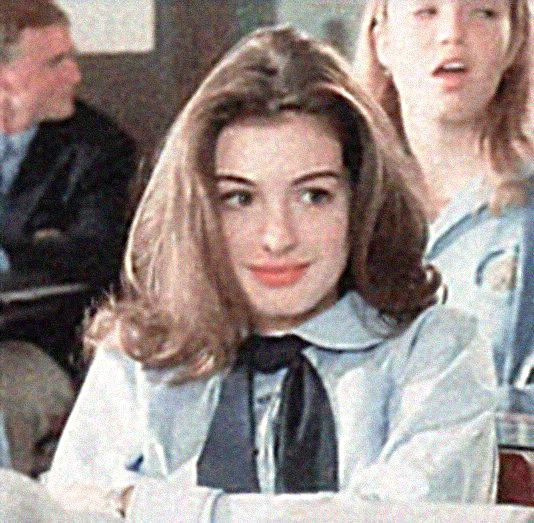}
            &
             \includegraphics[width=0.15\linewidth]{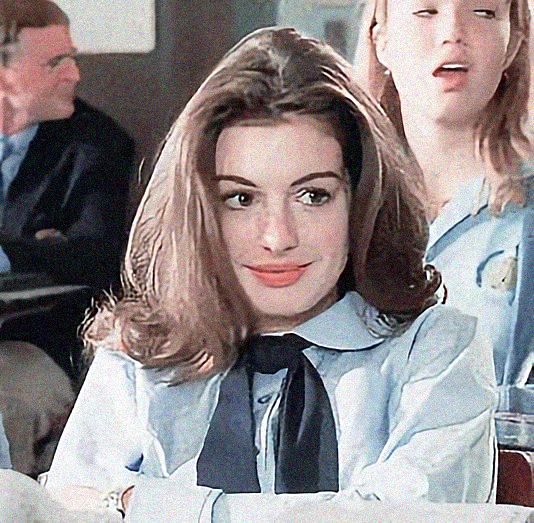}
             \vspace{-0.1cm} \\
            &
             \begin{tabular}{l}
             \vspace{-2.8cm} \\  \footnotesize  (0.8,0,0) 
             \end{tabular}
            &
             \includegraphics[width=0.15\linewidth]{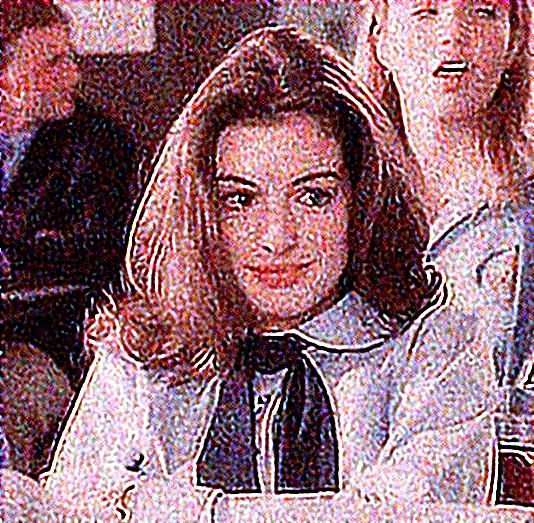}
            &
             \includegraphics[width=0.15\linewidth]{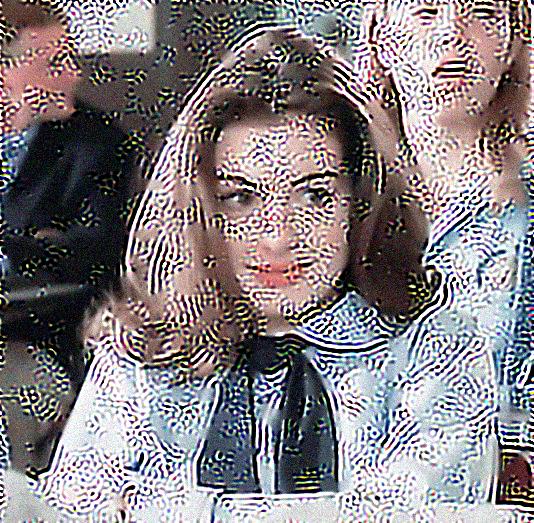}
            &
             \includegraphics[width=0.15\linewidth]{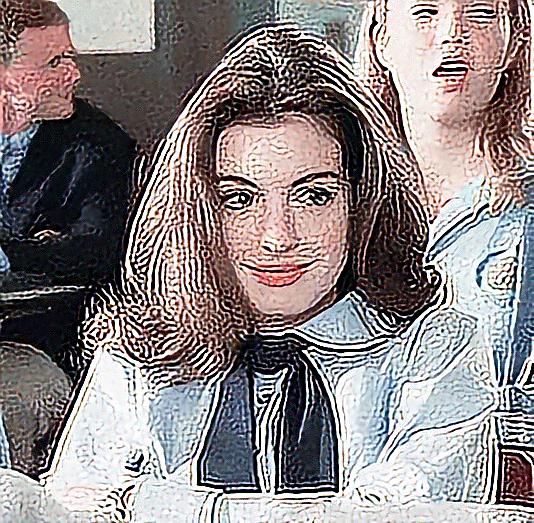}
            &
             \includegraphics[width=0.15\linewidth]{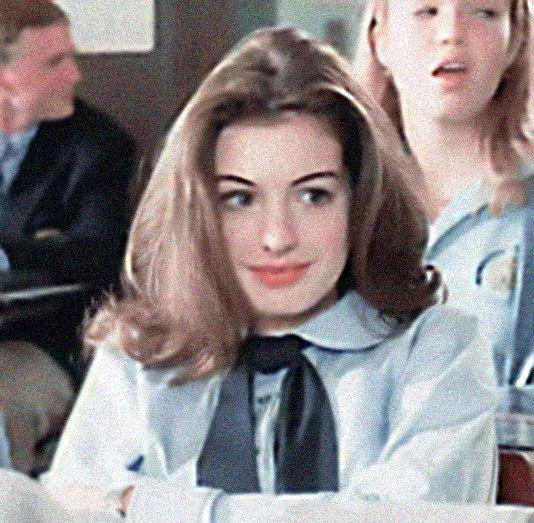}
            &
             \includegraphics[width=0.15\linewidth]{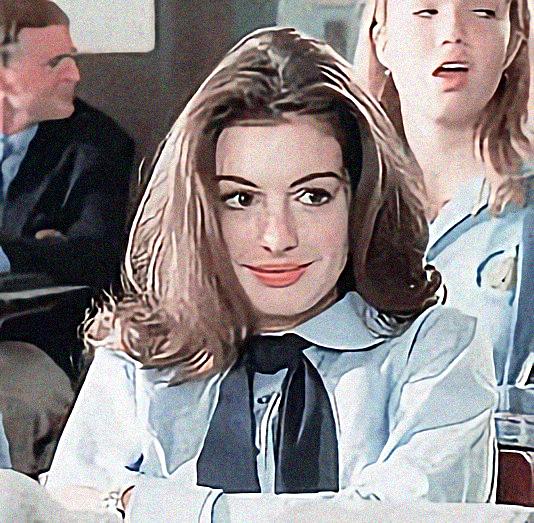}
             \vspace{-0.1cm} \\
            &
             \begin{tabular}{l}
             \vspace{-2.8cm} \\  \footnotesize  (1,0,0) 
             \end{tabular}
            &
             \includegraphics[width=0.15\linewidth]{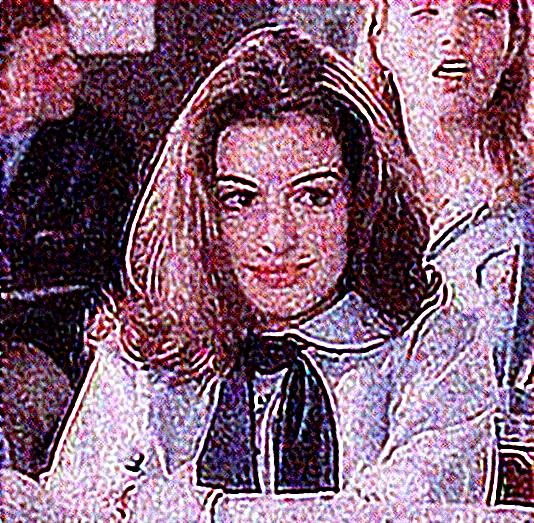}
            &
             \includegraphics[width=0.15\linewidth]{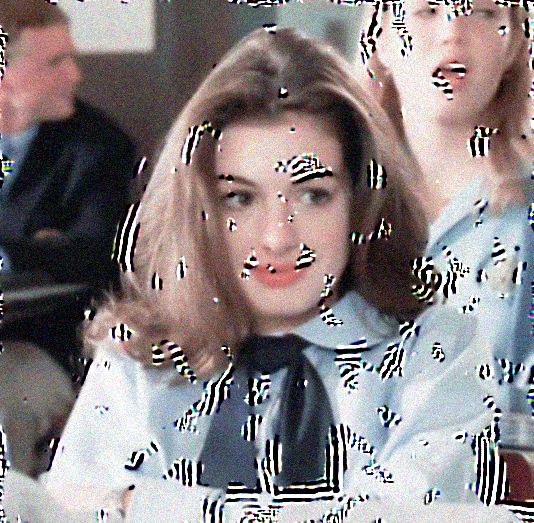}
            &
             \includegraphics[width=0.15\linewidth]{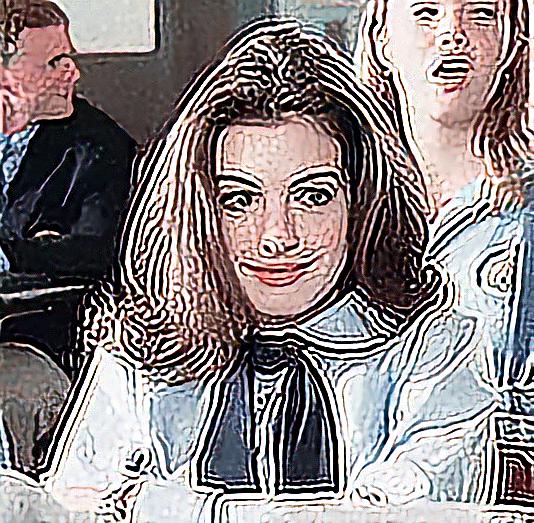}
            &
             \includegraphics[width=0.15\linewidth]{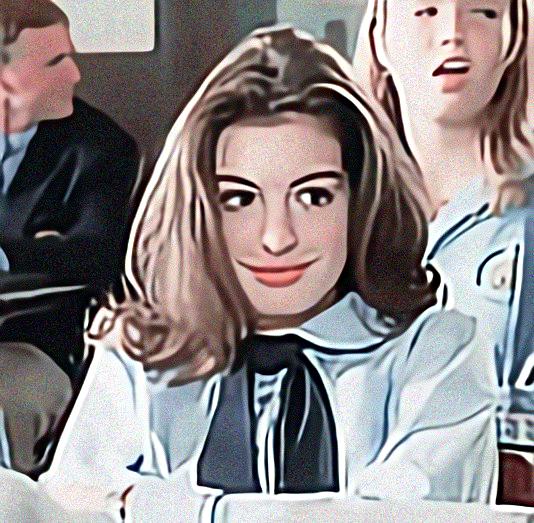}
            &
             \includegraphics[width=0.15\linewidth]{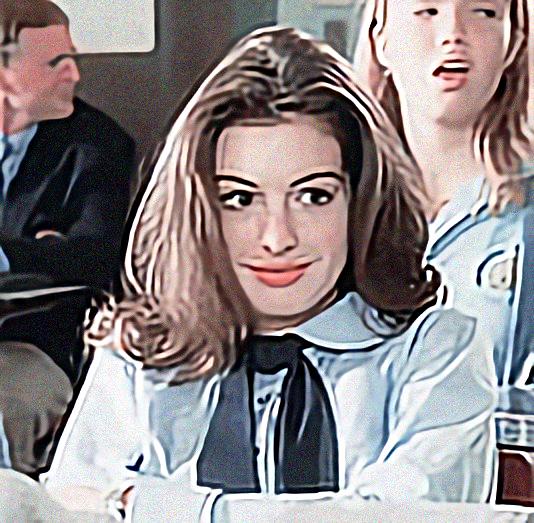}
             \vspace{-0.1cm} \\

	\toprule[1.5pt]
        \end{tabular}%\vspace{-0.3cm}
        \caption{\textbf{Deblur modulation examples to the image with blur, noise, and jpeg compression}. Our TA+TSNet-m modulates diverse imagery effects with respect to the given restoration tasks. It generates less auxiliary visual artifacts. The values of task vector denote restoration levels of (deblur, denoise, dejpeg), respectively.}
        \label{fig:synthetic_deblur}
    \end{center}
	\vspace{-0.2cm}
\end{figure*}

\begin{figure*}[p]
        \begin{center}\centering
        \setlength{\tabcolsep}{0.02cm}
        \setlength{\columnwidth}{2.81cm}
        \hspace*{-\tabcolsep}\begin{tabular}{ccccccc}
            \multicolumn{1}{c}{\footnotesize Input}    
            &
            \footnotesize Task vector
            &
            \multicolumn{1}{c}{\footnotesize CResMD~\cite{jingwen2020interactive}}
            &
            \multicolumn{1}{c}{ \footnotesize TSNet }
            &
            \multicolumn{1}{c}{\footnotesize TA+TSNet }
            &
            \multicolumn{1}{c}{\footnotesize TSNet-m }
            &
            \multicolumn{1}{c}{\footnotesize TA+TSNet-m }
            \\ \toprule[1.5pt]
            %\vspace{-0.1cm} \\
            &
             \begin{tabular}{c}
             \vspace{-2.8cm} \\  \footnotesize  (0,0,0) 
             \end{tabular}
            &
             \includegraphics[width=0.15\linewidth]{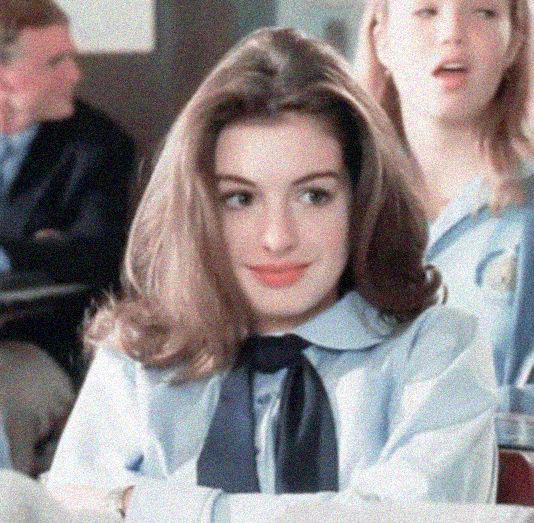}
            &
             \includegraphics[width=0.15\linewidth]{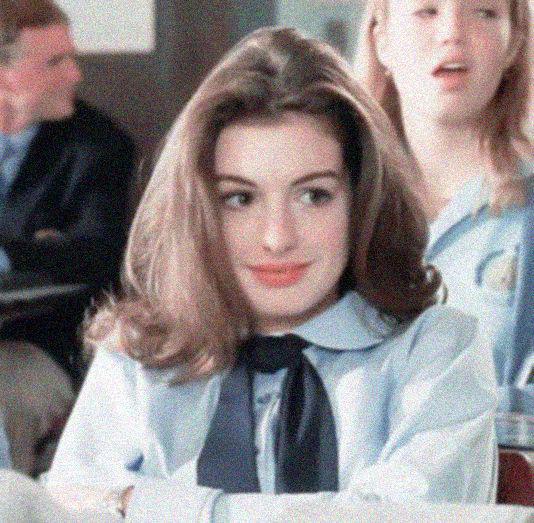}
            &
             \includegraphics[width=0.15\linewidth]{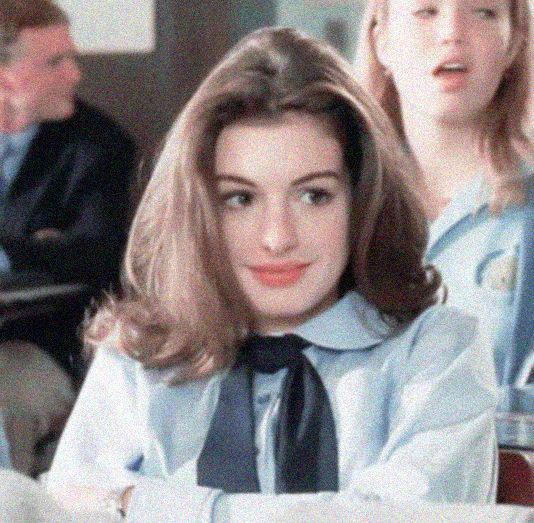}
            &
             \includegraphics[width=0.15\linewidth]{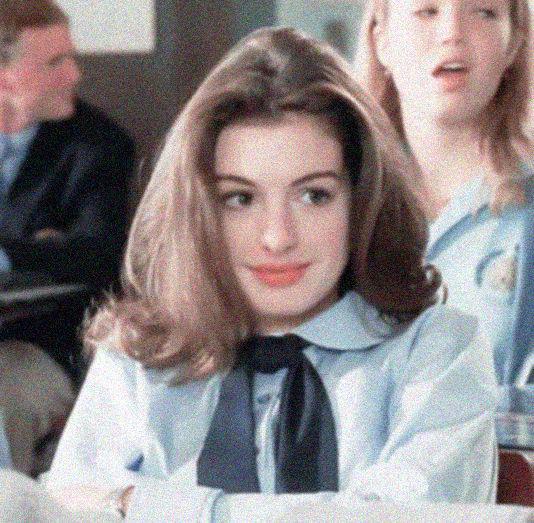}
            &
             \includegraphics[width=0.15\linewidth]{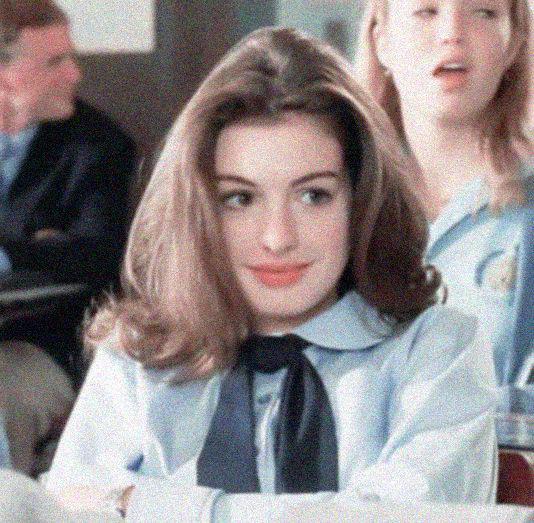}
             \vspace{-0.1cm} \\
             
            &
             \begin{tabular}{l}
             \vspace{-2.8cm} \\  \footnotesize  (0.2,0,0.2) 
             \end{tabular}
            &
             \includegraphics[width=0.15\linewidth]{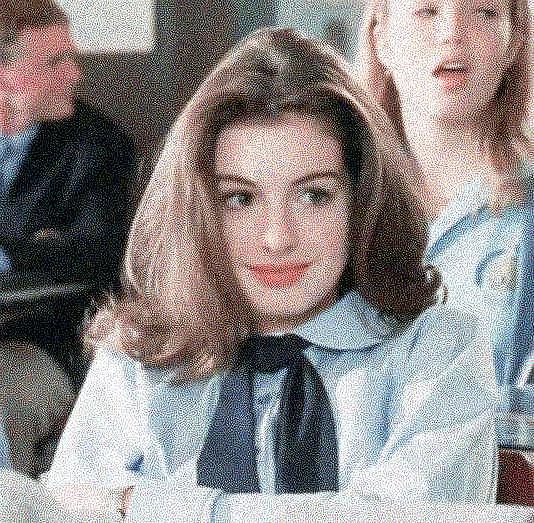}
            &
             \includegraphics[width=0.15\linewidth]{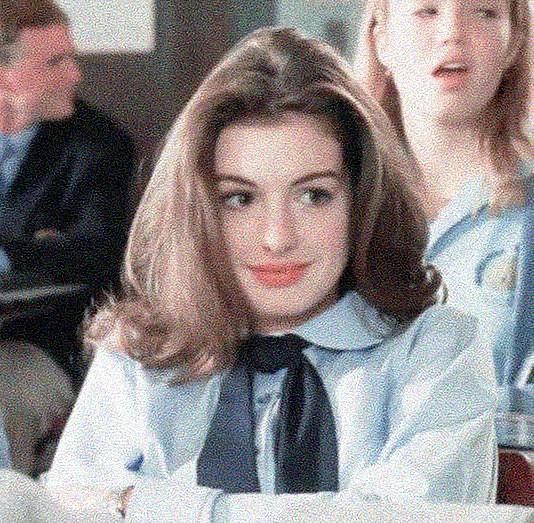}
            &
             \includegraphics[width=0.15\linewidth]{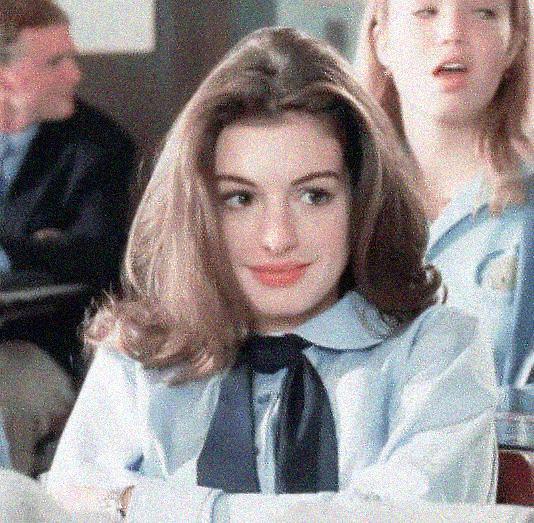}
            &
             \includegraphics[width=0.15\linewidth]{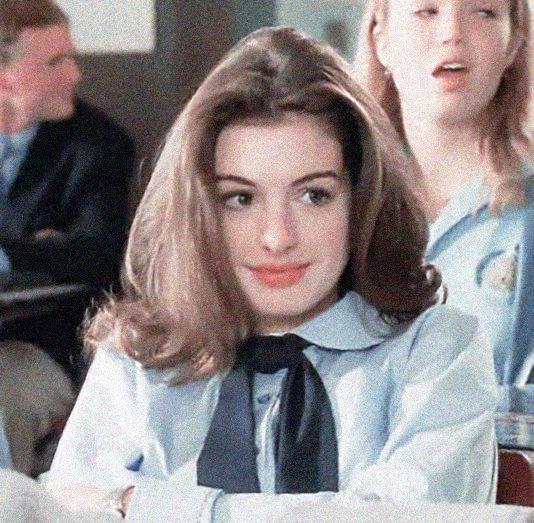}
            &
             \includegraphics[width=0.15\linewidth]{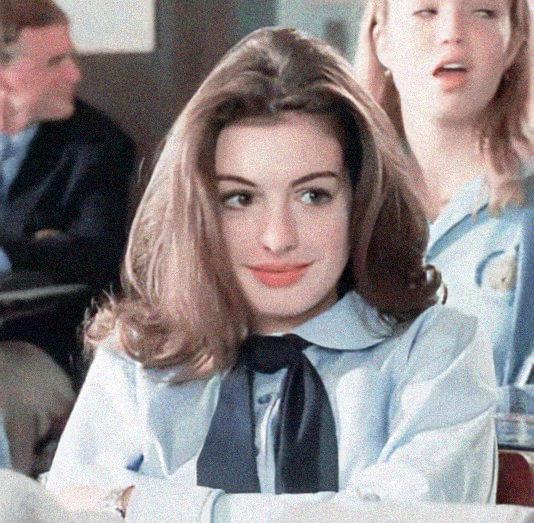}
             \vspace{-0.1cm} \\
            &
             \begin{tabular}{l}
             \vspace{-2.8cm} \\  \footnotesize  (0.4,0,0.4) 
             \end{tabular}
            &
             \includegraphics[width=0.15\linewidth]{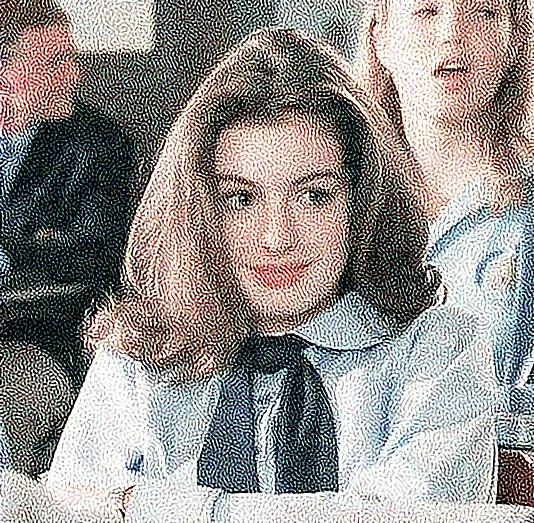}
            &
             \includegraphics[width=0.15\linewidth]{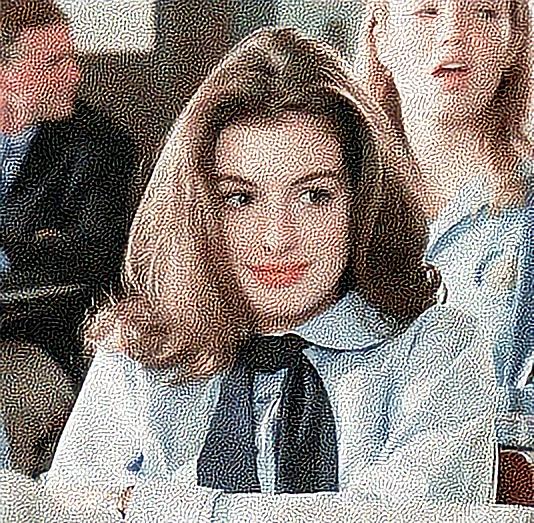}
            &
             \includegraphics[width=0.15\linewidth]{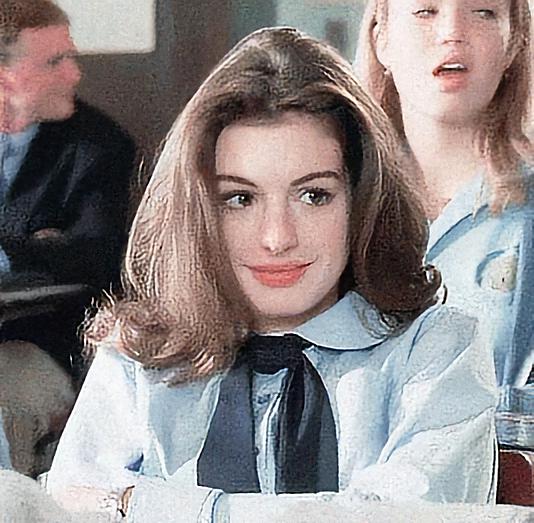}
            &
             \includegraphics[width=0.15\linewidth]{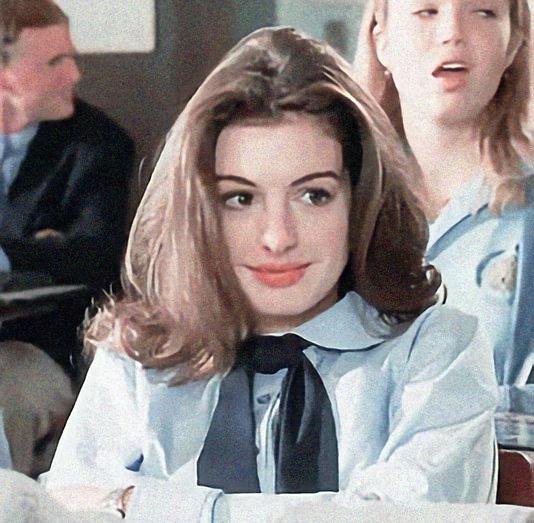}
            &
             \includegraphics[width=0.15\linewidth]{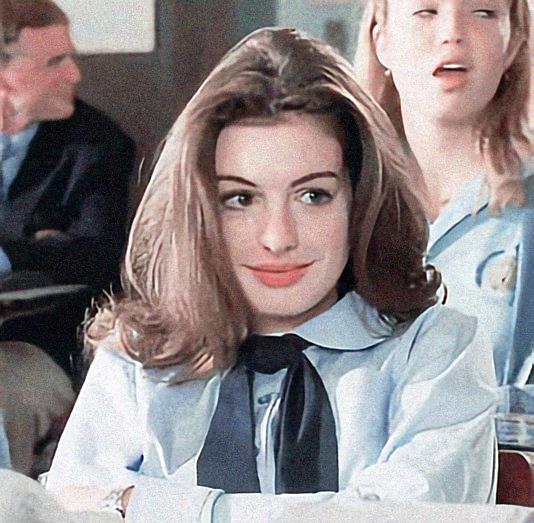}
             \vspace{-0.1cm} \\
             \begin{tabular}{c}
             \vspace{-2.8cm}\\
             \includegraphics[width=0.15\linewidth]{graphics/anne_able/anne_able.jpg}\\
             \vspace{-1cm} \\ Synthetic 
             \end{tabular}
            &
             \begin{tabular}{l}
             \vspace{-2.8cm} \\  \footnotesize  (0.5,0,0.5) 
             \end{tabular}
            &
             \includegraphics[width=0.15\linewidth]{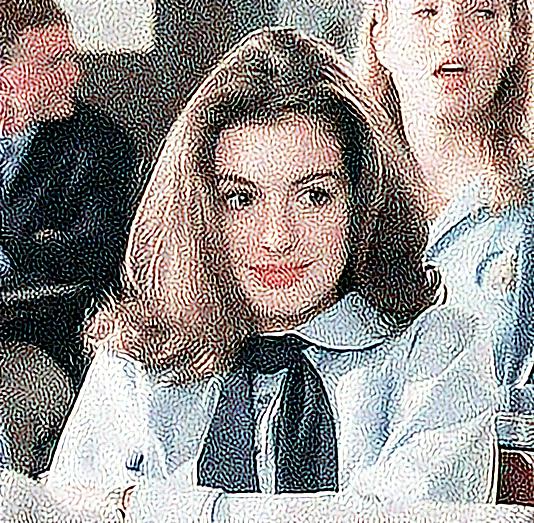}
            &
             \includegraphics[width=0.15\linewidth]{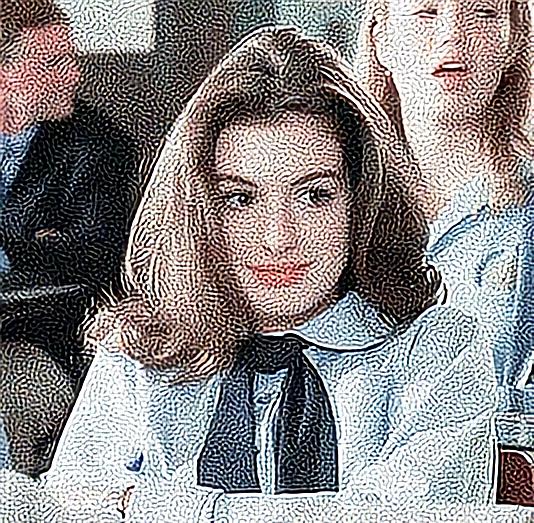}
            &
             \includegraphics[width=0.15\linewidth]{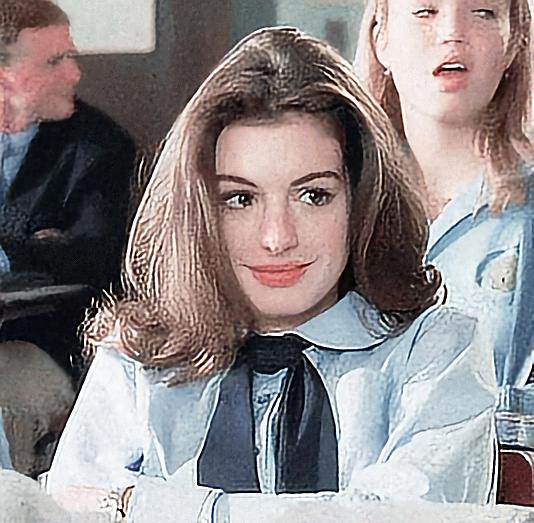}
            &
             \includegraphics[width=0.15\linewidth]{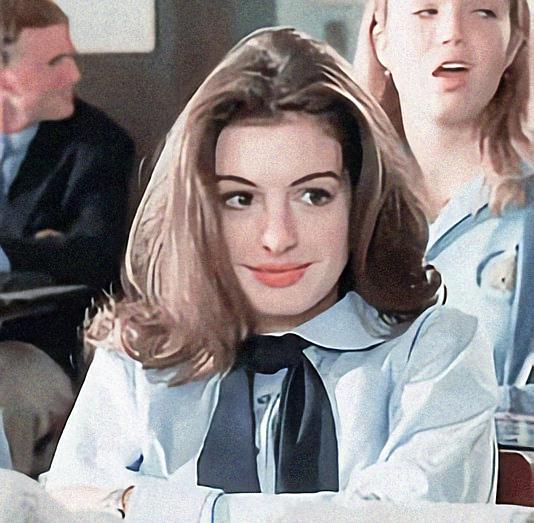}
            &
             \includegraphics[width=0.15\linewidth]{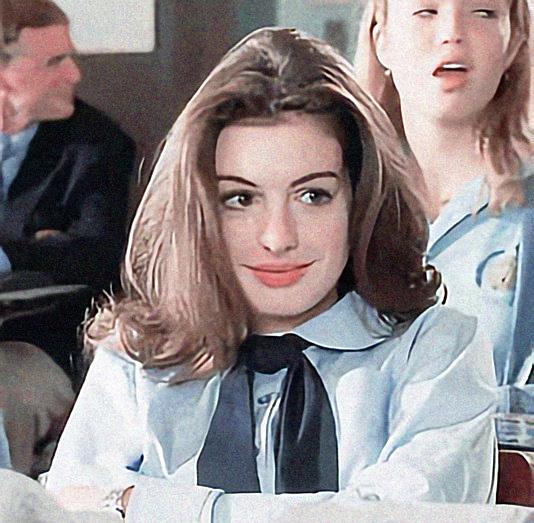}
             \vspace{-0.1cm} \\
             \begin{tabular}{c}
             \vspace{-4.2cm} \\ \footnotesize Optimal task vector  \vspace{-0.1cm} \\ \footnotesize =  \vspace{-0.1cm}\\ \footnotesize (0.3,0.3,0.3) 
             \end{tabular}
             &
             \begin{tabular}{l}
             \vspace{-2.8cm} \\  \footnotesize  (0.6,0,0.6) 
             \end{tabular}
            &
             \includegraphics[width=0.15\linewidth]{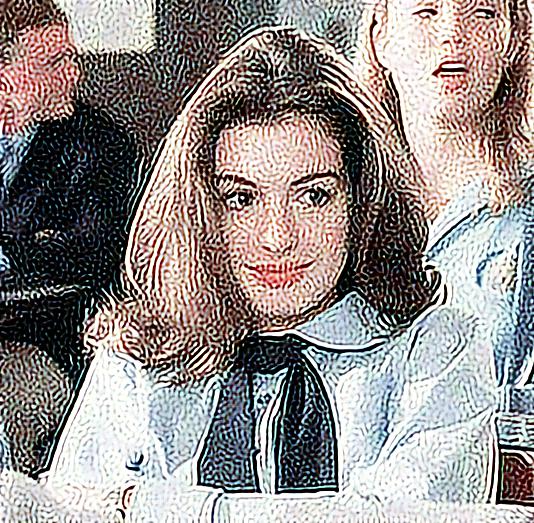}
            &
             \includegraphics[width=0.15\linewidth]{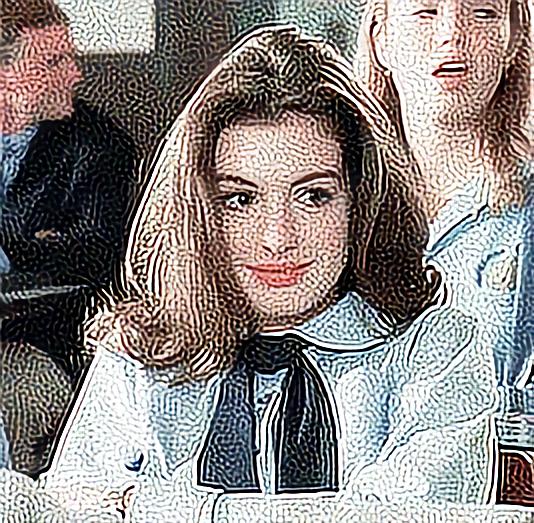}
            &
             \includegraphics[width=0.15\linewidth]{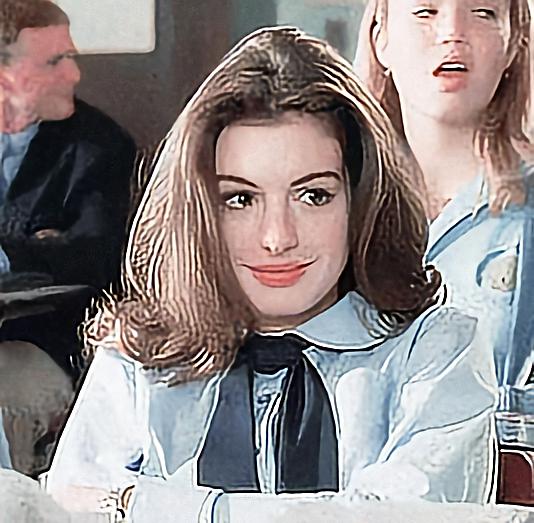}
            &
             \includegraphics[width=0.15\linewidth]{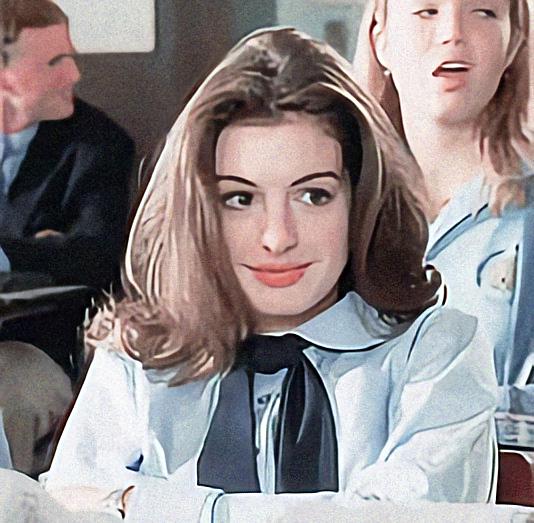}
            &
             \includegraphics[width=0.15\linewidth]{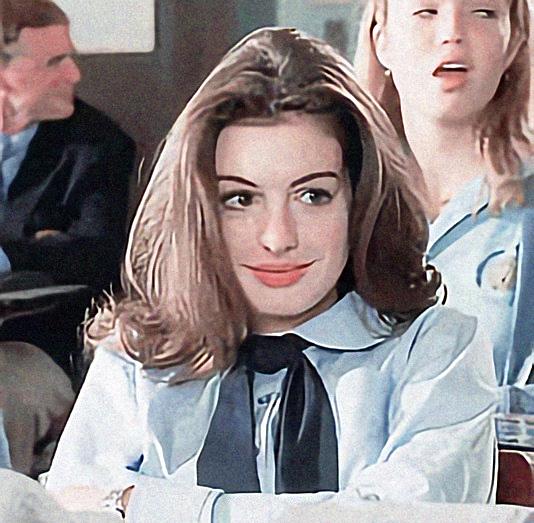}
             \vspace{-0.1cm} \\
            &
             \begin{tabular}{l}
             \vspace{-2.8cm} \\  \footnotesize  (0.8,0,0.8) 
             \end{tabular}
            &
             \includegraphics[width=0.15\linewidth]{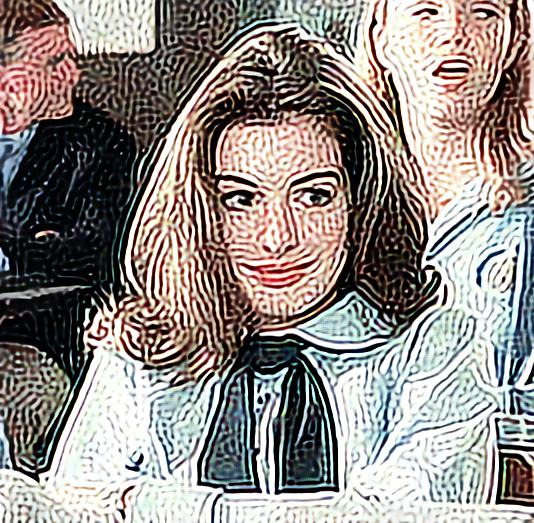}
            &
             \includegraphics[width=0.15\linewidth]{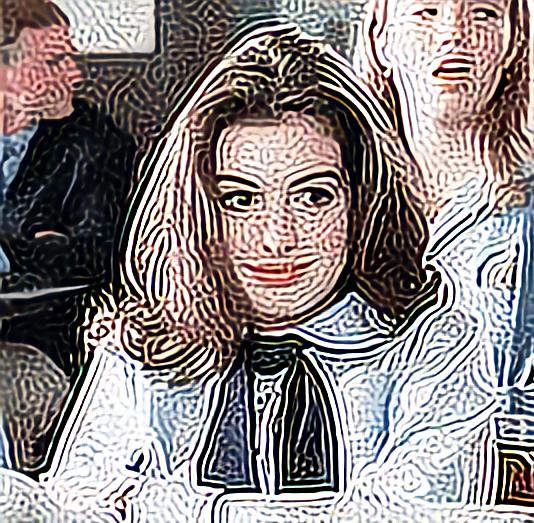}
            &
             \includegraphics[width=0.15\linewidth]{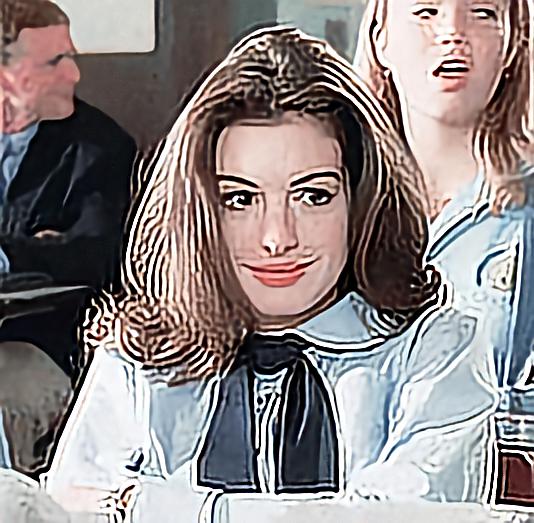}
            &
             \includegraphics[width=0.15\linewidth]{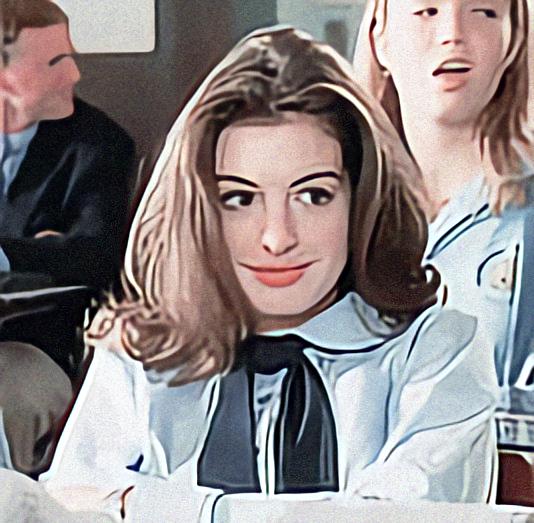}
            &
             \includegraphics[width=0.15\linewidth]{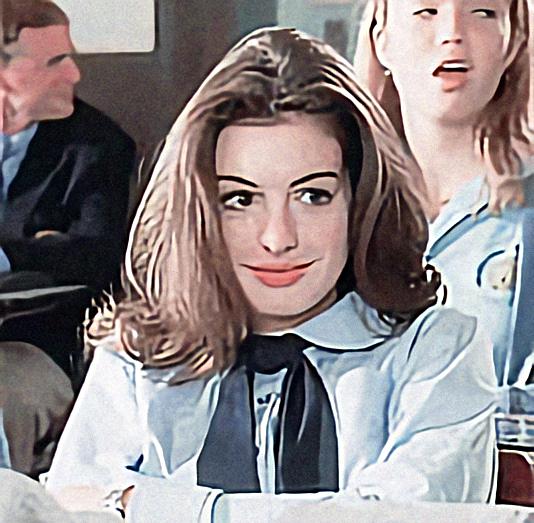}
             \vspace{-0.1cm} \\
            &
             \begin{tabular}{l}
             \vspace{-2.8cm} \\  \footnotesize  (1,0,1) 
             \end{tabular}
            &
             \includegraphics[width=0.15\linewidth]{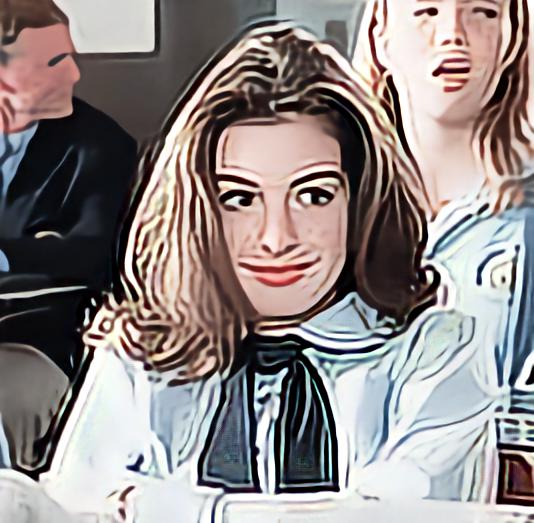}
            &
             \includegraphics[width=0.15\linewidth]{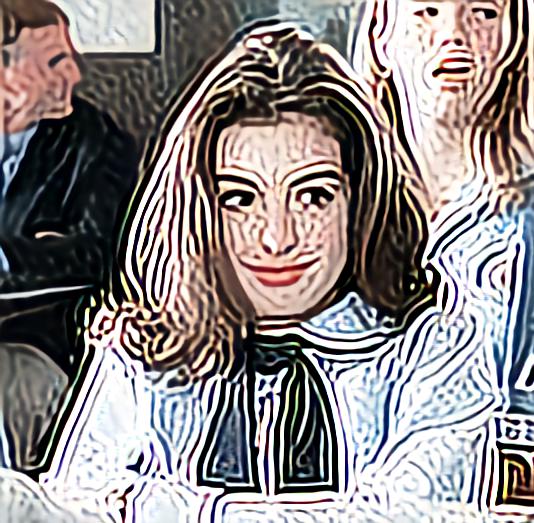}
            &
             \includegraphics[width=0.15\linewidth]{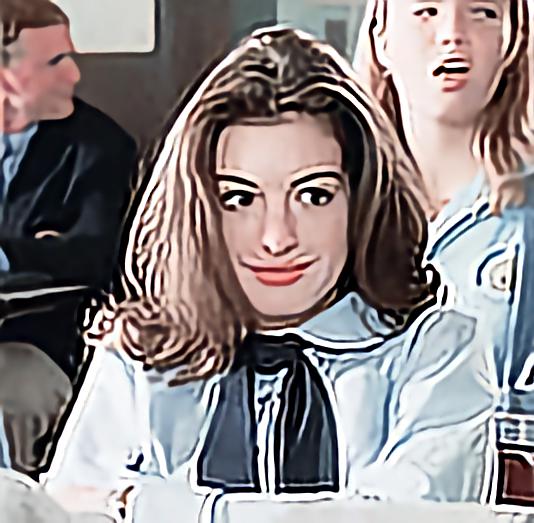}
            &
             \includegraphics[width=0.15\linewidth]{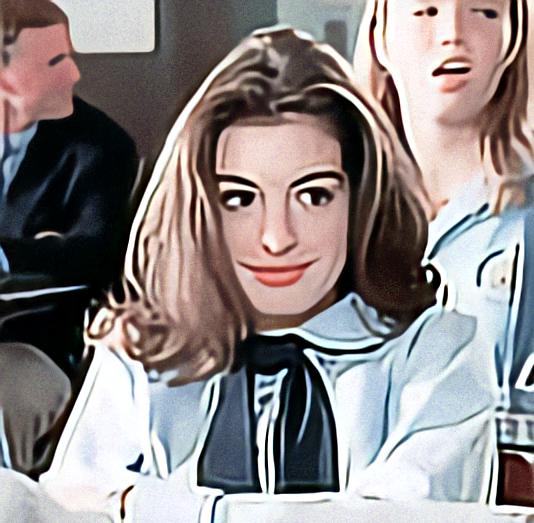}
            &
             \includegraphics[width=0.15\linewidth]{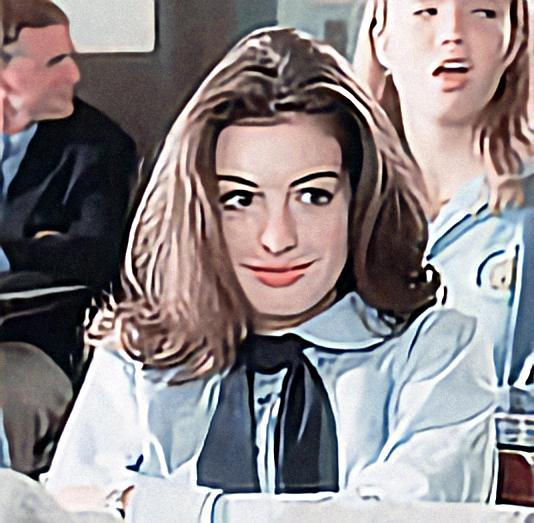}
             \vspace{-0.1cm} \\

	\toprule[1.5pt]
        \end{tabular}%\vspace{-0.3cm}
        \caption{\textbf{Deblur and dejpeg modulation examples to the image with blur, noise, and jpeg compression}. Our TA+TSNet-m modulates diverse imagery effects with respect to the given restoration tasks. It generates less auxiliary visual artifacts. The values of task vector denote restoration levels of (deblur, denoise, dejpeg), respectively.}\vspace{-0.0cm}
        \label{fig:synthetic_deblur_dejpeg}
    \end{center}
	\vspace{-0.2cm}
\end{figure*}

\begin{figure*}[p]
        \begin{center}\centering
        \setlength{\tabcolsep}{0.02cm}
        \setlength{\columnwidth}{2.81cm}
        \hspace*{-\tabcolsep}\begin{tabular}{ccccccc}
            \multicolumn{1}{c}{\footnotesize Input}    
            &
            \footnotesize Task vector
            &
            \multicolumn{1}{c}{\footnotesize CResMD~\cite{jingwen2020interactive}}
            &
            \multicolumn{1}{c}{ \footnotesize TSNet }
            &
            \multicolumn{1}{c}{\footnotesize TA+TSNet }
            &
            \multicolumn{1}{c}{\footnotesize TSNet-m }
            &
            \multicolumn{1}{c}{\footnotesize TA+TSNet-m }
            \\ \toprule[1.5pt]
            %\vspace{-0.1cm} \\
            &
             \begin{tabular}{c}
             \vspace{-2.8cm} \\  \footnotesize  (0,0,0) 
             \end{tabular}
            &
             \includegraphics[width=0.15\linewidth]{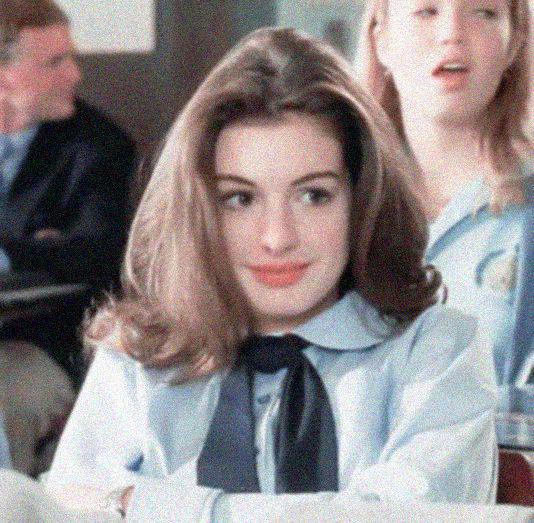}
            &
             \includegraphics[width=0.15\linewidth]{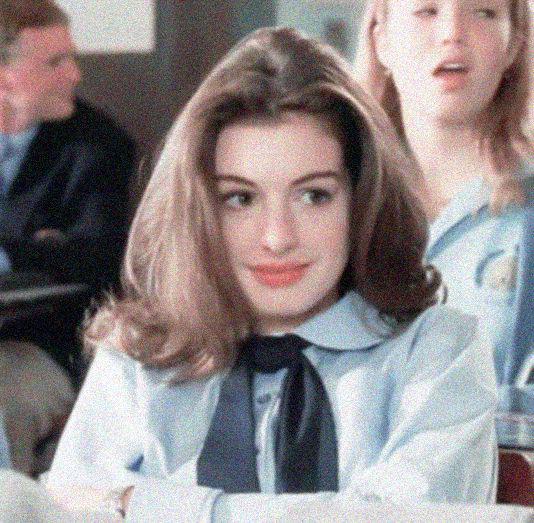}
            &
             \includegraphics[width=0.15\linewidth]{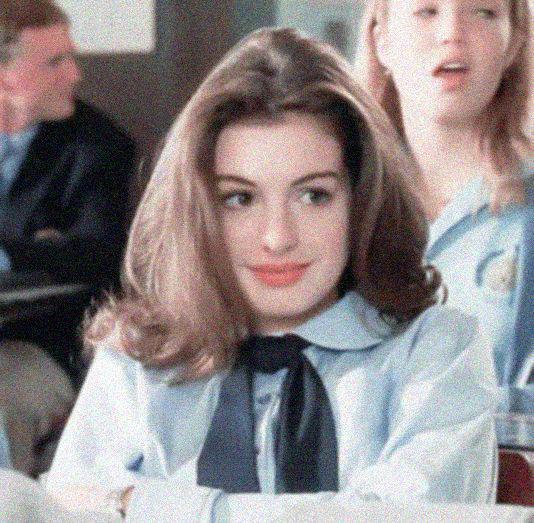}
            &
             \includegraphics[width=0.15\linewidth]{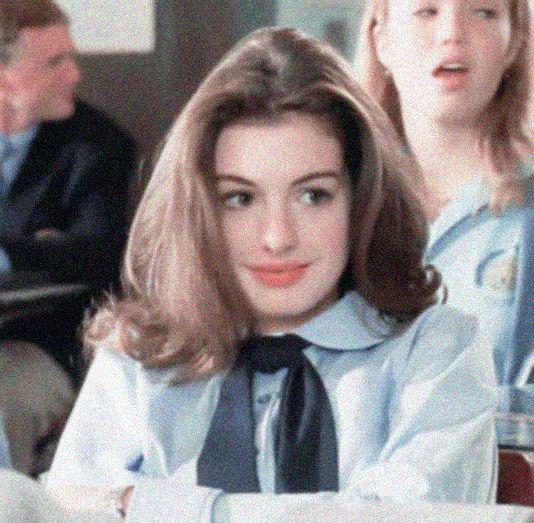}
            &
             \includegraphics[width=0.15\linewidth]{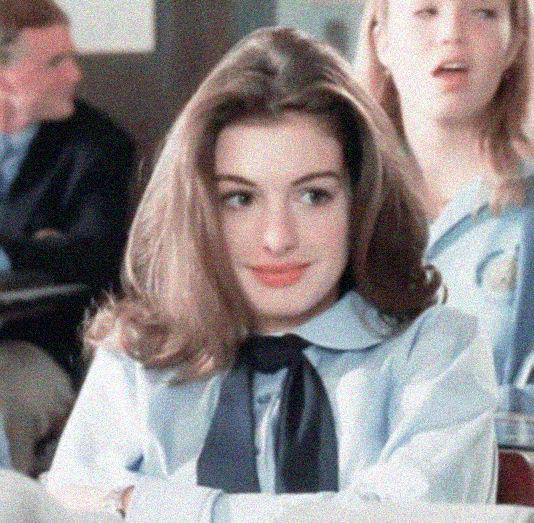}
             \vspace{-0.1cm} \\
             
            &
             \begin{tabular}{l}
             \vspace{-2.8cm} \\  \footnotesize  (0.2,0.2,0.2) 
             \end{tabular}
            &
             \includegraphics[width=0.15\linewidth]{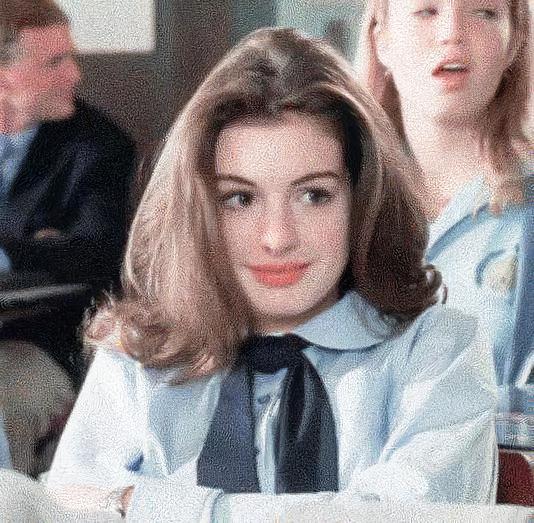}
            &
             \includegraphics[width=0.15\linewidth]{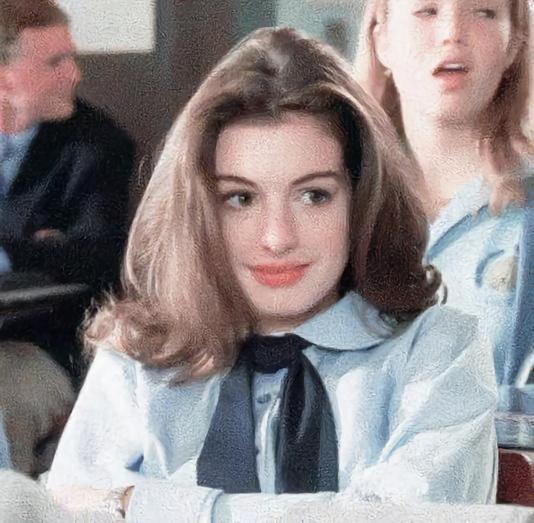}
            &
             \includegraphics[width=0.15\linewidth]{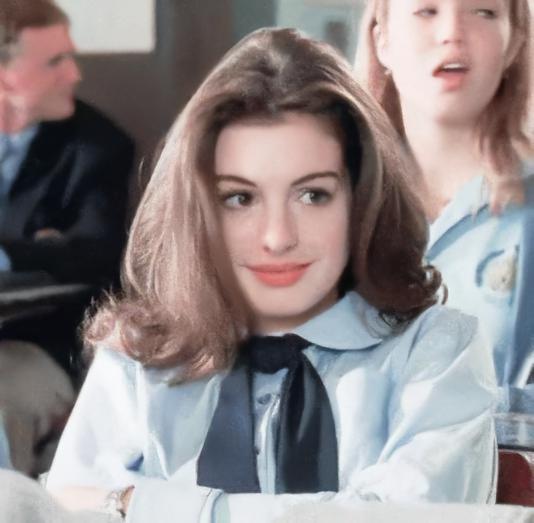}
            &
             \includegraphics[width=0.15\linewidth]{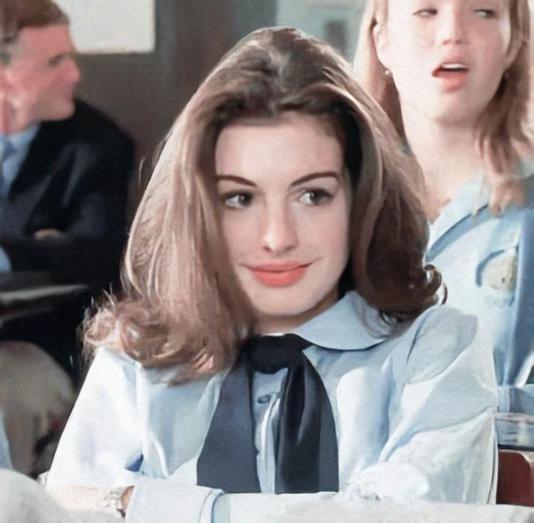}
            &
             \includegraphics[width=0.15\linewidth]{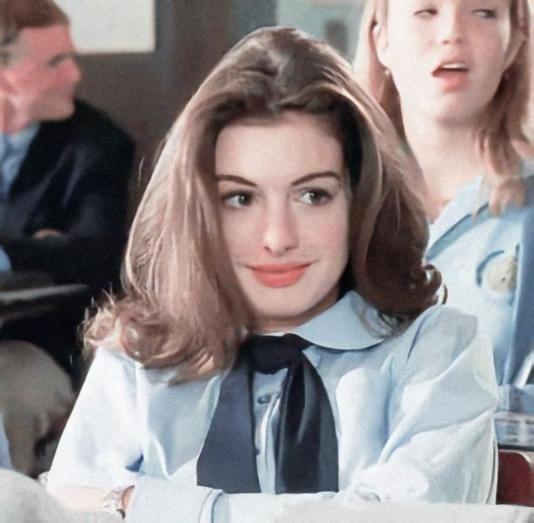}
             \vspace{-0.1cm} \\
            &
             \begin{tabular}{l}
             \vspace{-2.8cm} \\  \footnotesize  (0.4,0.4,0.4) 
             \end{tabular}
            &
             \includegraphics[width=0.15\linewidth]{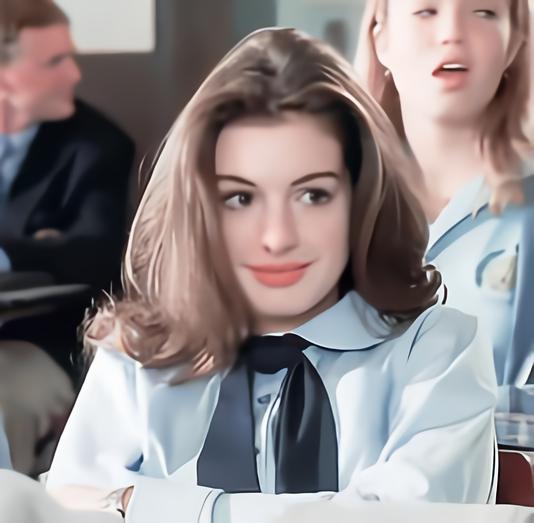}
            &
             \includegraphics[width=0.15\linewidth]{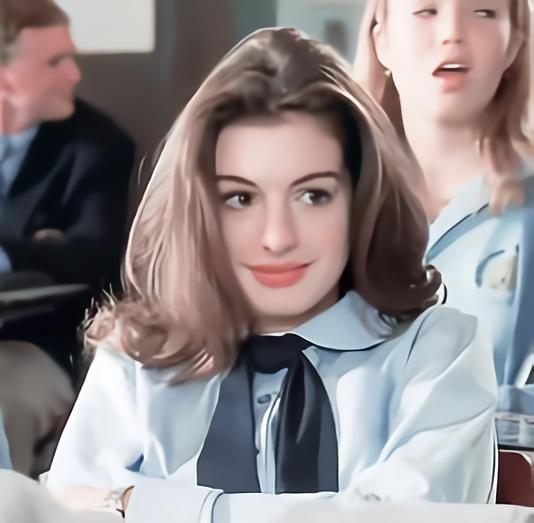}
            &
             \includegraphics[width=0.15\linewidth]{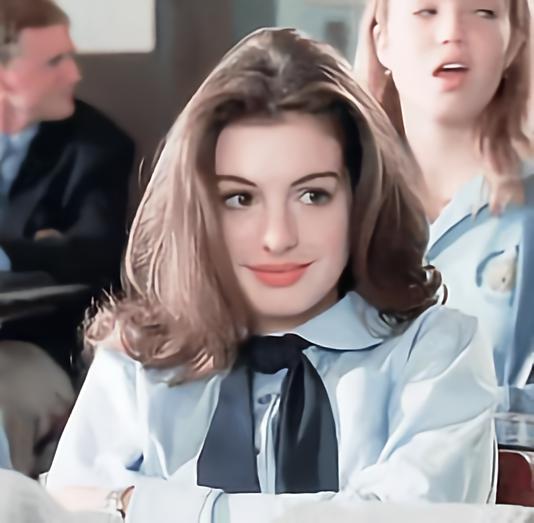}
            &
             \includegraphics[width=0.15\linewidth]{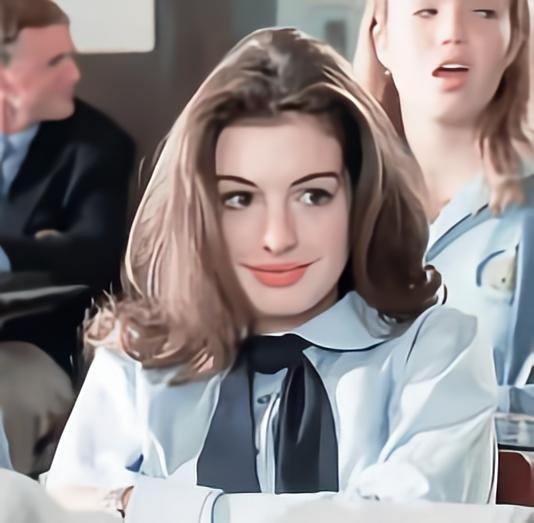}
            &
             \includegraphics[width=0.15\linewidth]{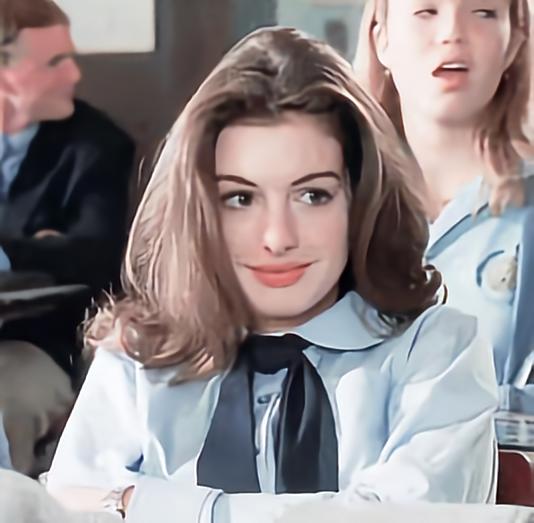}
             \vspace{-0.1cm} \\
             \begin{tabular}{c}
             \vspace{-2.8cm}\\
             \includegraphics[width=0.15\linewidth]{graphics/anne_able/anne_able.jpg}\\
             \vspace{-1cm} \\ Synthetic 
             \end{tabular}
            &
             \begin{tabular}{l}
             \vspace{-2.8cm} \\  \footnotesize  (0.5,0.5,0.5) 
             \end{tabular}
            &
             \includegraphics[width=0.15\linewidth]{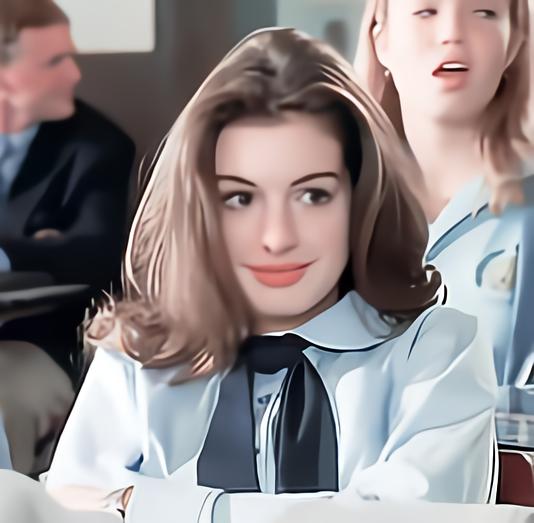}
            &
             \includegraphics[width=0.15\linewidth]{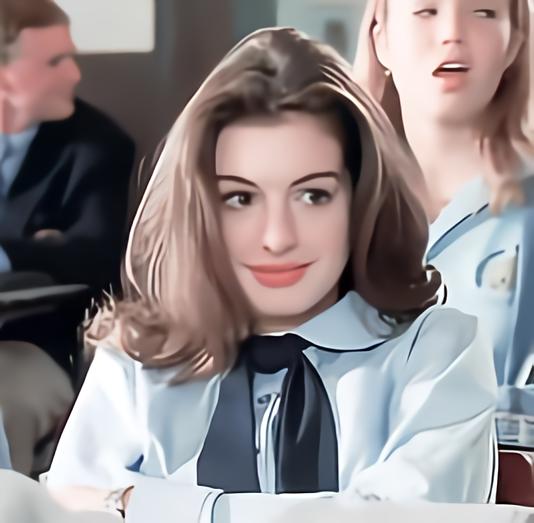}
            &
             \includegraphics[width=0.15\linewidth]{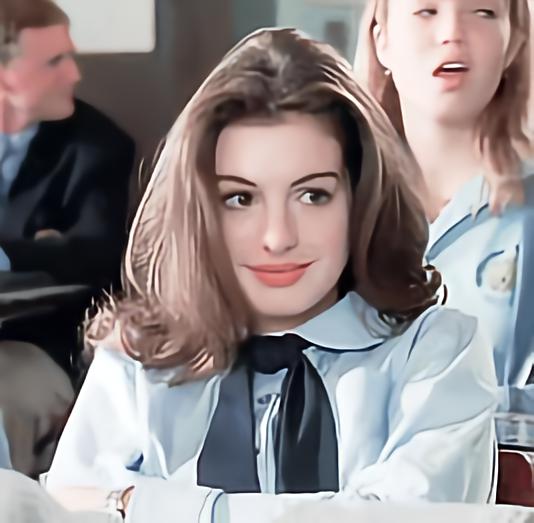}
            &
             \includegraphics[width=0.15\linewidth]{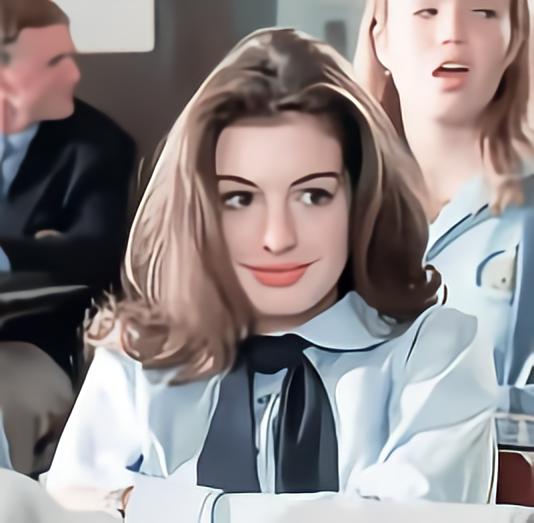}
            &
             \includegraphics[width=0.15\linewidth]{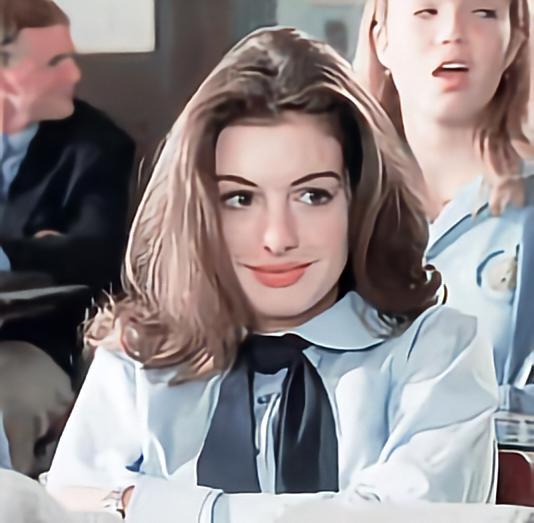}
             \vspace{-0.1cm} \\
             \begin{tabular}{c}
             \vspace{-4.2cm} \\ \footnotesize Optimal task vector  \vspace{-0.1cm} \\ \footnotesize =  \vspace{-0.1cm}\\ \footnotesize (0.3,0.3,0.3) 
             \end{tabular}
             &
             \begin{tabular}{l}
             \vspace{-2.8cm} \\  \footnotesize  (0.6,0.6,0.6) 
             \end{tabular}
            &
             \includegraphics[width=0.15\linewidth]{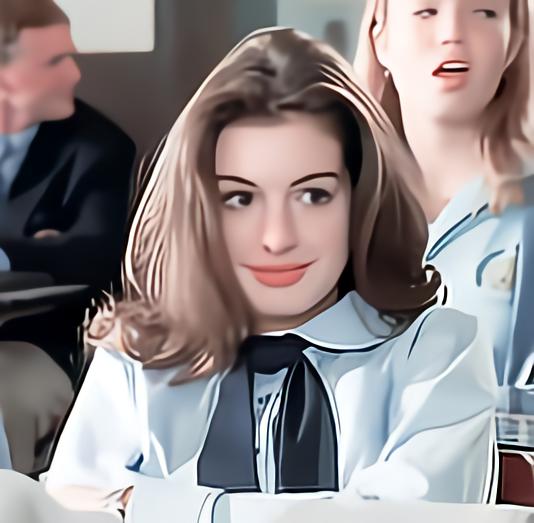}
            &
             \includegraphics[width=0.15\linewidth]{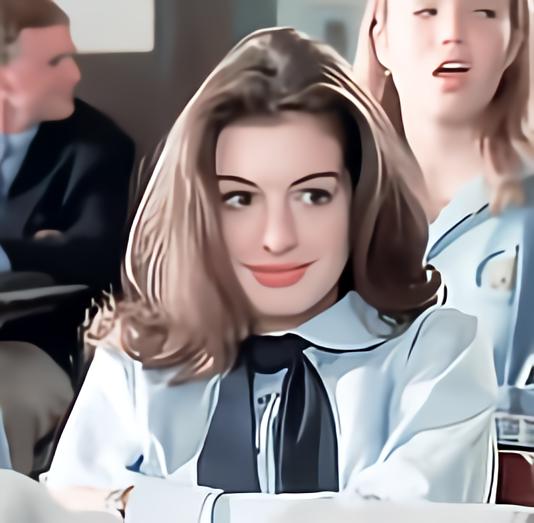}
            &
             \includegraphics[width=0.15\linewidth]{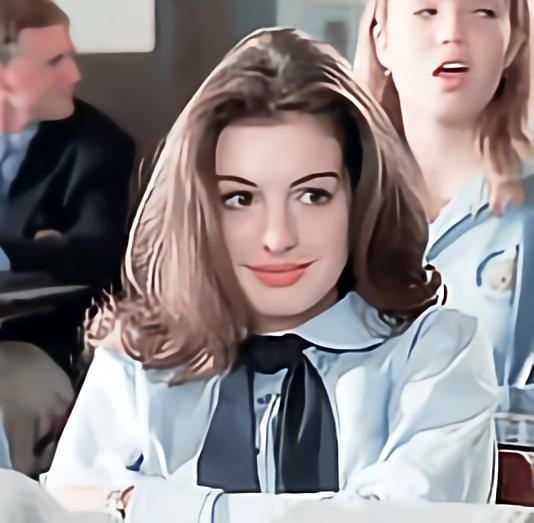}
            &
             \includegraphics[width=0.15\linewidth]{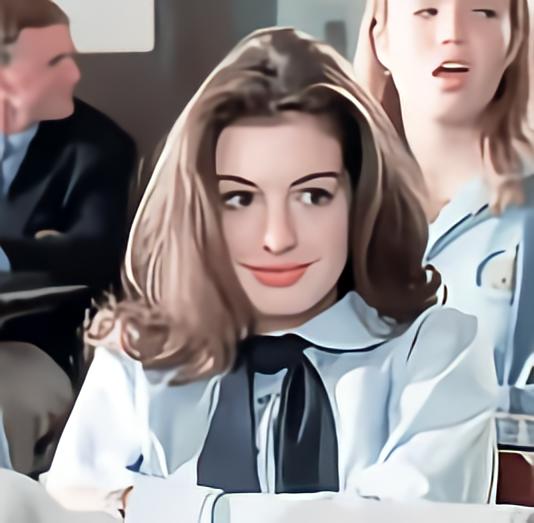}
            &
             \includegraphics[width=0.15\linewidth]{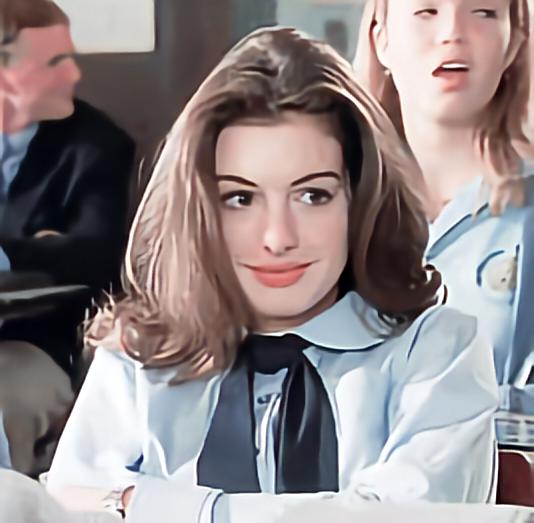}
             \vspace{-0.1cm} \\
            &
             \begin{tabular}{l}
             \vspace{-2.8cm} \\  \footnotesize  (0.8,0.8,0.8) 
             \end{tabular}
            &
             \includegraphics[width=0.15\linewidth]{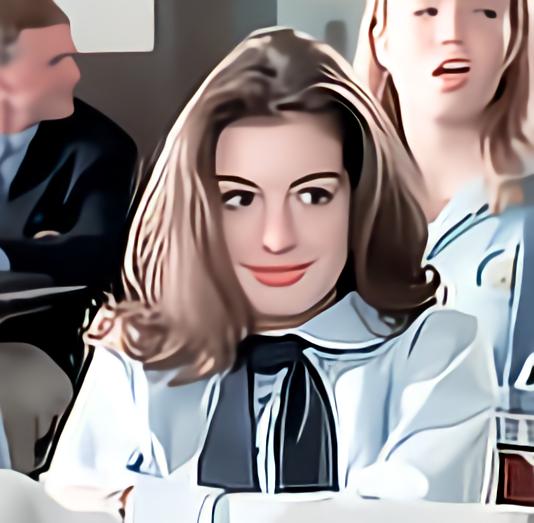}
            &
             \includegraphics[width=0.15\linewidth]{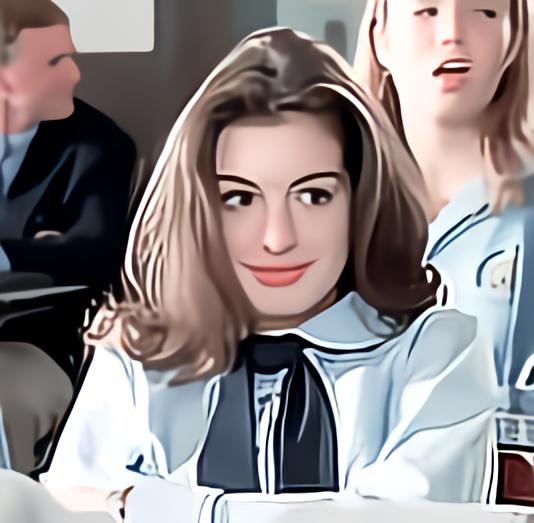}
            &
             \includegraphics[width=0.15\linewidth]{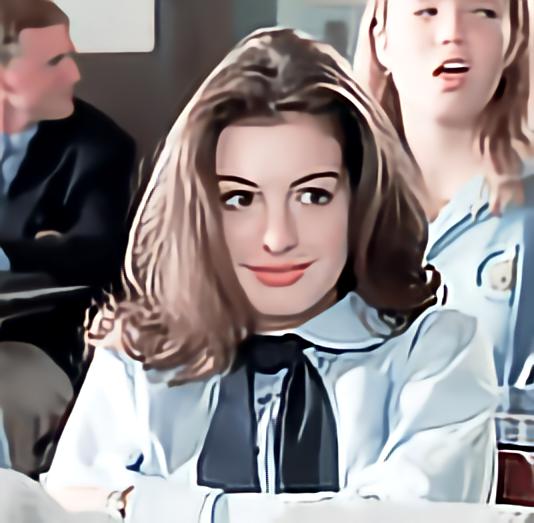}
            &
             \includegraphics[width=0.15\linewidth]{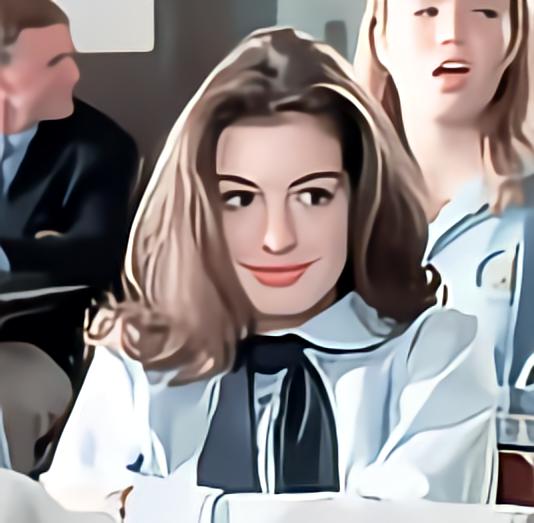}
            &
             \includegraphics[width=0.15\linewidth]{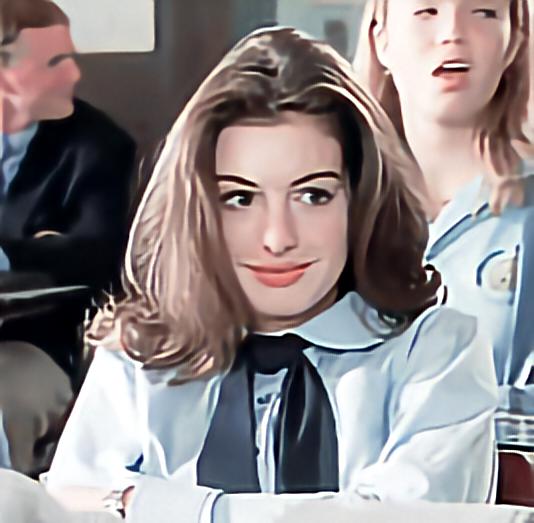}
             \vspace{-0.1cm} \\
            &
             \begin{tabular}{l}
             \vspace{-2.8cm} \\  \footnotesize  (1,1,1) 
             \end{tabular}
            &
             \includegraphics[width=0.15\linewidth]{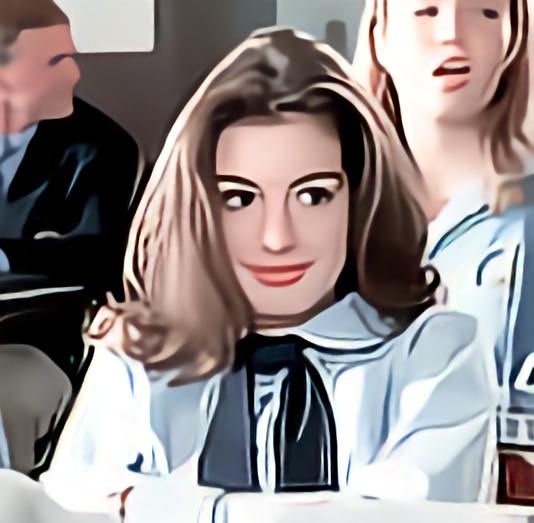}
            &
             \includegraphics[width=0.15\linewidth]{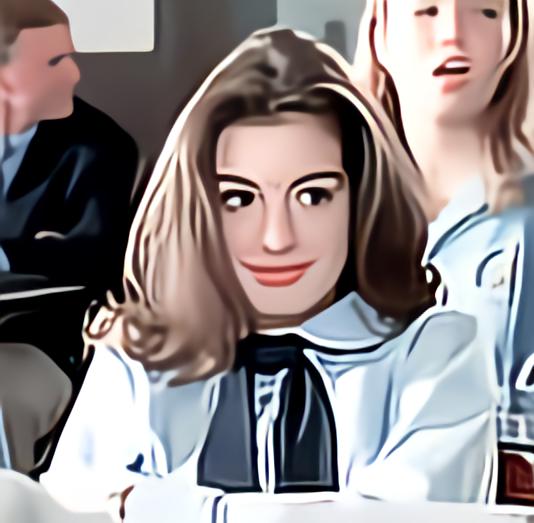}
            &
             \includegraphics[width=0.15\linewidth]{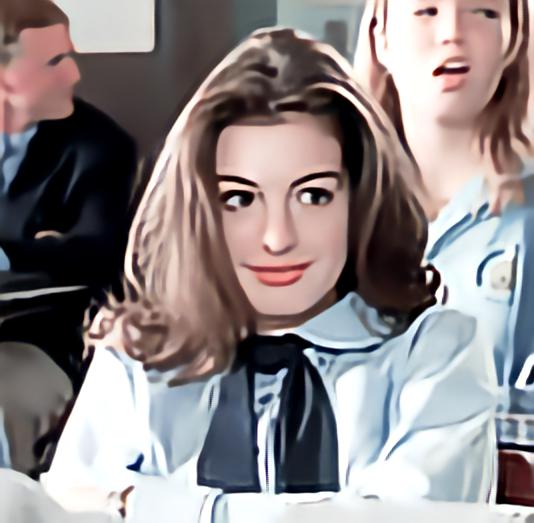}
            &
             \includegraphics[width=0.15\linewidth]{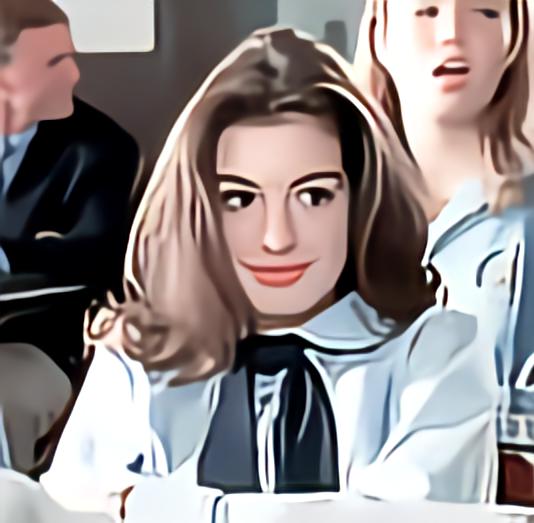}
            &
             \includegraphics[width=0.15\linewidth]{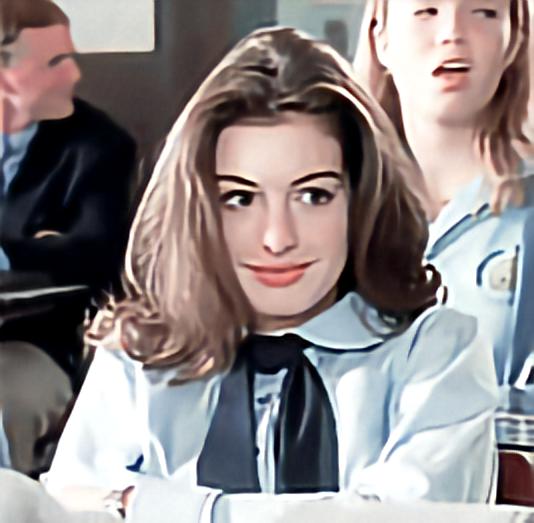}
             \vspace{-0.1cm} \\

	\toprule[1.5pt]
        \end{tabular}%\vspace{-0.3cm}
        \caption{\textbf{Deblur, denoise, and dejpeg modulation examples to the image with blur, noise, and jpeg compression}. Our TA+TSNet-m modulates diverse imagery effects with respect to the given restoration tasks. It generates less auxiliary visual artifacts and over-smoothed textures. The values of task vector denote restoration levels of (deblur, denoise, dejpeg), respectively.}\vspace{-0.0cm}
        \label{fig:synthetic_all}
    \end{center}
	\vspace{-0.2cm}
\end{figure*}

\begin{figure*}[p]
        \begin{center}\centering
        \setlength{\tabcolsep}{0.02cm}
        \setlength{\columnwidth}{2.81cm}
        \hspace*{-\tabcolsep}\begin{tabular}{ccccccc}
            \multicolumn{1}{c}{\footnotesize Input}    
            &
            \footnotesize Task vector
            &
            \multicolumn{1}{c}{\footnotesize CResMD~\cite{jingwen2020interactive}}
            &
            \multicolumn{1}{c}{ \footnotesize TSNet }
            &
            \multicolumn{1}{c}{\footnotesize TA+TSNet }
            &
            \multicolumn{1}{c}{\footnotesize TSNet-m }
            &
            \multicolumn{1}{c}{\footnotesize TA+TSNet-m }
            \\ \toprule[1.5pt]
            %\vspace{-0.1cm} \\
            &
             \begin{tabular}{c}
             \vspace{-2.8cm} \\  \footnotesize  (0,0,0) 
             \end{tabular}
            &
             \includegraphics[width=0.15\linewidth]{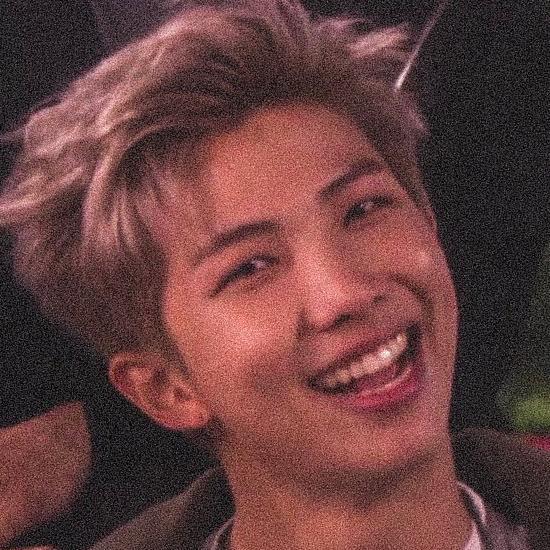}
            &
             \includegraphics[width=0.15\linewidth]{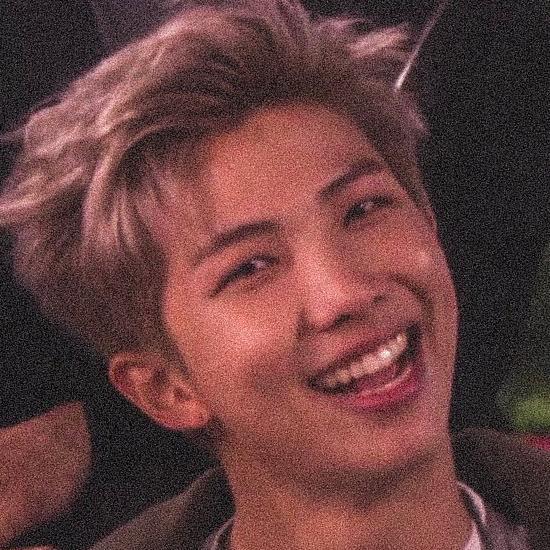}
            &
             \includegraphics[width=0.15\linewidth]{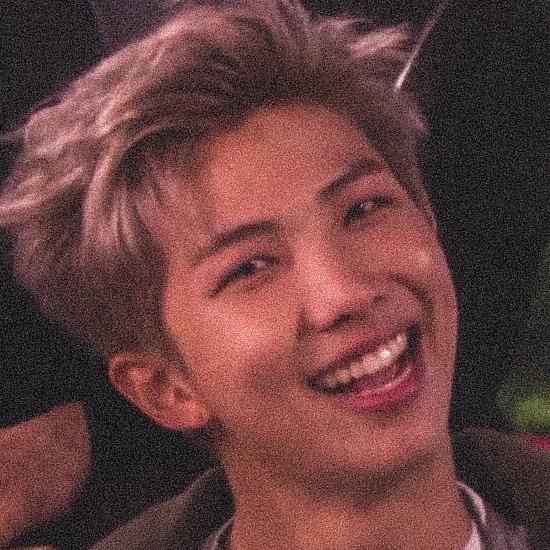}
            &
             \includegraphics[width=0.15\linewidth]{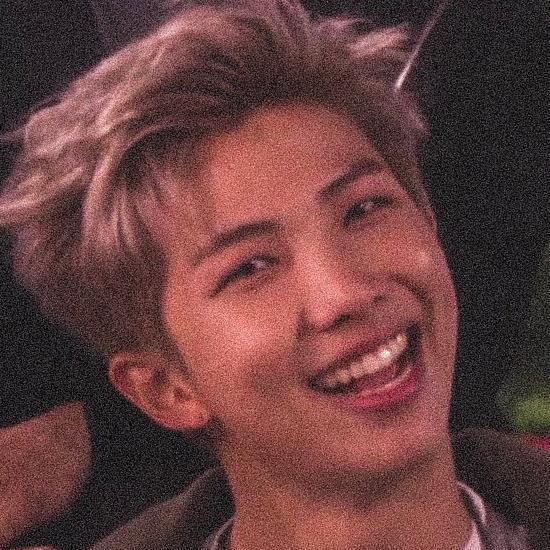}
            &
             \includegraphics[width=0.15\linewidth]{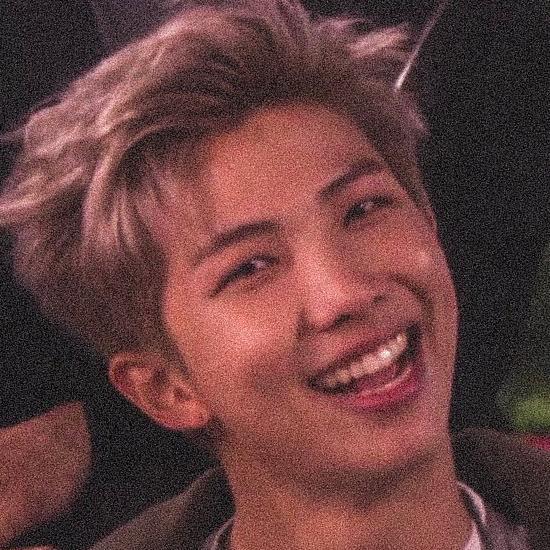}
             \vspace{-0.1cm} \\
             
            &
             \begin{tabular}{l}
             \vspace{-2.8cm} \\  \footnotesize  (0.2,0,0) 
             \end{tabular}
            &
             \includegraphics[width=0.15\linewidth]{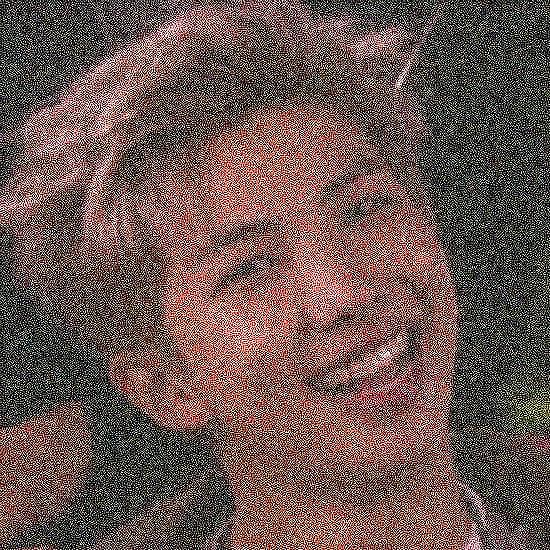}
            &
             \includegraphics[width=0.15\linewidth]{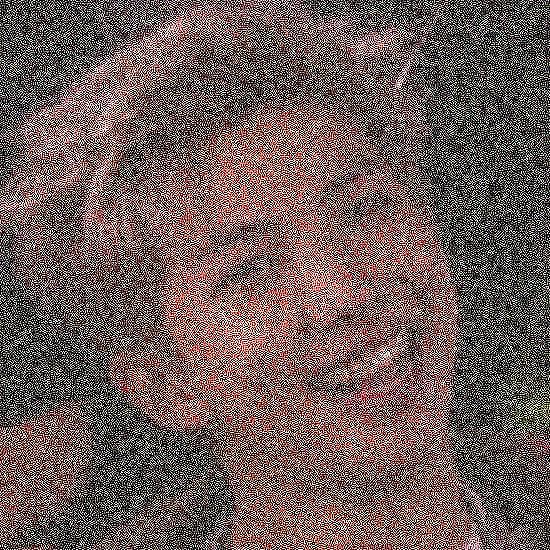}
            &
             \includegraphics[width=0.15\linewidth]{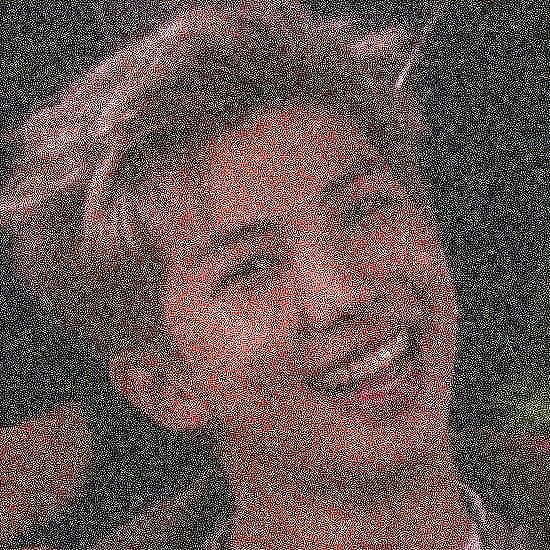}
            &
             \includegraphics[width=0.15\linewidth]{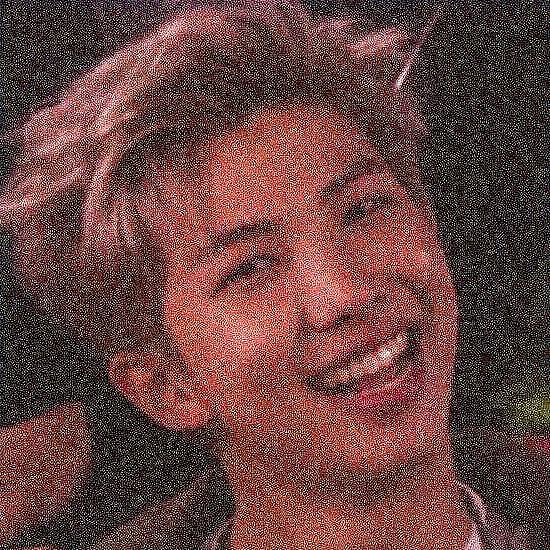}
            &
             \includegraphics[width=0.15\linewidth]{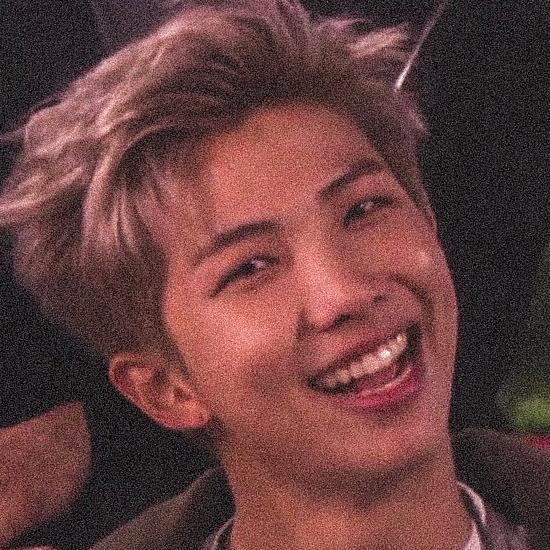}
             \vspace{-0.1cm} \\
            &
             \begin{tabular}{l}
             \vspace{-2.8cm} \\  \footnotesize  (0.4,0,0) 
             \end{tabular}
            &
             \includegraphics[width=0.15\linewidth]{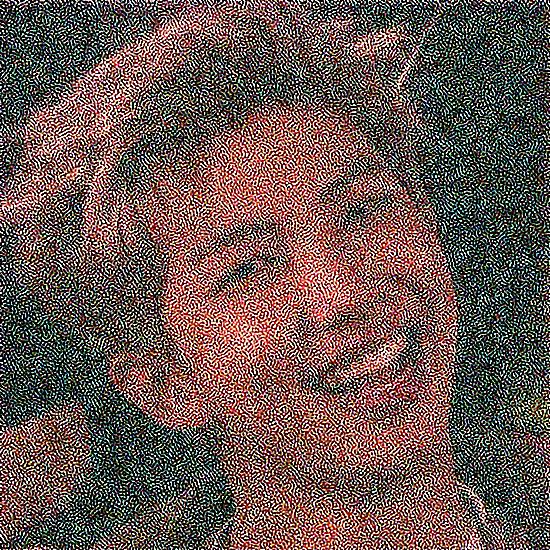}
            &
             \includegraphics[width=0.15\linewidth]{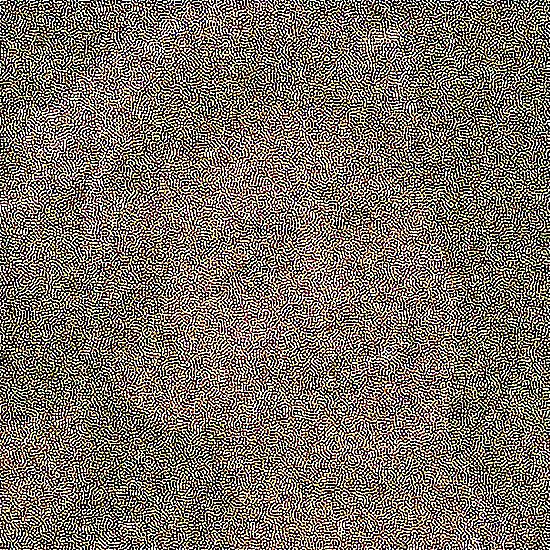}
            &
             \includegraphics[width=0.15\linewidth]{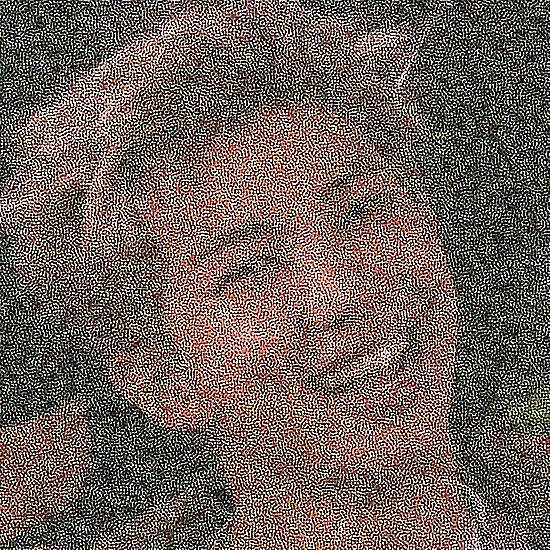}
            &
             \includegraphics[width=0.15\linewidth]{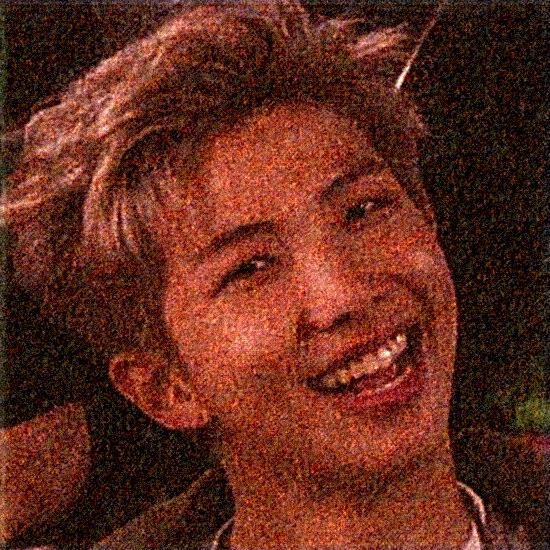}
            &
             \includegraphics[width=0.15\linewidth]{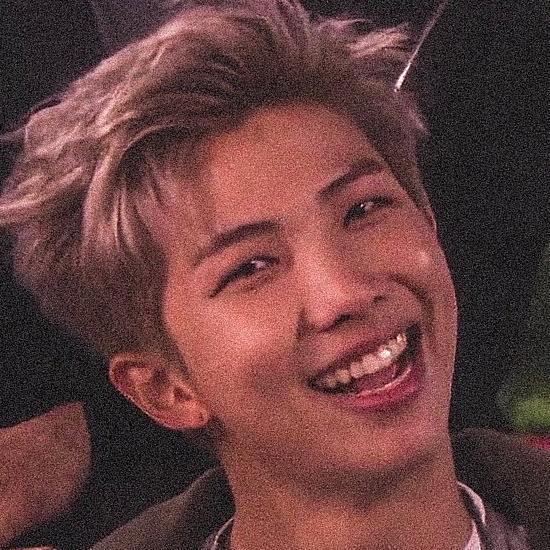}
             \vspace{-0.1cm} \\
             \begin{tabular}{c}
             \vspace{-2.8cm}\\
             \includegraphics[width=0.15\linewidth]{graphics/bts/bts.jpg}\\
             \vspace{-1cm} \\ {\color{white} Real } 
             \end{tabular}
            &
             \begin{tabular}{l}
             \vspace{-2.8cm} \\  \footnotesize  (0.5,0,0) 
             \end{tabular}
            &
             \includegraphics[width=0.15\linewidth]{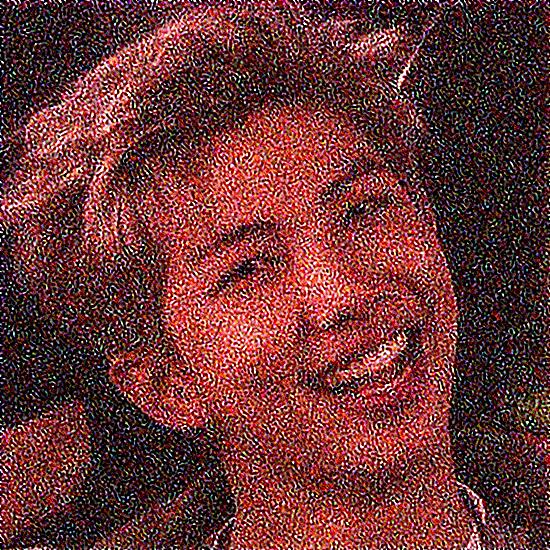}
            &
             \includegraphics[width=0.15\linewidth]{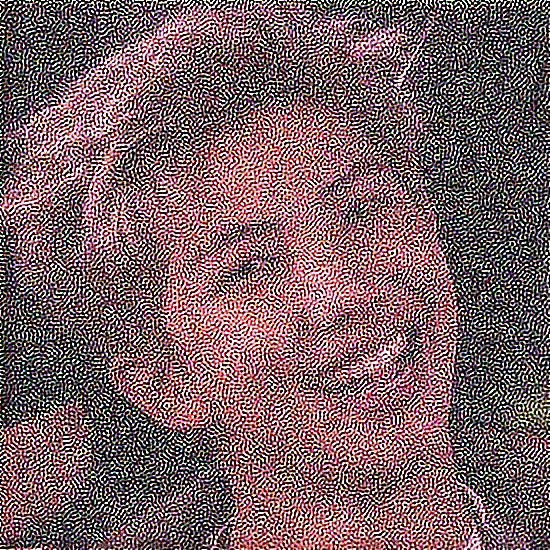}
            &
             \includegraphics[width=0.15\linewidth]{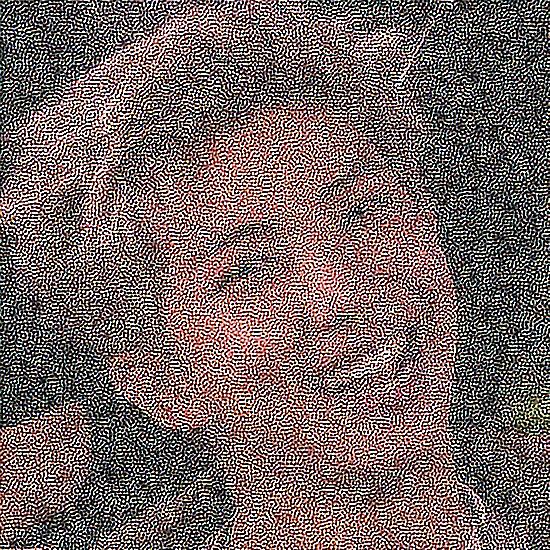}
            &
             \includegraphics[width=0.15\linewidth]{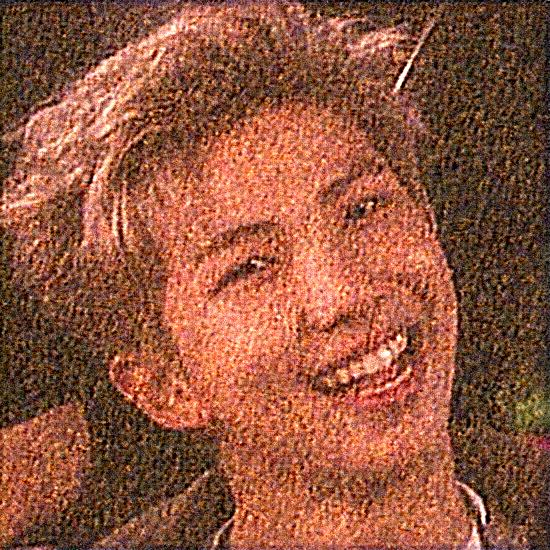}
            &
             \includegraphics[width=0.15\linewidth]{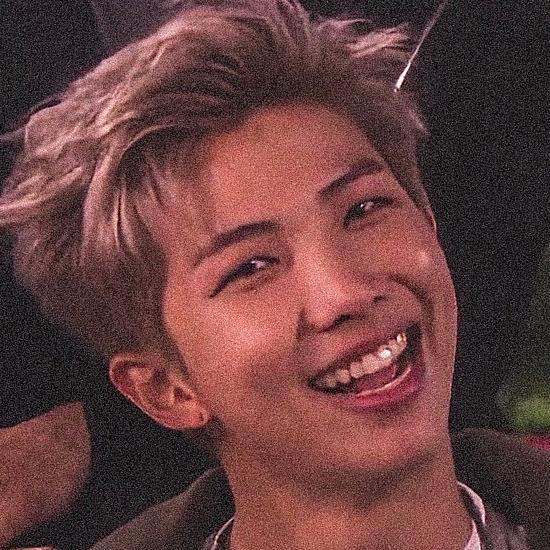}
             \vspace{-0.1cm} \\
             \begin{tabular}{c}
             \vspace{-4.2cm} \\ \footnotesize Optimal task vector  \vspace{-0.1cm} \\ \footnotesize =  \vspace{-0.1cm}\\ \footnotesize Unknown
             \end{tabular}
             &
             \begin{tabular}{l}
             \vspace{-2.8cm} \\  \footnotesize  (0.6,0,0) 
             \end{tabular}
            &
             \includegraphics[width=0.15\linewidth]{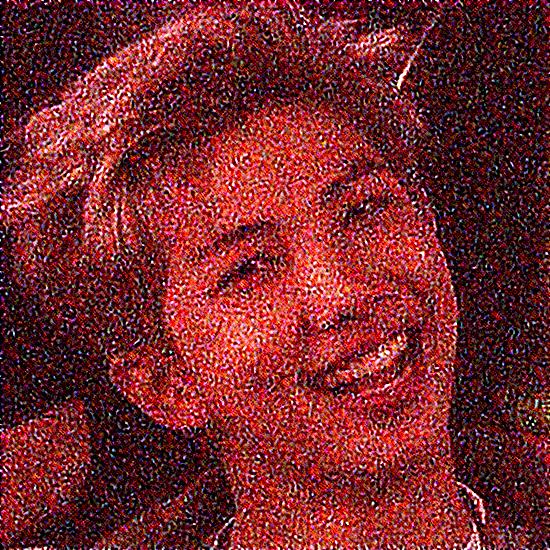}
            &
             \includegraphics[width=0.15\linewidth]{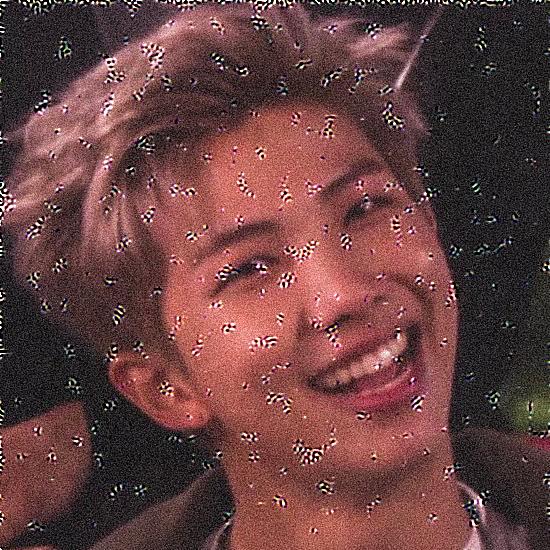}
            &
             \includegraphics[width=0.15\linewidth]{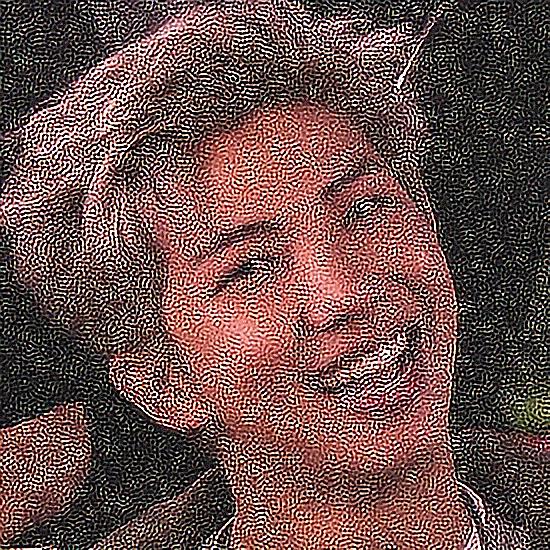}
            &
             \includegraphics[width=0.15\linewidth]{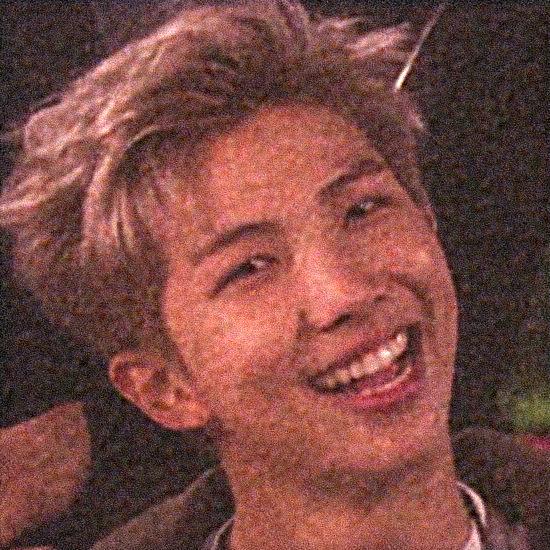}
            &
             \includegraphics[width=0.15\linewidth]{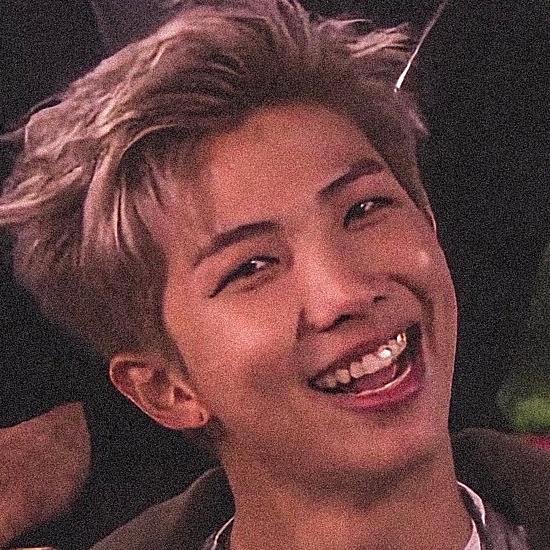}
             \vspace{-0.1cm} \\
            &
             \begin{tabular}{l}
             \vspace{-2.8cm} \\  \footnotesize  (0.8,0,0) 
             \end{tabular}
            &
             \includegraphics[width=0.15\linewidth]{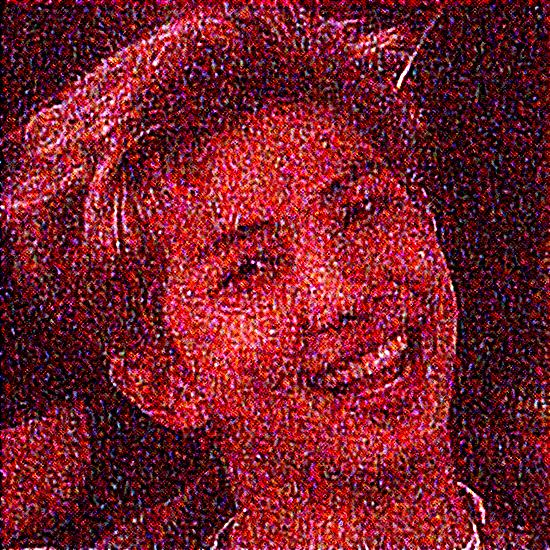}
            &
             \includegraphics[width=0.15\linewidth]{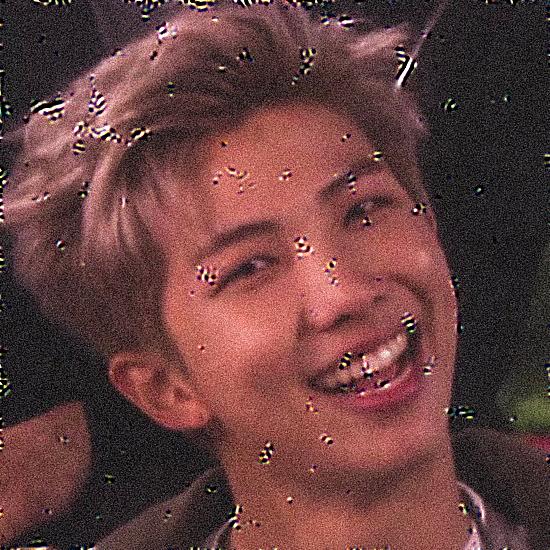}
            &
             \includegraphics[width=0.15\linewidth]{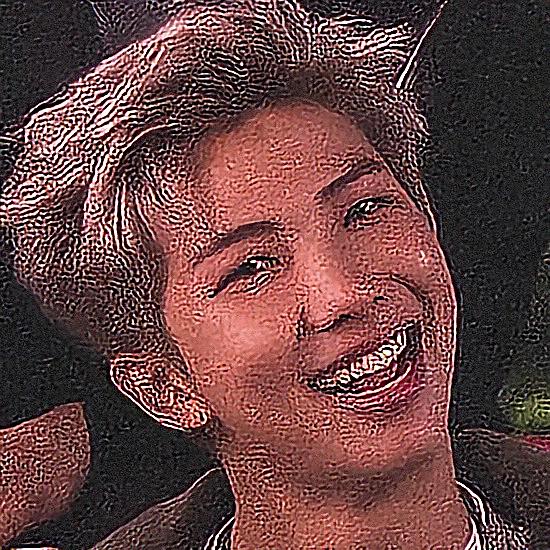}
            &
             \includegraphics[width=0.15\linewidth]{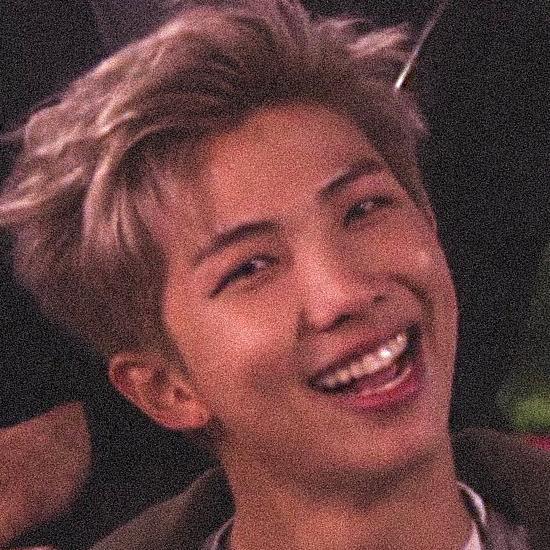}
            &
             \includegraphics[width=0.15\linewidth]{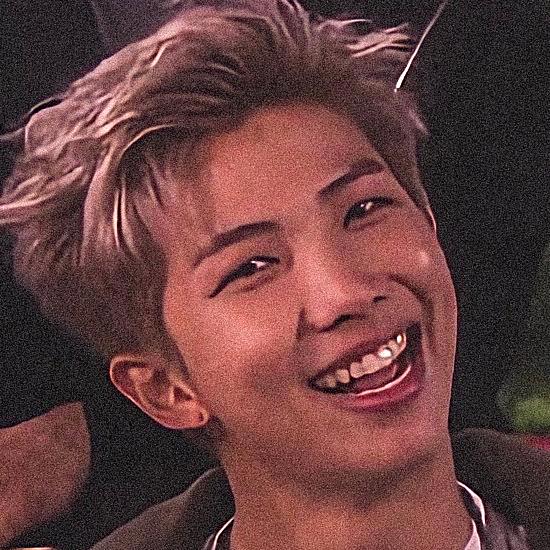}
             \vspace{-0.1cm} \\
            &
             \begin{tabular}{l}
             \vspace{-2.8cm} \\  \footnotesize  (1,0,0) 
             \end{tabular}
            &
             \includegraphics[width=0.15\linewidth]{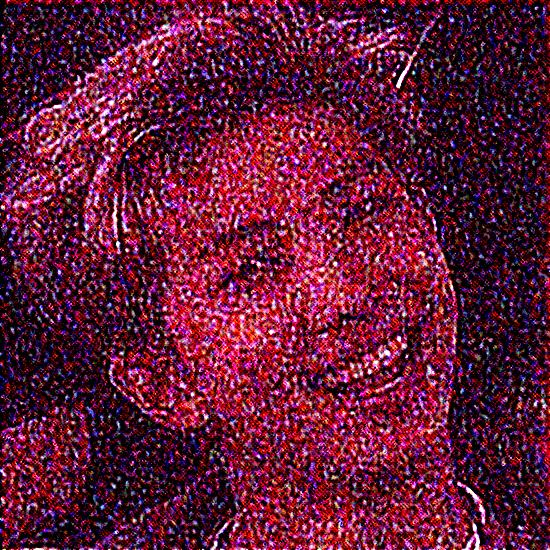}
            &
             \includegraphics[width=0.15\linewidth]{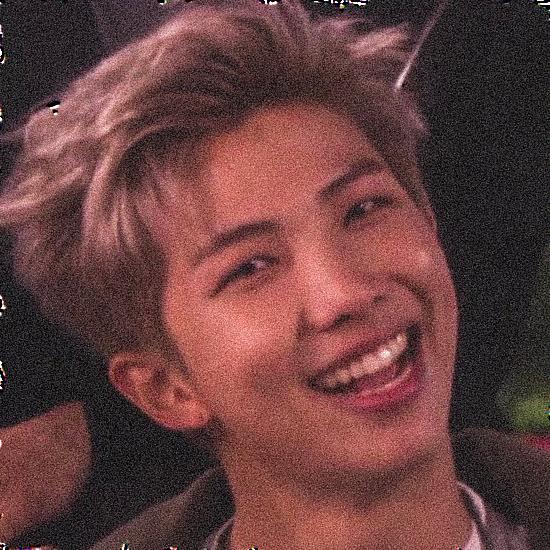}
            &
             \includegraphics[width=0.15\linewidth]{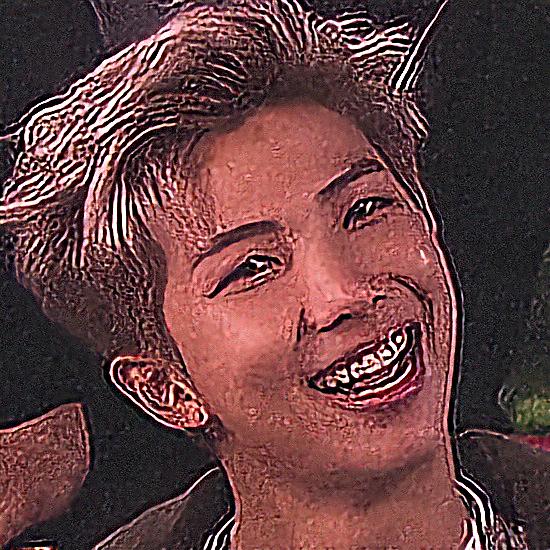}
            &
             \includegraphics[width=0.15\linewidth]{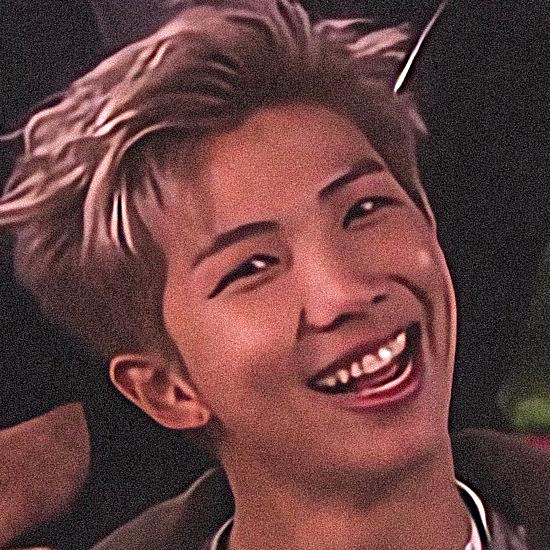}
            &
             \includegraphics[width=0.15\linewidth]{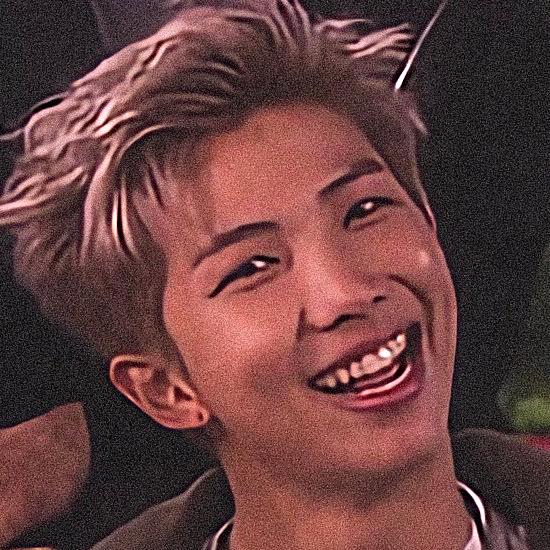}
             \vspace{-0.1cm} \\

	\toprule[1.5pt]
        \end{tabular}%\vspace{-0.3cm}
        \caption{\textbf{Deblur modulation examples to the real world image on the Internet}. Our TA+TSNet-m modulates diverse imagery effects with respect to the given restoration tasks. It generates less auxiliary visual artifacts. The values of task vector denote restoration levels of (deblur, denoise, dejpeg), respectively.}\vspace{-0.0cm}
        \label{fig:real_deblur}
    \end{center}
	\vspace{-0.2cm}
\end{figure*}

\begin{figure*}[p]
        \begin{center}\centering
        \setlength{\tabcolsep}{0.02cm}
        \setlength{\columnwidth}{2.81cm}
        \hspace*{-\tabcolsep}\begin{tabular}{ccccccc}
            \multicolumn{1}{c}{\footnotesize Input}    
            &
            \footnotesize Task vector
            &
            \multicolumn{1}{c}{\footnotesize CResMD~\cite{jingwen2020interactive}}
            &
            \multicolumn{1}{c}{ \footnotesize TSNet }
            &
            \multicolumn{1}{c}{\footnotesize TA+TSNet }
            &
            \multicolumn{1}{c}{\footnotesize TSNet-m }
            &
            \multicolumn{1}{c}{\footnotesize TA+TSNet-m }
            \\ \toprule[1.5pt]
            %\vspace{-0.1cm} \\
            &
             \begin{tabular}{c}
             \vspace{-2.8cm} \\  \footnotesize  (0,0,0) 
             \end{tabular}
            &
             \includegraphics[width=0.15\linewidth]{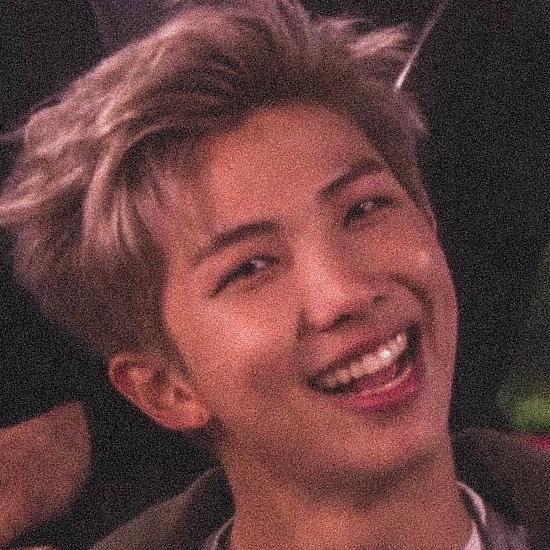}
            &
             \includegraphics[width=0.15\linewidth]{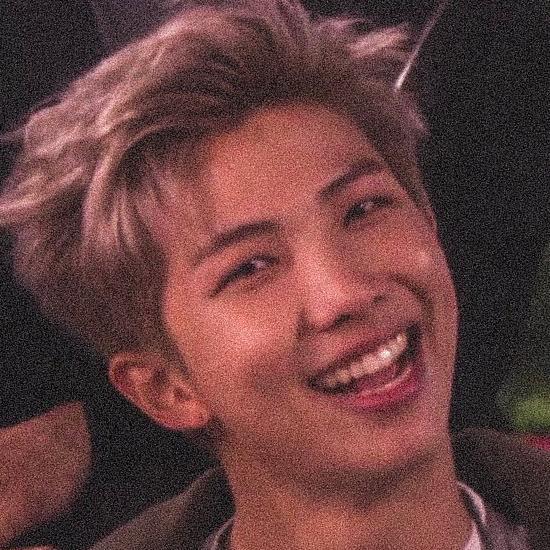}
            &
             \includegraphics[width=0.15\linewidth]{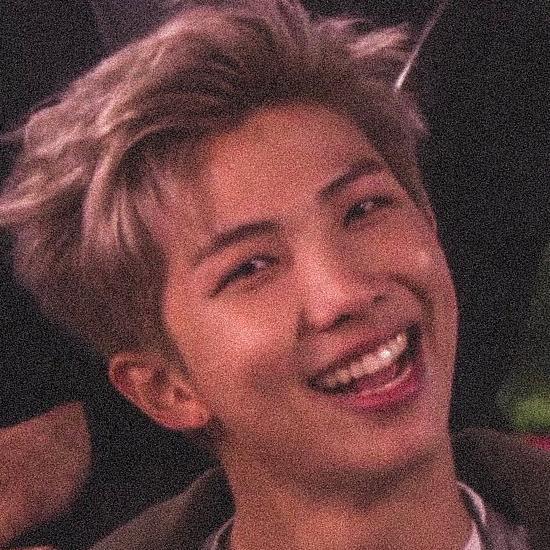}
            &
             \includegraphics[width=0.15\linewidth]{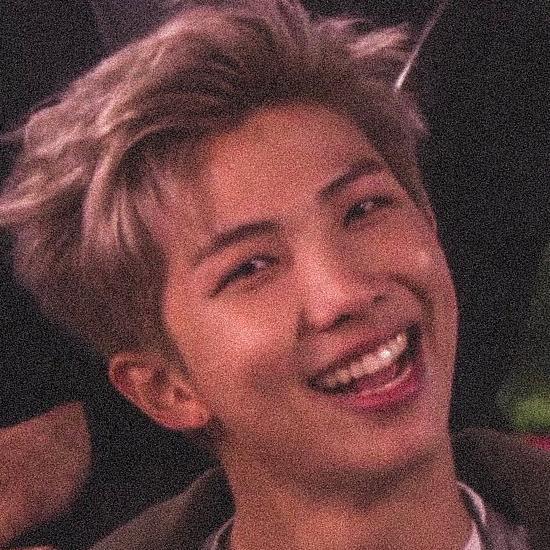}
            &
             \includegraphics[width=0.15\linewidth]{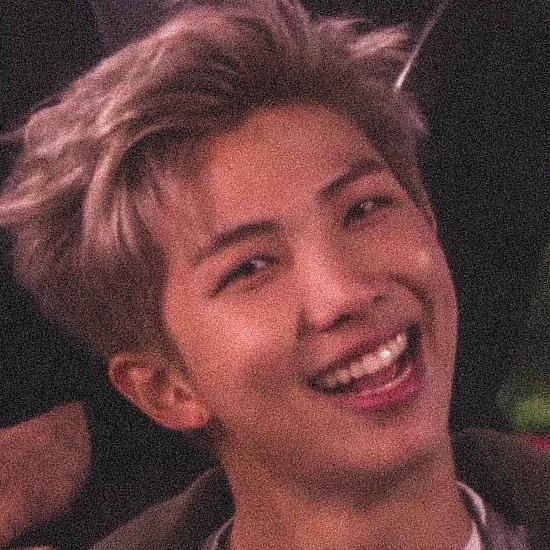}
             \vspace{-0.1cm} \\
             
            &
             \begin{tabular}{l}
             \vspace{-2.8cm} \\  \footnotesize  (0,0.2,0) 
             \end{tabular}
            &
             \includegraphics[width=0.15\linewidth]{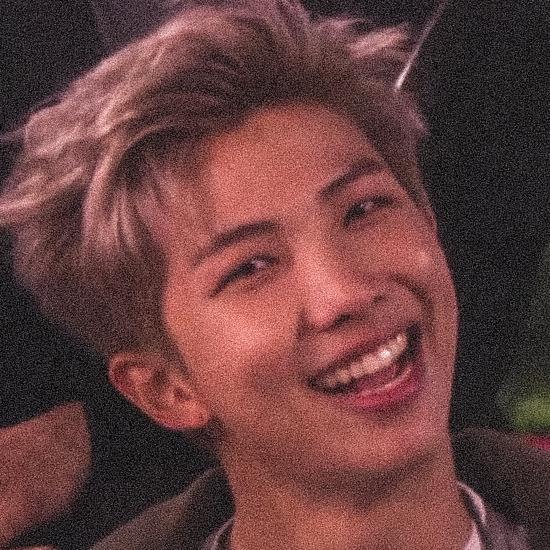}
            &
             \includegraphics[width=0.15\linewidth]{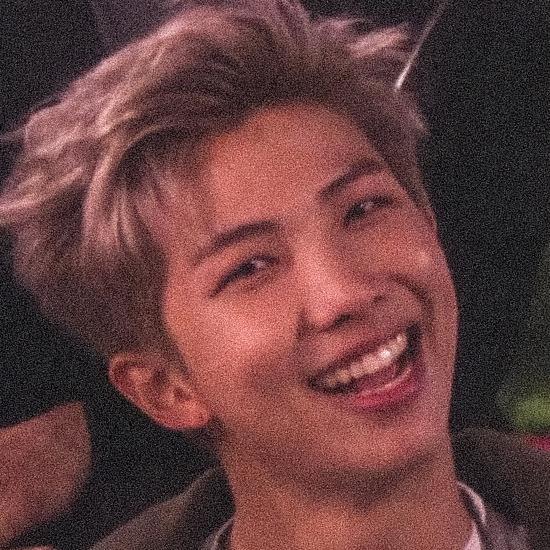}
            &
             \includegraphics[width=0.15\linewidth]{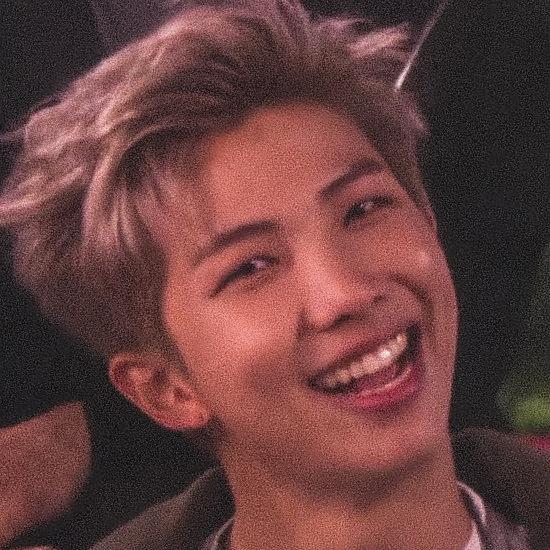}
            &
             \includegraphics[width=0.15\linewidth]{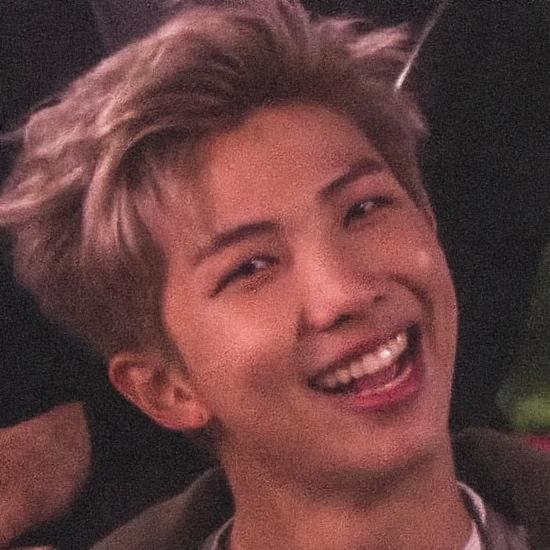}
            &
             \includegraphics[width=0.15\linewidth]{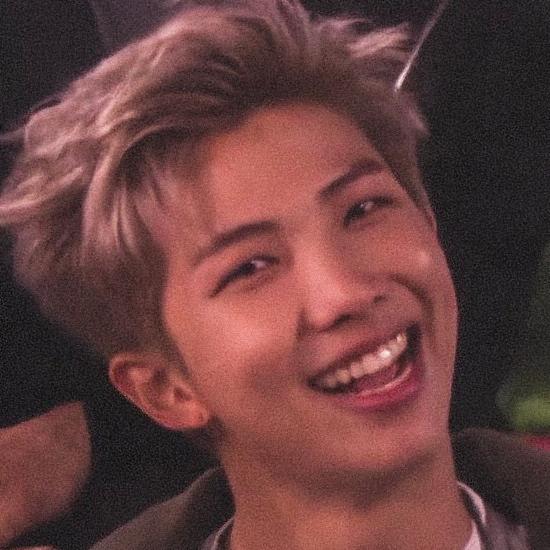}
             \vspace{-0.1cm} \\
            &
             \begin{tabular}{l}
             \vspace{-2.8cm} \\  \footnotesize  (0,0.4,0) 
             \end{tabular}
            &
             \includegraphics[width=0.15\linewidth]{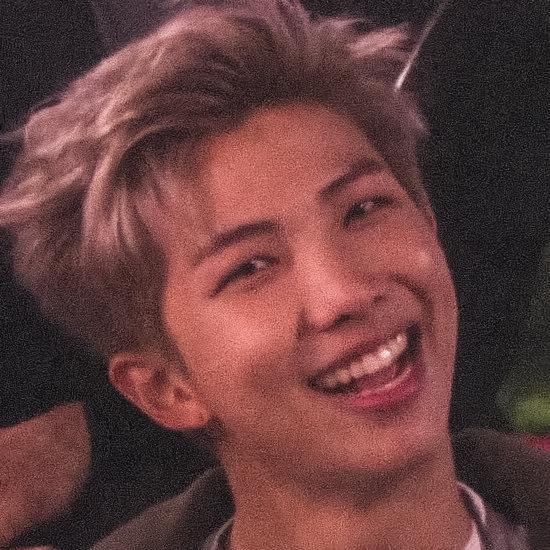}
            &
             \includegraphics[width=0.15\linewidth]{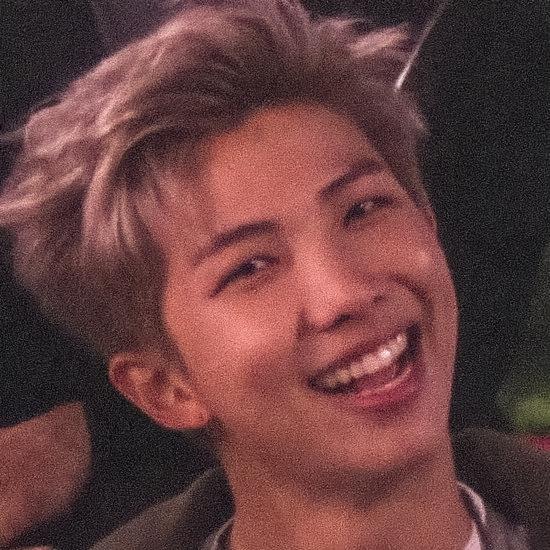}
            &
             \includegraphics[width=0.15\linewidth]{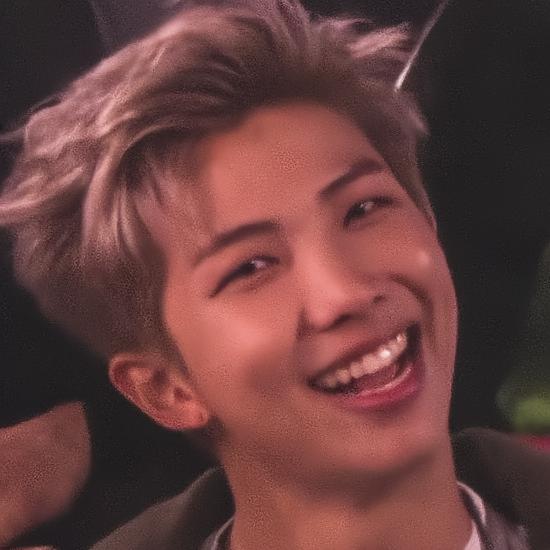}
            &
             \includegraphics[width=0.15\linewidth]{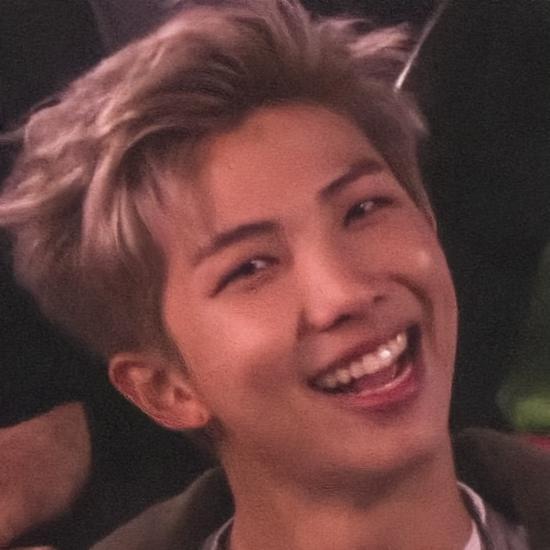}
            &
             \includegraphics[width=0.15\linewidth]{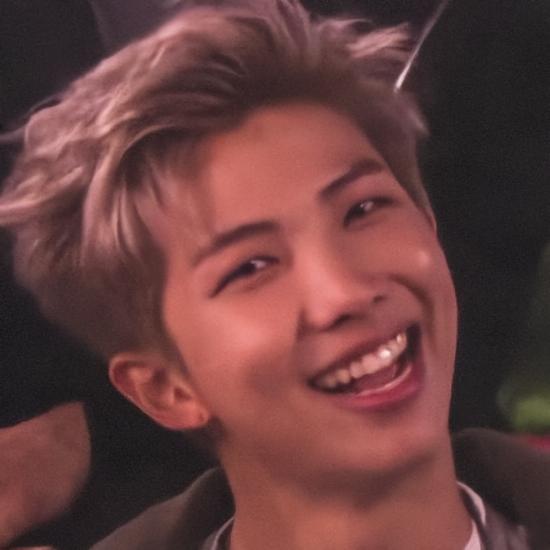}
             \vspace{-0.1cm} \\
             \begin{tabular}{c}
             \vspace{-2.8cm}\\
             \includegraphics[width=0.15\linewidth]{graphics/bts/bts.jpg}\\
             \vspace{-1cm} \\ {\color{white} Real } 
             \end{tabular}
            &
             \begin{tabular}{l}
             \vspace{-2.8cm} \\  \footnotesize  (0,0.5,0) 
             \end{tabular}
            &
             \includegraphics[width=0.15\linewidth]{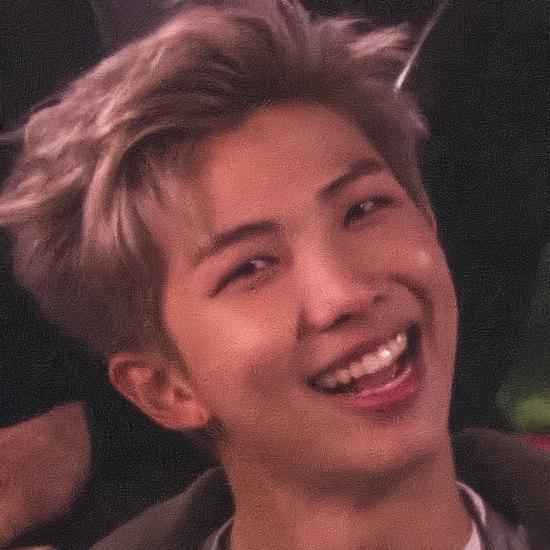}
            &
             \includegraphics[width=0.15\linewidth]{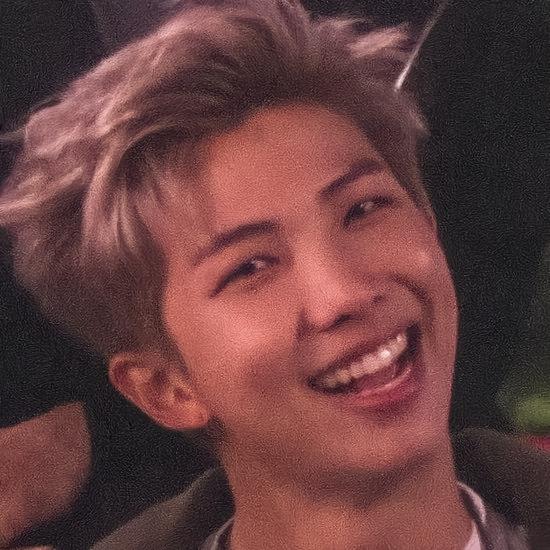}
            &
             \includegraphics[width=0.15\linewidth]{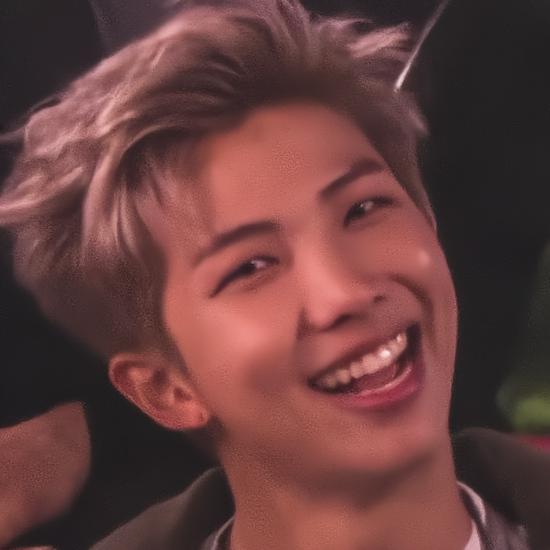}
            &
             \includegraphics[width=0.15\linewidth]{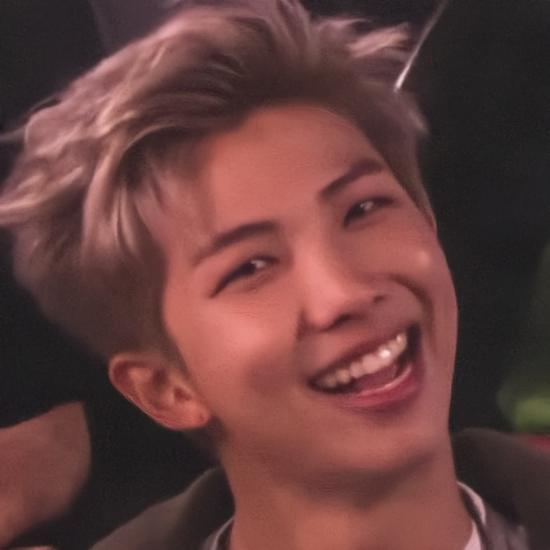}
            &
             \includegraphics[width=0.15\linewidth]{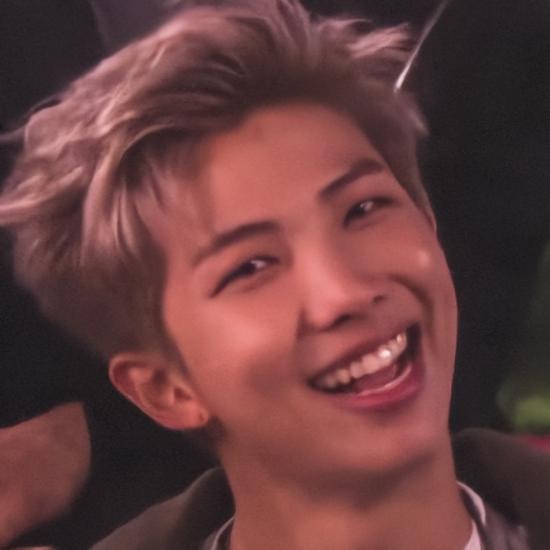}
             \vspace{-0.1cm} \\
             \begin{tabular}{c}
             \vspace{-4.2cm} \\ \footnotesize Optimal task vector  \vspace{-0.1cm} \\ \footnotesize =  \vspace{-0.1cm}\\ \footnotesize Unknown
             \end{tabular}
             &
             \begin{tabular}{l}
             \vspace{-2.8cm} \\  \footnotesize  (0,0.6,0) 
             \end{tabular}
            &
             \includegraphics[width=0.15\linewidth]{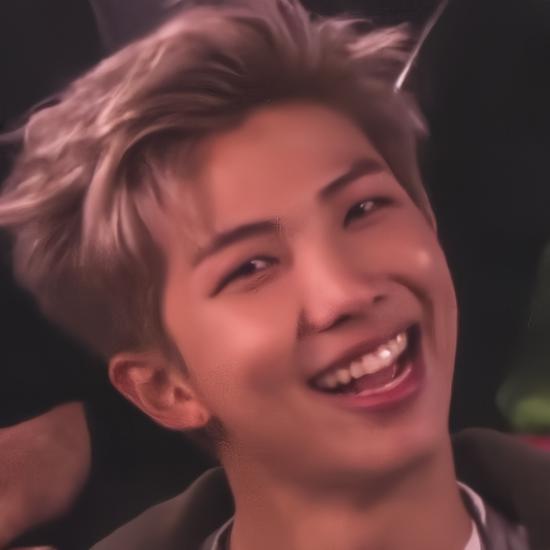}
            &
             \includegraphics[width=0.15\linewidth]{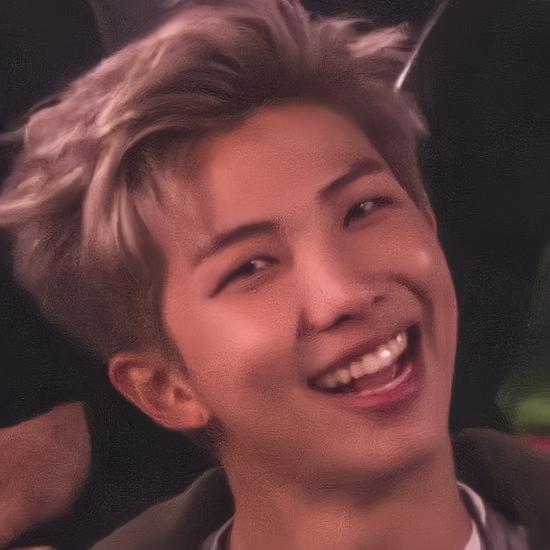}
            &
             \includegraphics[width=0.15\linewidth]{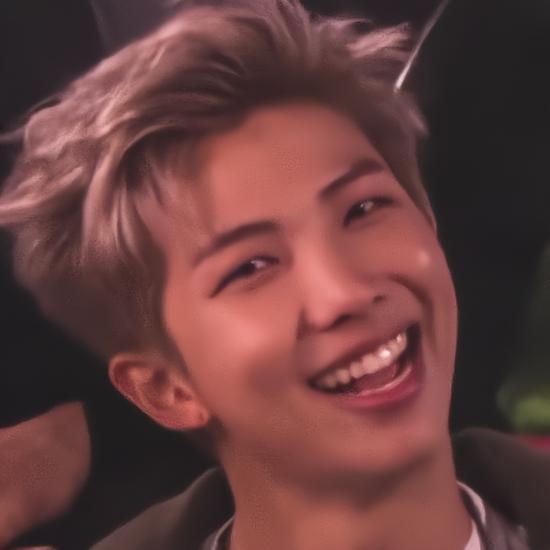}
            &
             \includegraphics[width=0.15\linewidth]{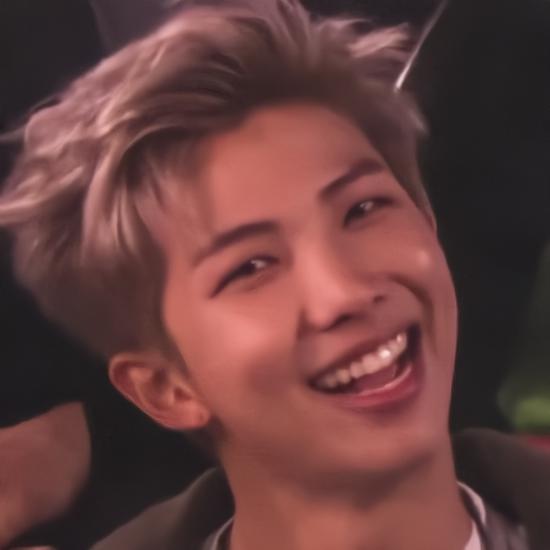}
            &
             \includegraphics[width=0.15\linewidth]{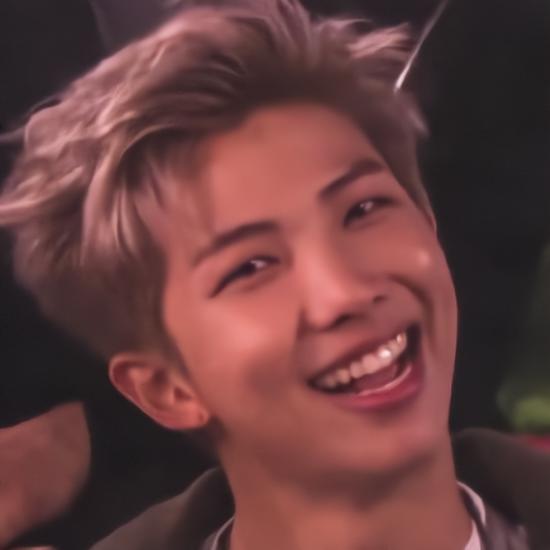}
             \vspace{-0.1cm} \\
            &
             \begin{tabular}{l}
             \vspace{-2.8cm} \\  \footnotesize  (0,0.8,0) 
             \end{tabular}
            &
             \includegraphics[width=0.15\linewidth]{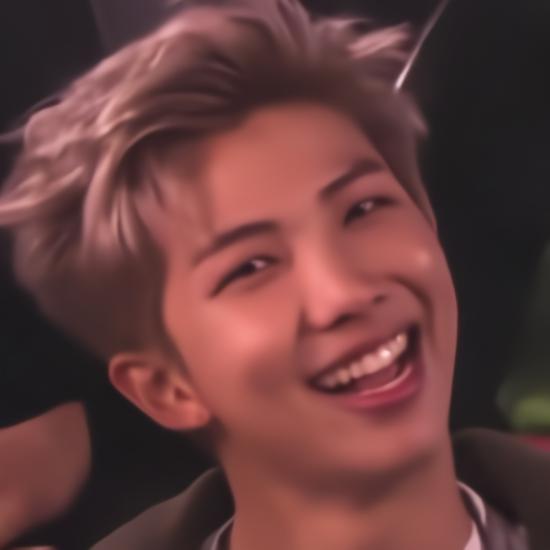}
            &
             \includegraphics[width=0.15\linewidth]{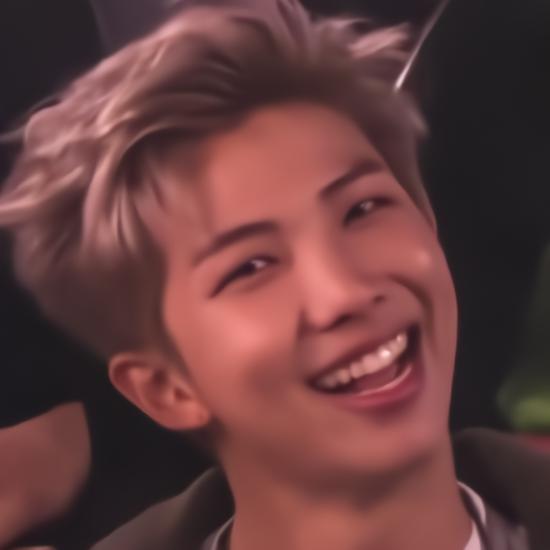}
            &
             \includegraphics[width=0.15\linewidth]{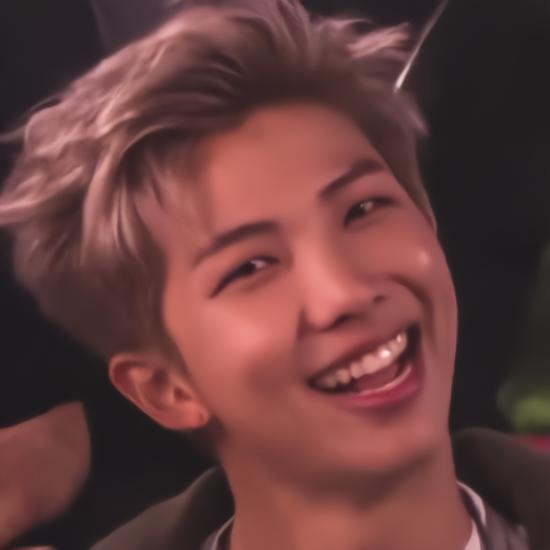}
            &
             \includegraphics[width=0.15\linewidth]{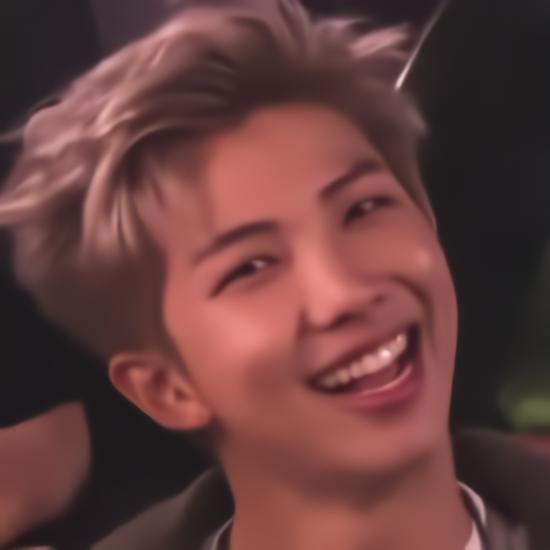}
            &
             \includegraphics[width=0.15\linewidth]{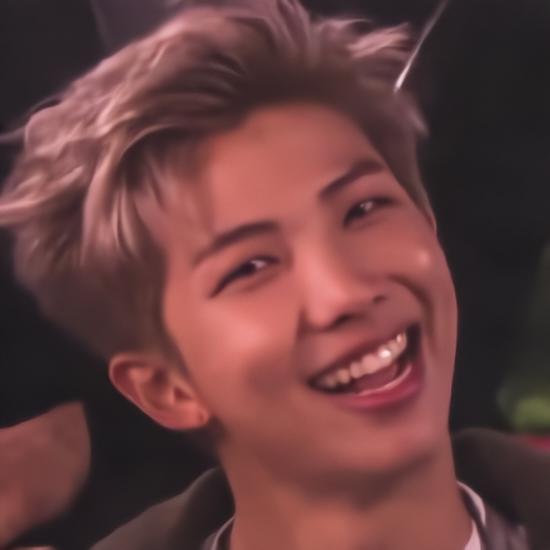}
             \vspace{-0.1cm} \\
            &
             \begin{tabular}{l}
             \vspace{-2.8cm} \\  \footnotesize  (0,1,0) 
             \end{tabular}
            &
             \includegraphics[width=0.15\linewidth]{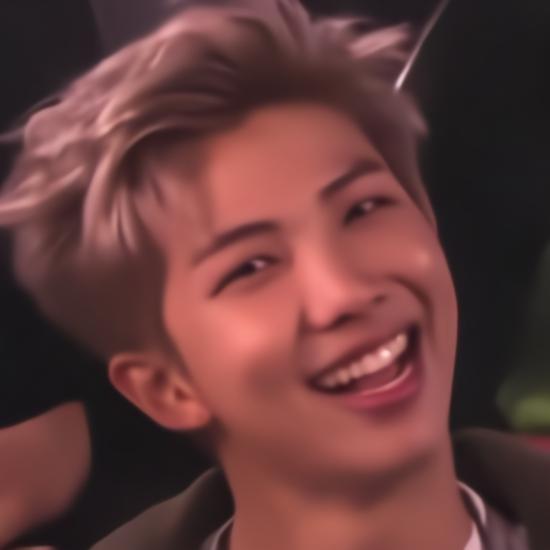}
            &
             \includegraphics[width=0.15\linewidth]{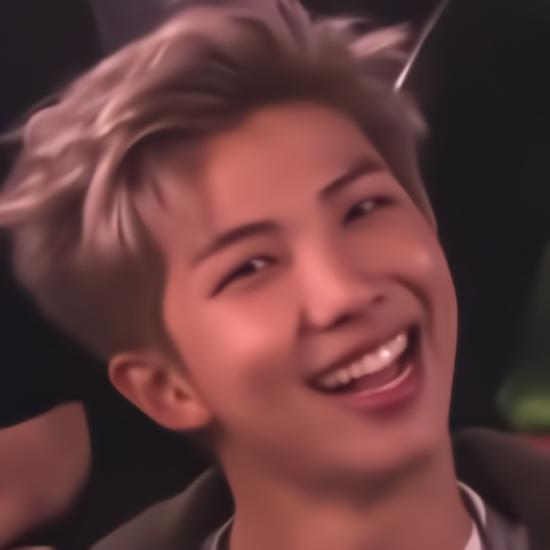}
            &
             \includegraphics[width=0.15\linewidth]{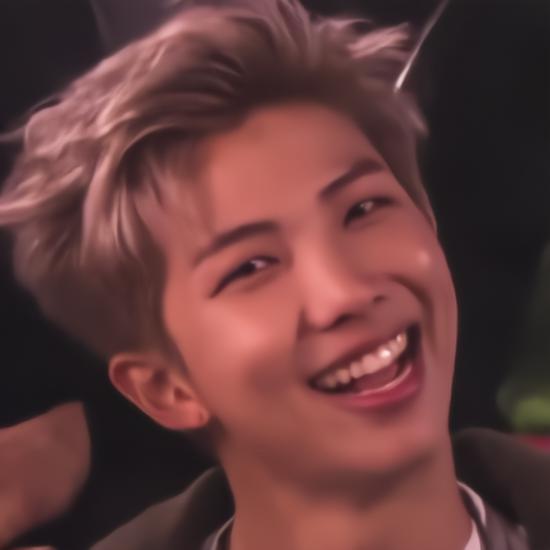}
            &
             \includegraphics[width=0.15\linewidth]{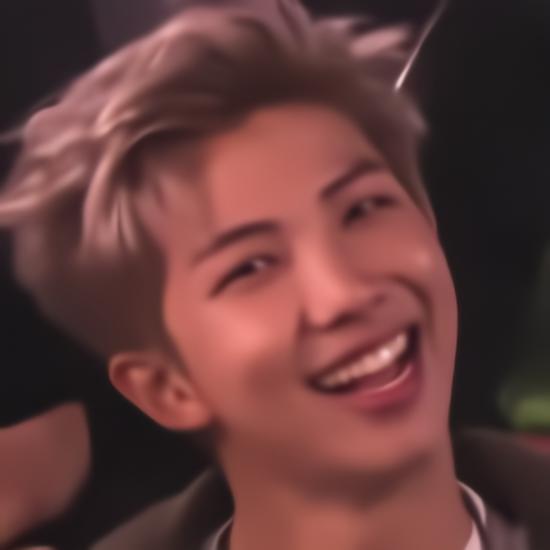}
            &
             \includegraphics[width=0.15\linewidth]{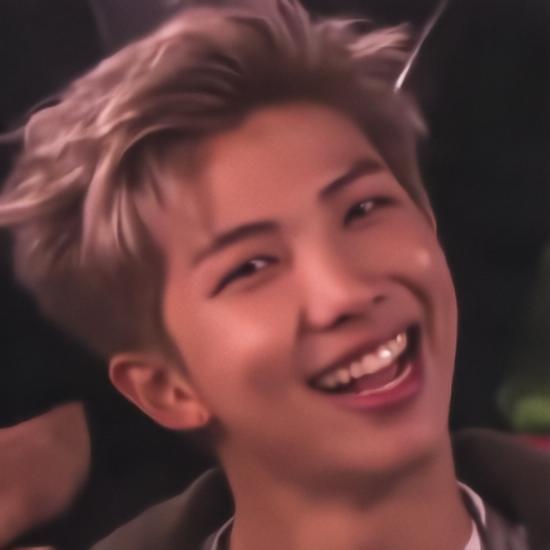}
             \vspace{-0.1cm} \\

	\toprule[1.5pt]
        \end{tabular}%\vspace{-0.3cm}
        \caption{\textbf{Denoise modulation examples to the real world image on the Internet}. Our TA+TSNet-m modulates diverse imagery effects with respect to the given restoration tasks. It generates less auxiliary visual artifacts and over-smoothed textures. The values of task vector denote restoration levels of (deblur, denoise, dejpeg), respectively.}\vspace{-0.0cm}
        \label{fig:real_denoise}
    \end{center}
	\vspace{-0.2cm}
\end{figure*}

\begin{figure*}[p]
        \begin{center}\centering
        \setlength{\tabcolsep}{0.02cm}
        \setlength{\columnwidth}{2.81cm}
        \hspace*{-\tabcolsep}\begin{tabular}{ccccccc}
            \multicolumn{1}{c}{\footnotesize Input}    
            &
            \footnotesize Task vector
            &
            \multicolumn{1}{c}{\footnotesize CResMD~\cite{jingwen2020interactive}}
            &
            \multicolumn{1}{c}{ \footnotesize TSNet }
            &
            \multicolumn{1}{c}{\footnotesize TA+TSNet }
            &
            \multicolumn{1}{c}{\footnotesize TSNet-m }
            &
            \multicolumn{1}{c}{\footnotesize TA+TSNet-m }
            \\ \toprule[1.5pt]
            %\vspace{-0.1cm} \\
            &
             \begin{tabular}{c}
             \vspace{-2.8cm} \\  \footnotesize  (0,0,0) 
             \end{tabular}
            &
             \includegraphics[width=0.15\linewidth]{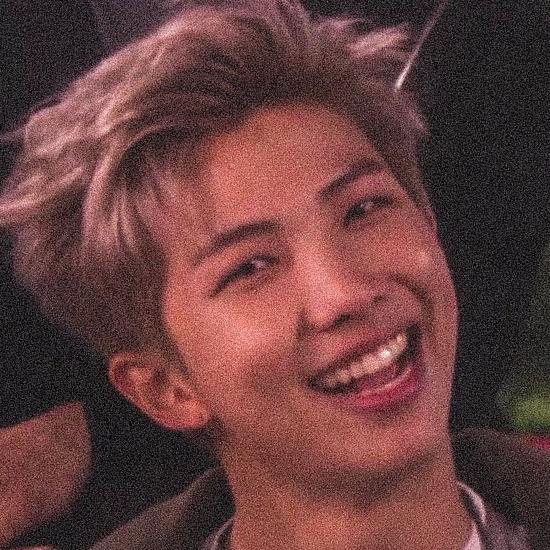}
            &
             \includegraphics[width=0.15\linewidth]{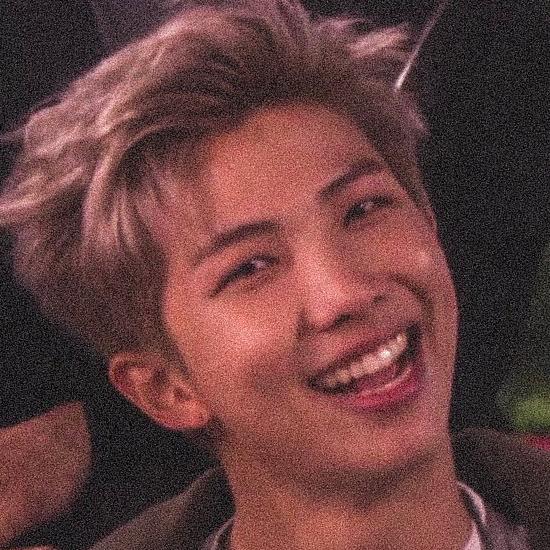}
            &
             \includegraphics[width=0.15\linewidth]{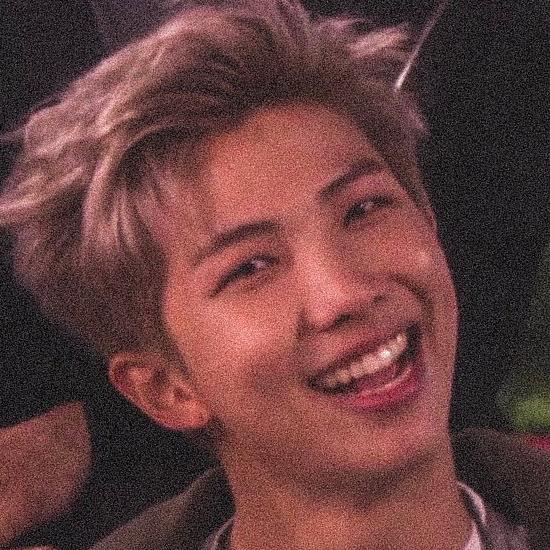}
            &
             \includegraphics[width=0.15\linewidth]{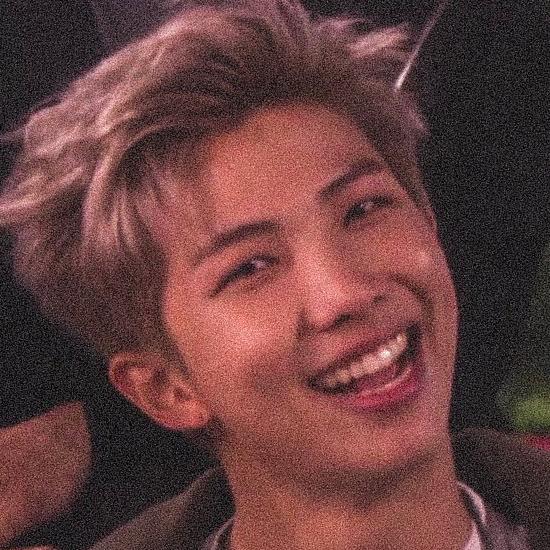}
            &
             \includegraphics[width=0.15\linewidth]{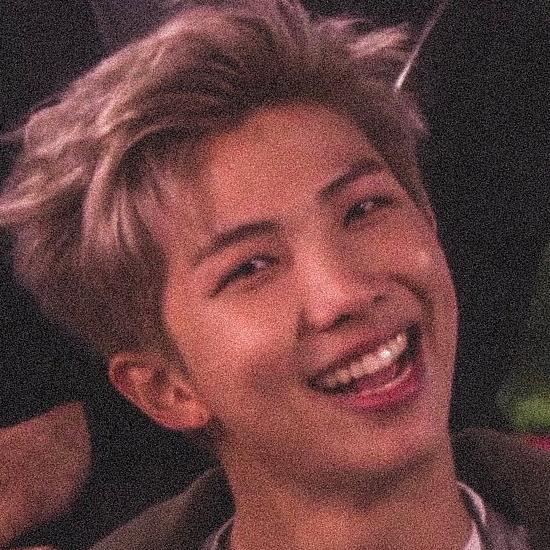}
             \vspace{-0.1cm} \\
             
            &
             \begin{tabular}{l}
             \vspace{-2.8cm} \\  \footnotesize  (0.2,0.2,0.2) 
             \end{tabular}
            &
             \includegraphics[width=0.15\linewidth]{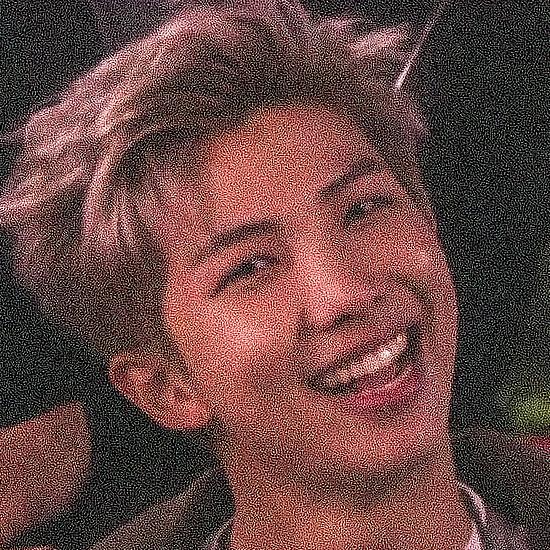}
            &
             \includegraphics[width=0.15\linewidth]{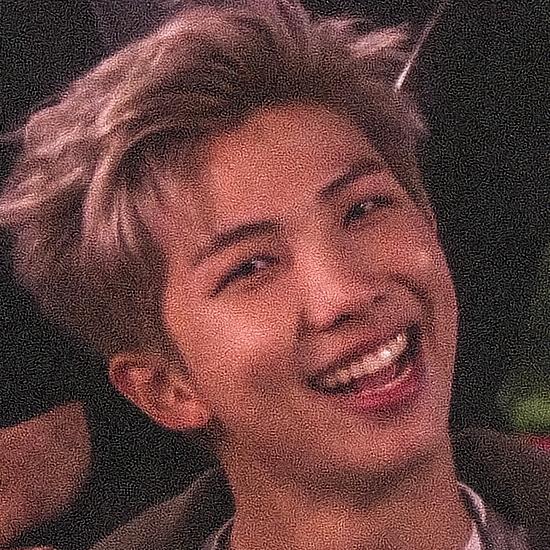}
            &
             \includegraphics[width=0.15\linewidth]{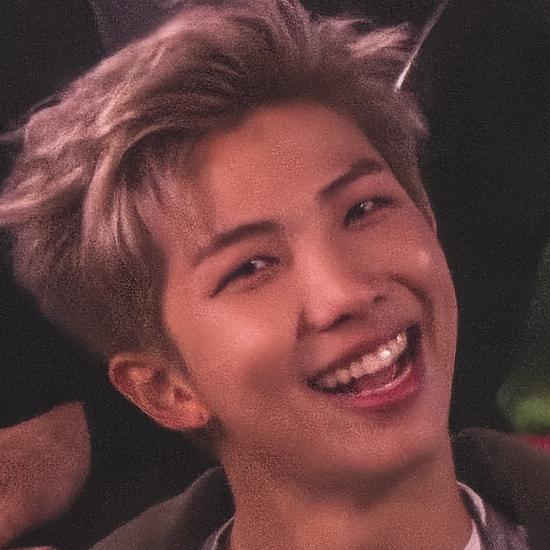}
            &
             \includegraphics[width=0.15\linewidth]{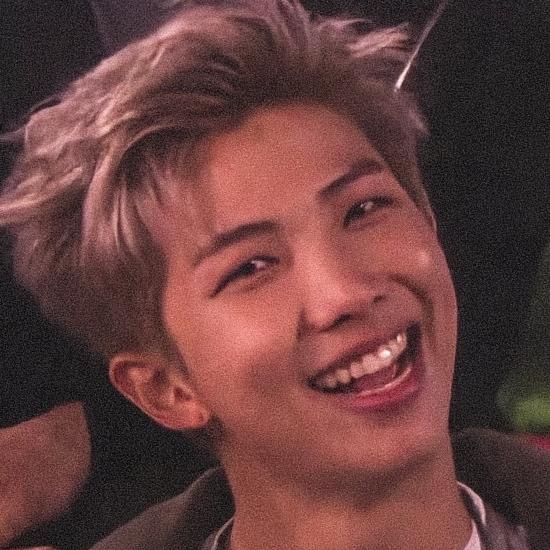}
            &
             \includegraphics[width=0.15\linewidth]{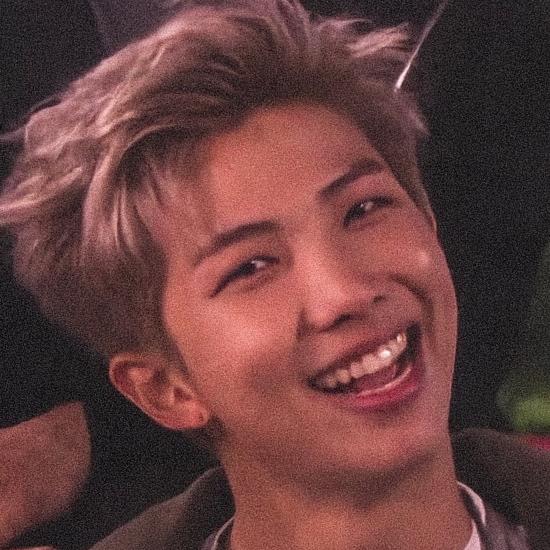}
             \vspace{-0.1cm} \\
            &
             \begin{tabular}{l}
             \vspace{-2.8cm} \\  \footnotesize  (0.4,0.4,0.4) 
             \end{tabular}
            &
             \includegraphics[width=0.15\linewidth]{graphics/bts/CResMD/all/04.jpg}
            &
             \includegraphics[width=0.15\linewidth]{graphics/bts/TSNet/all/04.jpg}
            &
             \includegraphics[width=0.15\linewidth]{graphics/bts/TA+TSNet/all/04.jpg}
            &
             \includegraphics[width=0.15\linewidth]{graphics/bts/TSNet-m/all/04.jpg}
            &
             \includegraphics[width=0.15\linewidth]{graphics/bts/TA+TSNet-m/all/04.jpg}
             \vspace{-0.1cm} \\
             \begin{tabular}{c}
             \vspace{-2.8cm}\\
             \includegraphics[width=0.15\linewidth]{graphics/bts/bts.jpg}\\
             \vspace{-1cm} \\ {\color{white} Real } 
             \end{tabular}
            &
             \begin{tabular}{l}
             \vspace{-2.8cm} \\  \footnotesize  (0.5,0.5,0.5) 
             \end{tabular}
            &
             \includegraphics[width=0.15\linewidth]{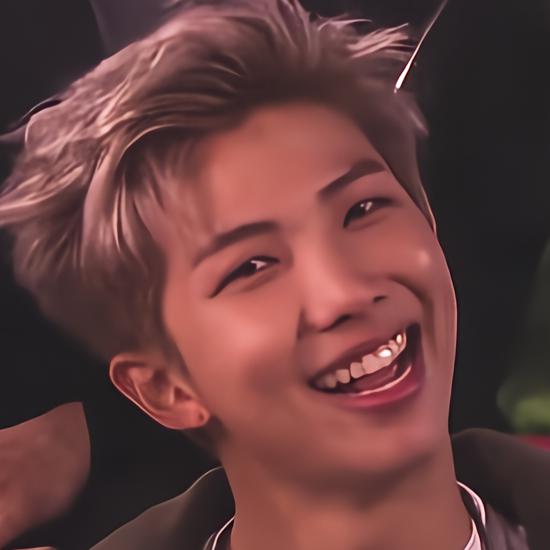}
            &
             \includegraphics[width=0.15\linewidth]{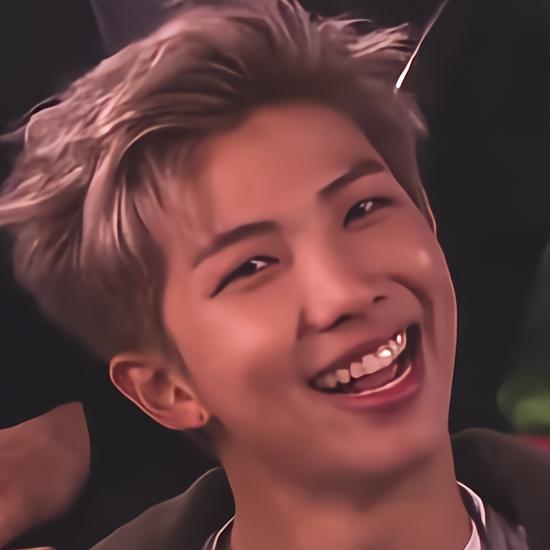}
            &
             \includegraphics[width=0.15\linewidth]{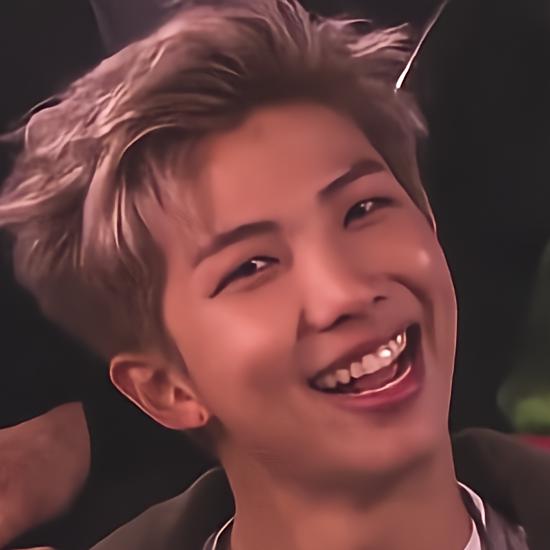}
            &
             \includegraphics[width=0.15\linewidth]{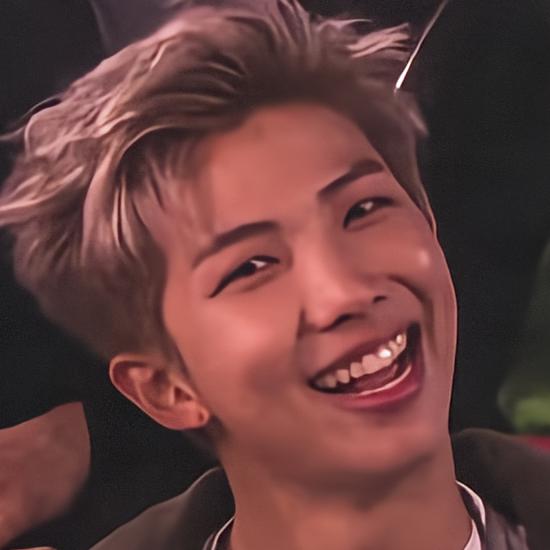}
            &
             \includegraphics[width=0.15\linewidth]{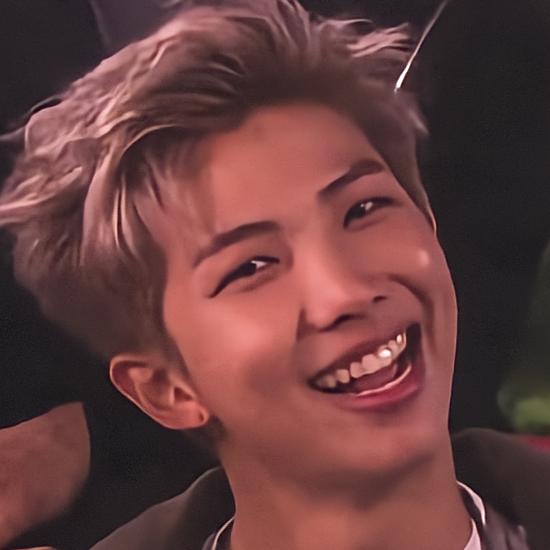}
             \vspace{-0.1cm} \\
             \begin{tabular}{c}
             \vspace{-4.2cm} \\ \footnotesize Optimal task vector  \vspace{-0.1cm} \\ \footnotesize =  \vspace{-0.1cm}\\ \footnotesize Unknown
             \end{tabular}
             &
             \begin{tabular}{l}
             \vspace{-2.8cm} \\  \footnotesize  (0.6,0.6,0.6) 
             \end{tabular}
            &
             \includegraphics[width=0.15\linewidth]{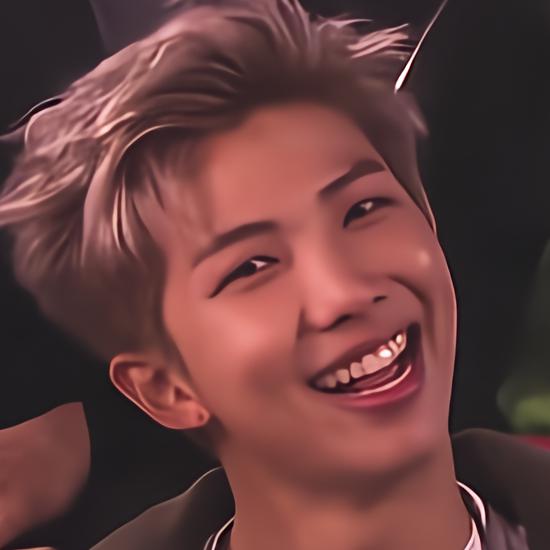}
            &
             \includegraphics[width=0.15\linewidth]{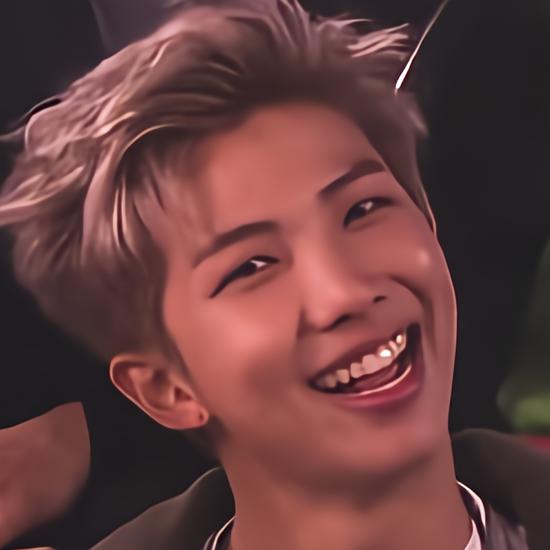}
            &
             \includegraphics[width=0.15\linewidth]{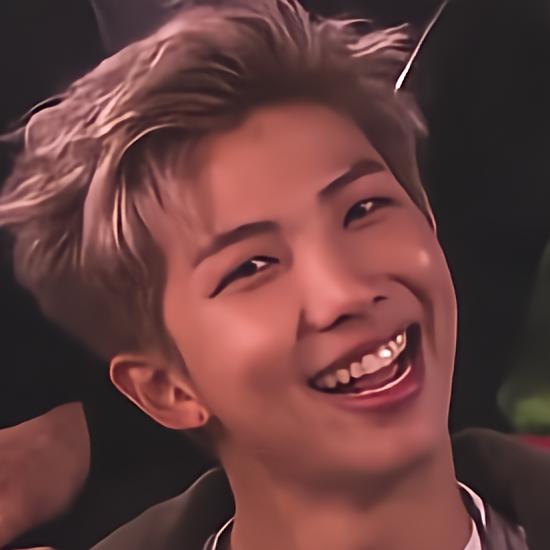}
            &
             \includegraphics[width=0.15\linewidth]{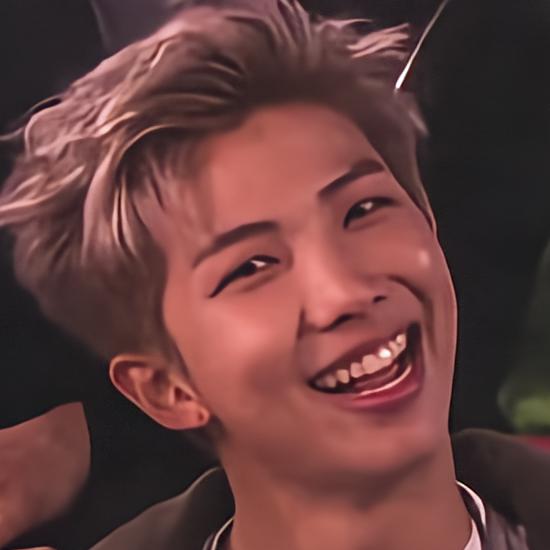}
            &
             \includegraphics[width=0.15\linewidth]{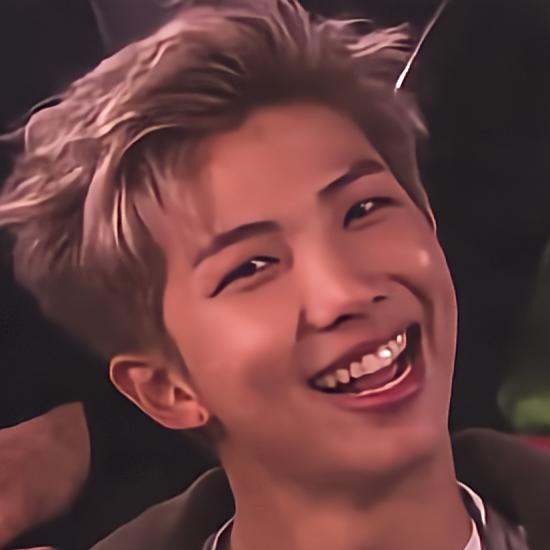}
             \vspace{-0.1cm} \\
            &
             \begin{tabular}{l}
             \vspace{-2.8cm} \\  \footnotesize  (0.8,0.8,0.8) 
             \end{tabular}
            &
             \includegraphics[width=0.15\linewidth]{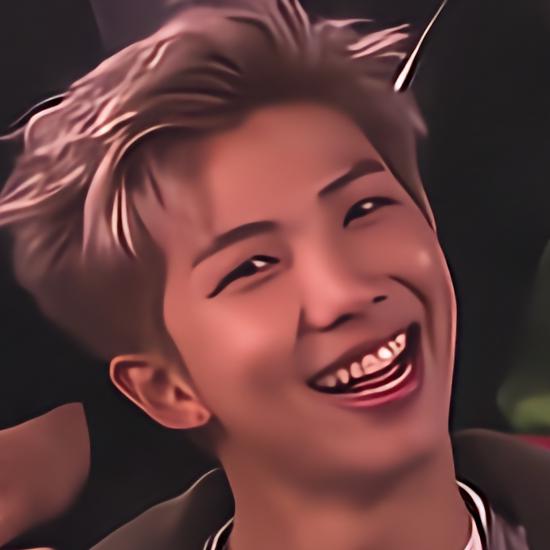}
            &
             \includegraphics[width=0.15\linewidth]{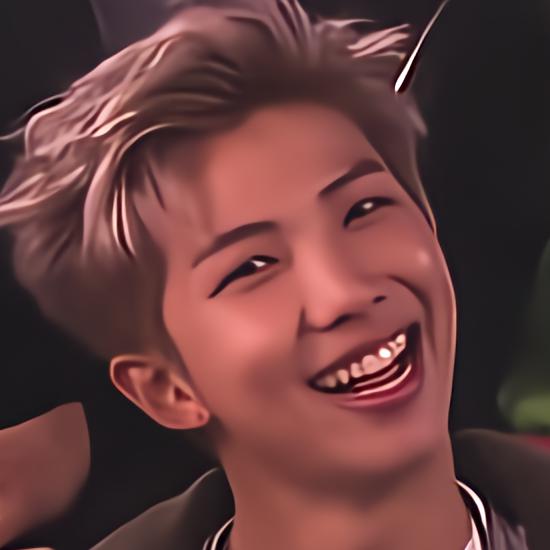}
            &
             \includegraphics[width=0.15\linewidth]{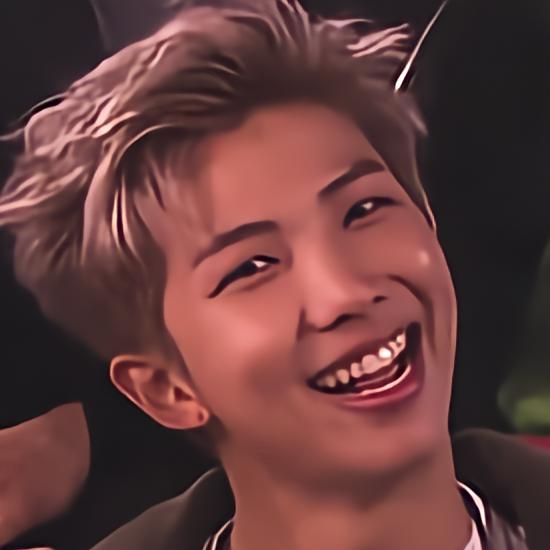}
            &
             \includegraphics[width=0.15\linewidth]{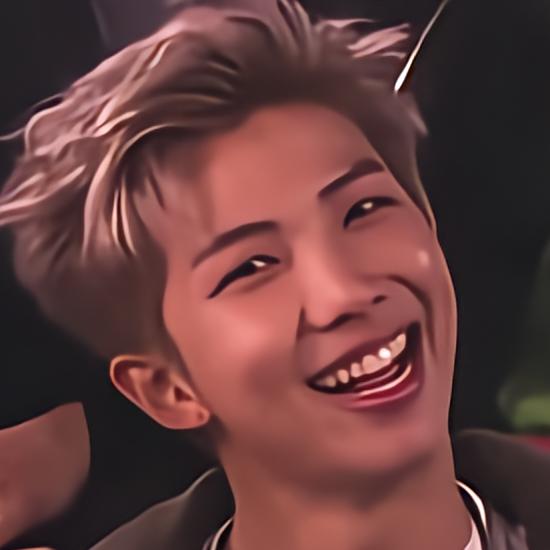}
            &
             \includegraphics[width=0.15\linewidth]{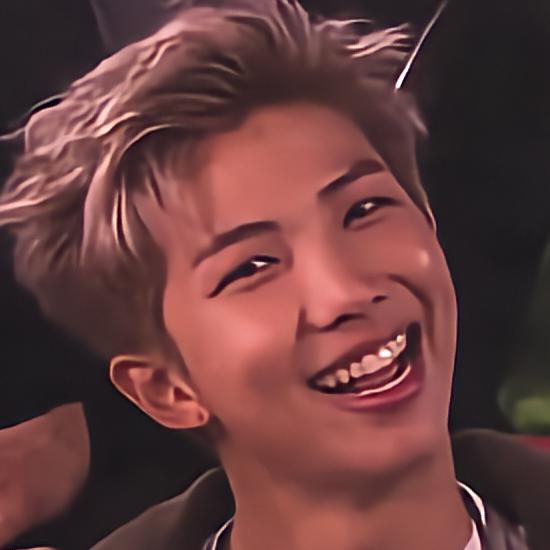}
             \vspace{-0.1cm} \\
            &
             \begin{tabular}{l}
             \vspace{-2.8cm} \\  \footnotesize  (1,1,1) 
             \end{tabular}
            &
             \includegraphics[width=0.15\linewidth]{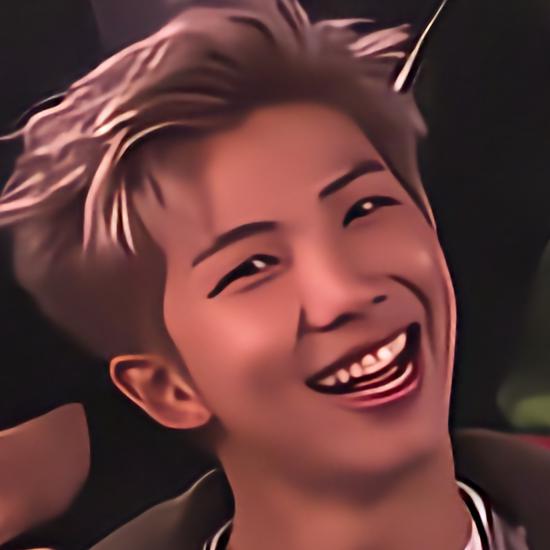}
            &
             \includegraphics[width=0.15\linewidth]{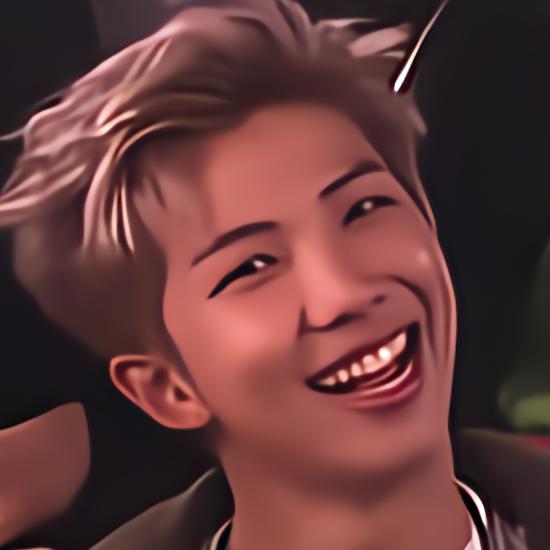}
            &
             \includegraphics[width=0.15\linewidth]{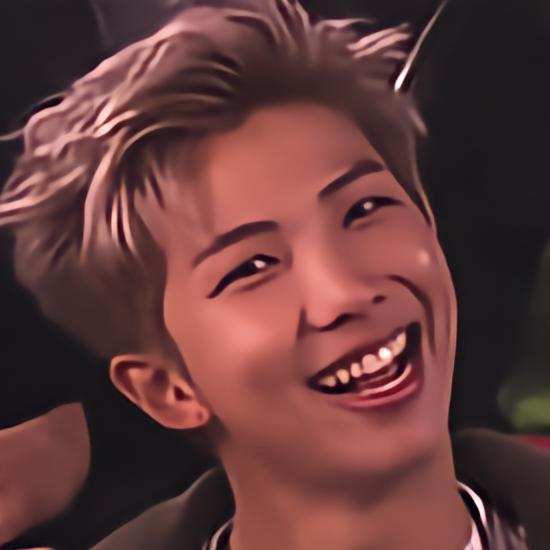}
            &
             \includegraphics[width=0.15\linewidth]{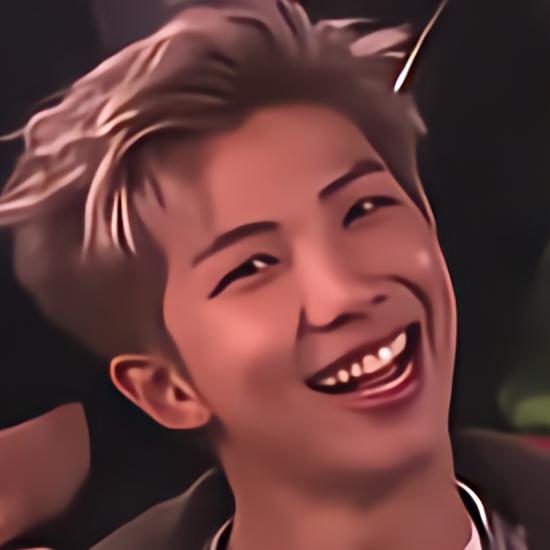}
            &
             \includegraphics[width=0.15\linewidth]{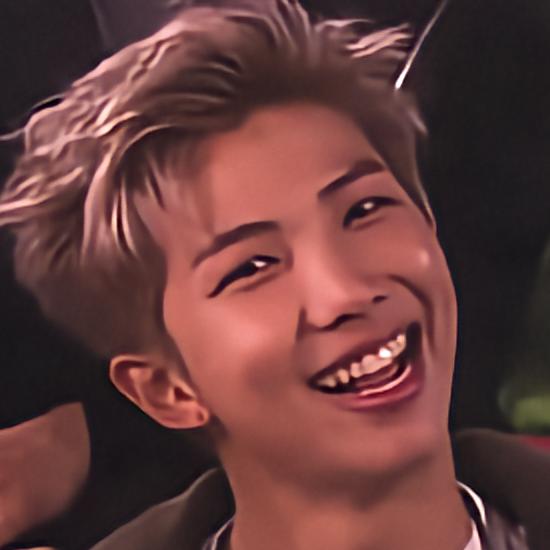}
             \vspace{-0.1cm} \\

	\toprule[1.5pt]
        \end{tabular}%\vspace{-0.3cm}
        \caption{\textbf{Deblur, denoise, and dejpeg modulation examples to the real world image on the Internet}. Our TA+TSNet-m modulates diverse imagery effects with respect to the given restoration tasks. It generates less auxiliary visual artifacts and over-smoothed textures. The values of task vector denote restoration levels of (deblur, denoise, dejpeg), respectively.}\vspace{-0.0cm}
        \label{fig:real_all}
    \end{center}
	\vspace{-0.2cm}
\end{figure*}

\begin{figure*}[p]
        \begin{center}\centering
        \setlength{\tabcolsep}{0.02cm}
        \setlength{\columnwidth}{2.81cm}
        \hspace*{-\tabcolsep}\begin{tabular}{ccccccc}
            \multicolumn{1}{c}{\footnotesize Input}    
            &
            \footnotesize Task vector
            &
            \multicolumn{1}{c}{\footnotesize CResMD~\cite{jingwen2020interactive}}
            &
            \multicolumn{1}{c}{ \footnotesize TSNet }
            &
            \multicolumn{1}{c}{\footnotesize TA+TSNet }
            &
            \multicolumn{1}{c}{\footnotesize TSNet-m }
            &
            \multicolumn{1}{c}{\footnotesize TA+TSNet-m }
            \\ \toprule[1.5pt]
            %\vspace{-0.1cm} \\
            &
             \begin{tabular}{c}
             \vspace{-2.8cm} \\  \footnotesize  (0,0,0) 
             \end{tabular}
            &
             \includegraphics[width=0.15\linewidth]{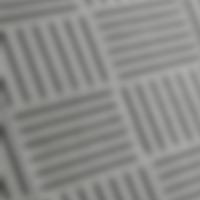}
            &
             \includegraphics[width=0.15\linewidth]{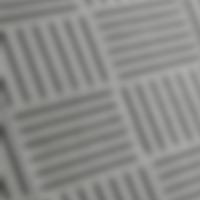}
            &
             \includegraphics[width=0.15\linewidth]{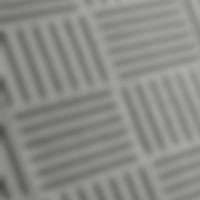}
            &
             \includegraphics[width=0.15\linewidth]{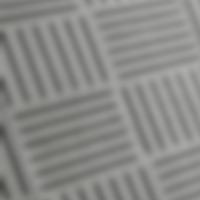}
            &
             \includegraphics[width=0.15\linewidth]{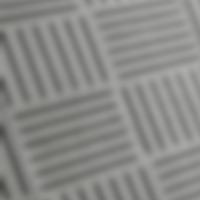}
             \vspace{-0.1cm} \\
             
            &
             \begin{tabular}{l}
             \vspace{-2.8cm} \\  \footnotesize  (0.2,0,0) 
             \end{tabular}
            &
             \includegraphics[width=0.15\linewidth]{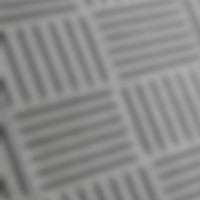}
            &
             \includegraphics[width=0.15\linewidth]{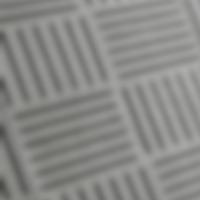}
            &
             \includegraphics[width=0.15\linewidth]{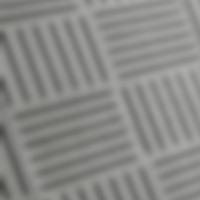}
            &
             \includegraphics[width=0.15\linewidth]{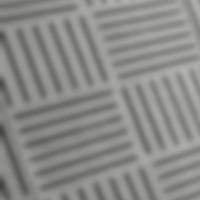}
            &
             \includegraphics[width=0.15\linewidth]{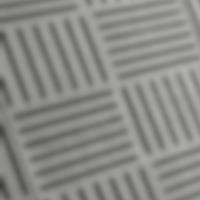}
             \vspace{-0.1cm} \\
            &
             \begin{tabular}{l}
             \vspace{-2.8cm} \\  \footnotesize  (0.4,0,0) 
             \end{tabular}
            &
             \includegraphics[width=0.15\linewidth]{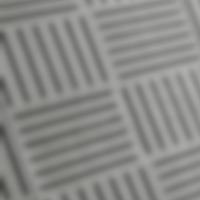}
            &
             \includegraphics[width=0.15\linewidth]{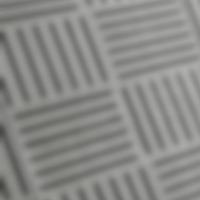}
            &
             \includegraphics[width=0.15\linewidth]{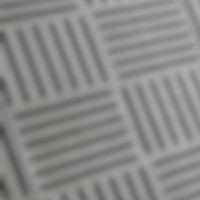}
            &
             \includegraphics[width=0.15\linewidth]{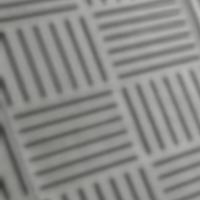}
            &
             \includegraphics[width=0.15\linewidth]{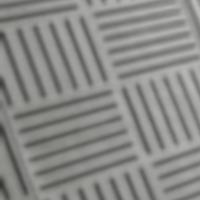}
             \vspace{-0.1cm} \\
             \begin{tabular}{c}
             \vspace{-2.8cm}\\
             \includegraphics[width=0.15\linewidth]{graphics/urban/urban.jpg}\\
             \vspace{-1cm} \\ {\color{black} Synthetic } 
             \end{tabular}
            &
             \begin{tabular}{l}
             \vspace{-2.8cm} \\  \footnotesize  (0.5,0,0) 
             \end{tabular}
            &
             \includegraphics[width=0.15\linewidth]{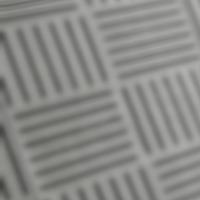}
            &
             \includegraphics[width=0.15\linewidth]{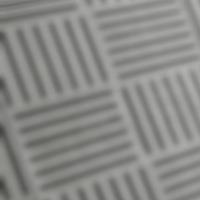}
            &
             \includegraphics[width=0.15\linewidth]{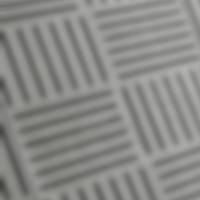}
            &
             \includegraphics[width=0.15\linewidth]{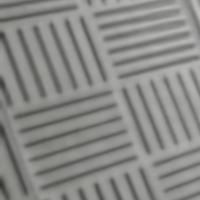}
            &
             \includegraphics[width=0.15\linewidth]{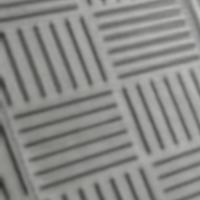}
             \vspace{-0.1cm} \\
             \begin{tabular}{c}
             \vspace{-4.2cm} \\ \footnotesize Optimal task vector  \vspace{-0.1cm} \\ \footnotesize =  \vspace{-0.1cm}\\ \footnotesize (1,0,0)
             \end{tabular}
             &
             \begin{tabular}{l}
             \vspace{-2.8cm} \\  \footnotesize  (0.6,0,0) 
             \end{tabular}
            &
             \includegraphics[width=0.15\linewidth]{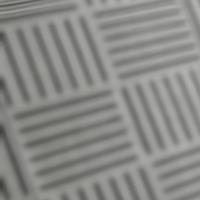}
            &
             \includegraphics[width=0.15\linewidth]{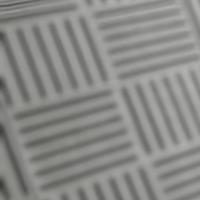}
            &
             \includegraphics[width=0.15\linewidth]{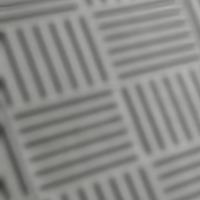}
            &
             \includegraphics[width=0.15\linewidth]{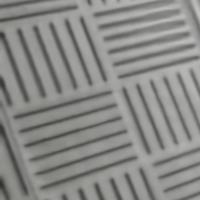}
            &
             \includegraphics[width=0.15\linewidth]{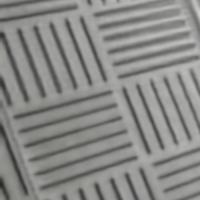}
             \vspace{-0.1cm} \\
            &
             \begin{tabular}{l}
             \vspace{-2.8cm} \\  \footnotesize  (0.8,0,0) 
             \end{tabular}
            &
             \includegraphics[width=0.15\linewidth]{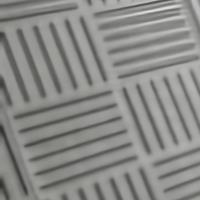}
            &
             \includegraphics[width=0.15\linewidth]{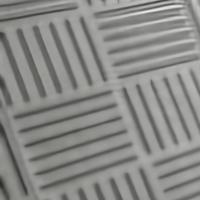}
            &
             \includegraphics[width=0.15\linewidth]{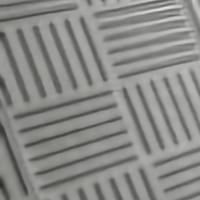}
            &
             \includegraphics[width=0.15\linewidth]{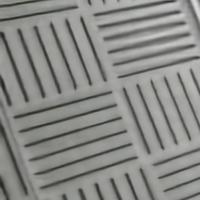}
            &
             \includegraphics[width=0.15\linewidth]{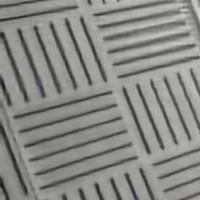}
             \vspace{-0.1cm} \\
            &
             \begin{tabular}{l}
             \vspace{-2.8cm} \\  \footnotesize  (1,0,0) 
             \end{tabular}
            &
             \includegraphics[width=0.15\linewidth]{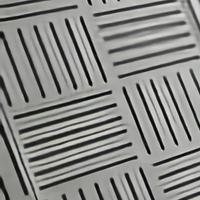}
            &
             \includegraphics[width=0.15\linewidth]{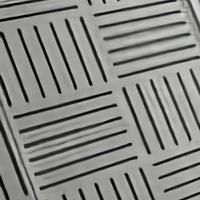}
            &
             \includegraphics[width=0.15\linewidth]{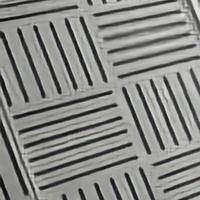}
            &
             \includegraphics[width=0.15\linewidth]{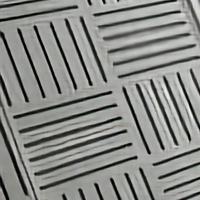}
            &
             \includegraphics[width=0.15\linewidth]{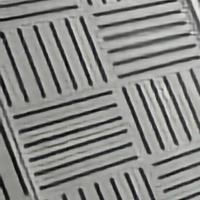}
             \vspace{-0.1cm} \\

	\toprule[1.5pt]
        \end{tabular}%\vspace{-0.3cm}
        \caption{\textbf{Deblur modulation examples to the image with blur}. Our TA+TSNet-m modulates diverse imagery effects with respect to the given restoration tasks. The values of task vector denote restoration levels of (deblur, denoise, dejpeg), respectively.}\vspace{-0.0cm}
        \label{fig:drastic_blur}
    \end{center}
	\vspace{-0.2cm}
\end{figure*}

\begin{figure*}[p]
        \begin{center}\centering
        \setlength{\tabcolsep}{0.02cm}
        \setlength{\columnwidth}{2.81cm}
        \hspace*{-\tabcolsep}\begin{tabular}{ccccccc}
            \multicolumn{1}{c}{\footnotesize Input}    
            &
            \footnotesize Task vector
            &
            \multicolumn{1}{c}{\footnotesize CResMD~\cite{jingwen2020interactive}}
            &
            \multicolumn{1}{c}{ \footnotesize TSNet }
            &
            \multicolumn{1}{c}{\footnotesize TA+TSNet }
            &
            \multicolumn{1}{c}{\footnotesize TSNet-m }
            &
            \multicolumn{1}{c}{\footnotesize TA+TSNet-m }
            \\ \toprule[1.5pt]
            %\vspace{-0.1cm} \\
            &
             \begin{tabular}{c}
             \vspace{-2.8cm} \\  \footnotesize  (0,0,0) 
             \end{tabular}
            &
             \includegraphics[width=0.15\linewidth]{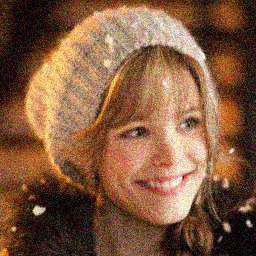}
            &
             \includegraphics[width=0.15\linewidth]{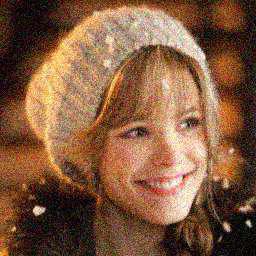}
            &
             \includegraphics[width=0.15\linewidth]{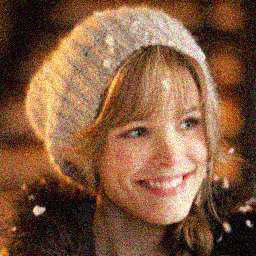}
            &
             \includegraphics[width=0.15\linewidth]{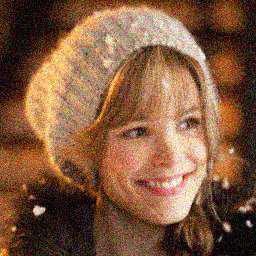}
            &
             \includegraphics[width=0.15\linewidth]{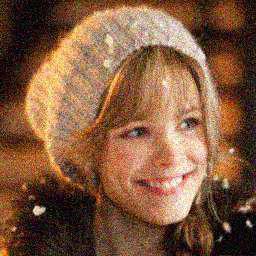}
             \vspace{-0.1cm} \\
             
            &
             \begin{tabular}{l}
             \vspace{-2.8cm} \\  \footnotesize  (0,0.2,0.2) 
             \end{tabular}
            &
             \includegraphics[width=0.15\linewidth]{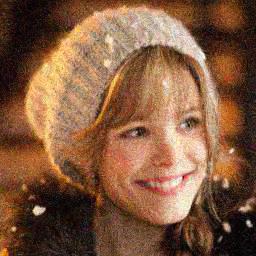}
            &
             \includegraphics[width=0.15\linewidth]{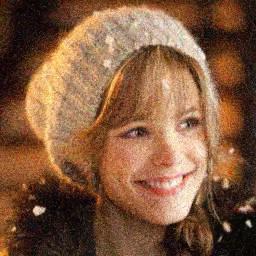}
            &
             \includegraphics[width=0.15\linewidth]{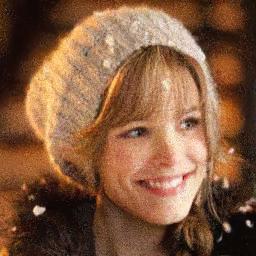}
            &
             \includegraphics[width=0.15\linewidth]{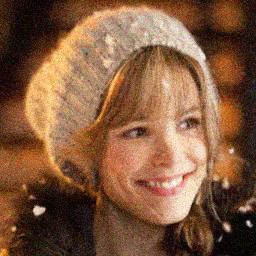}
            &
             \includegraphics[width=0.15\linewidth]{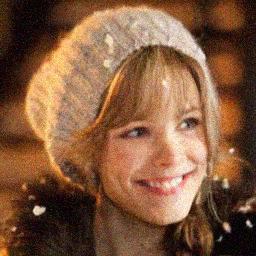}
             \vspace{-0.1cm} \\
            &
             \begin{tabular}{l}
             \vspace{-2.8cm} \\  \footnotesize  (0,0.4,0.4) 
             \end{tabular}
            &
             \includegraphics[width=0.15\linewidth]{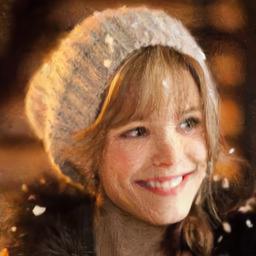}
            &
             \includegraphics[width=0.15\linewidth]{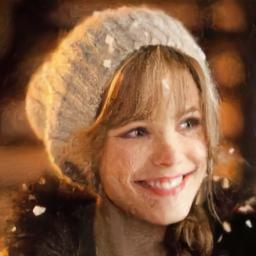}
            &
             \includegraphics[width=0.15\linewidth]{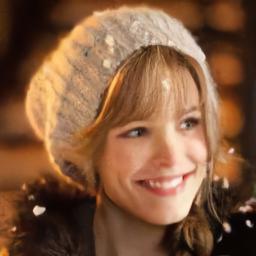}
            &
             \includegraphics[width=0.15\linewidth]{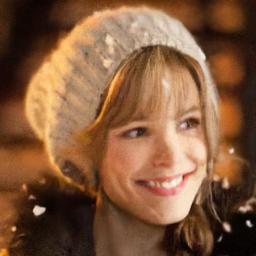}
            &
             \includegraphics[width=0.15\linewidth]{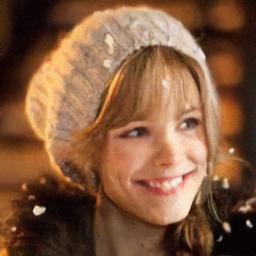}
             \vspace{-0.1cm} \\
             \begin{tabular}{c}
             \vspace{-2.8cm}\\
             \includegraphics[width=0.15\linewidth]{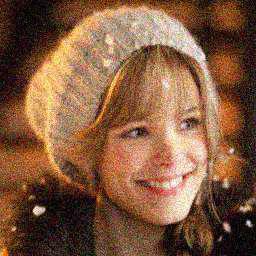}\\
             \vspace{-1cm} \\ {\color{black} Synthetic } 
             \end{tabular}
            &
             \begin{tabular}{l}
             \vspace{-2.8cm} \\  \footnotesize  (0,0.5,0.5) 
             \end{tabular}
            &
             \includegraphics[width=0.15\linewidth]{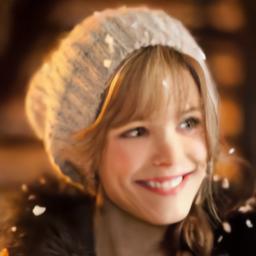}
            &
             \includegraphics[width=0.15\linewidth]{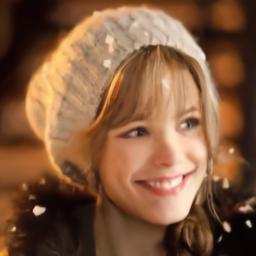}
            &
             \includegraphics[width=0.15\linewidth]{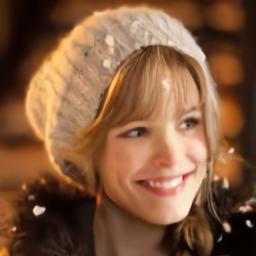}
            &
             \includegraphics[width=0.15\linewidth]{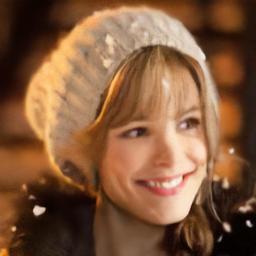}
            &
             \includegraphics[width=0.15\linewidth]{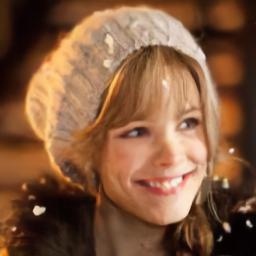}
             \vspace{-0.1cm} \\
             \begin{tabular}{c}
             \vspace{-4.2cm} \\ \footnotesize Optimal task vector  \vspace{-0.1cm} \\ \footnotesize =  \vspace{-0.1cm}\\ \footnotesize (0,0.5,0.5)
             \end{tabular}
             &
             \begin{tabular}{l}
             \vspace{-2.8cm} \\  \footnotesize  (0,0.6,0.6) 
             \end{tabular}
            &
             \includegraphics[width=0.15\linewidth]{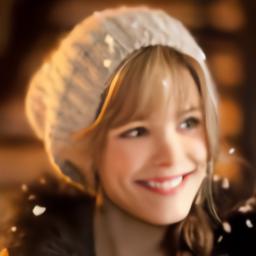}
            &
             \includegraphics[width=0.15\linewidth]{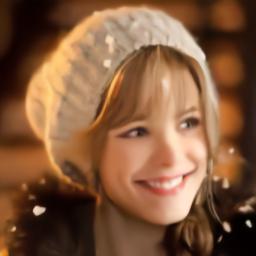}
            &
             \includegraphics[width=0.15\linewidth]{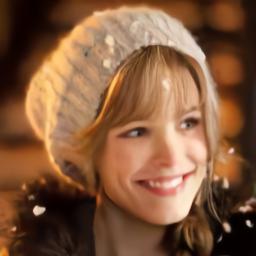}
            &
             \includegraphics[width=0.15\linewidth]{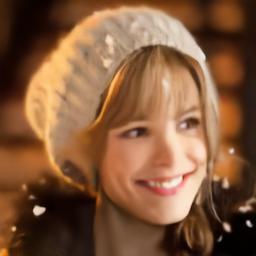}
            &
             \includegraphics[width=0.15\linewidth]{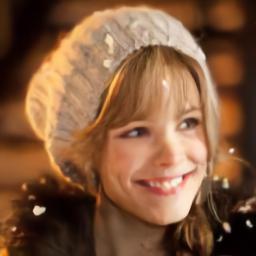}
             \vspace{-0.1cm} \\
            &
             \begin{tabular}{l}
             \vspace{-2.8cm} \\  \footnotesize  (0,0.8,0.8) 
             \end{tabular}
            &
             \includegraphics[width=0.15\linewidth]{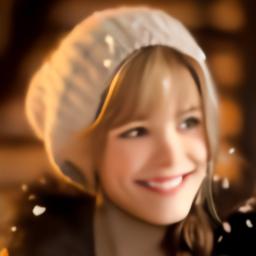}
            &
             \includegraphics[width=0.15\linewidth]{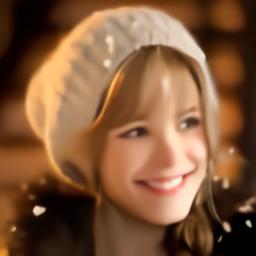}
            &
             \includegraphics[width=0.15\linewidth]{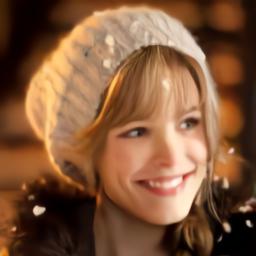}
            &
             \includegraphics[width=0.15\linewidth]{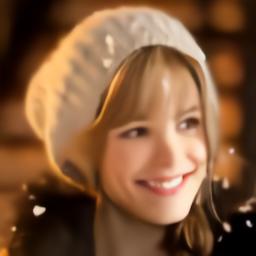}
            &
             \includegraphics[width=0.15\linewidth]{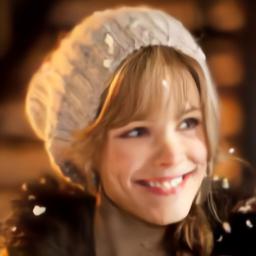}
             \vspace{-0.1cm} \\
            &
             \begin{tabular}{l}
             \vspace{-2.8cm} \\  \footnotesize  (0,1,1) 
             \end{tabular}
            &
             \includegraphics[width=0.15\linewidth]{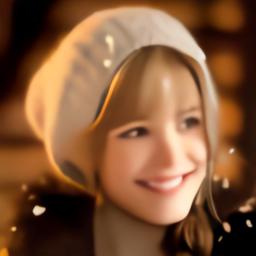}
            &
             \includegraphics[width=0.15\linewidth]{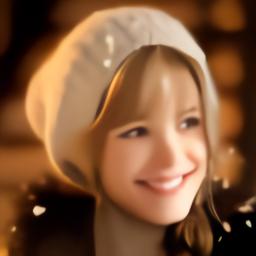}
            &
             \includegraphics[width=0.15\linewidth]{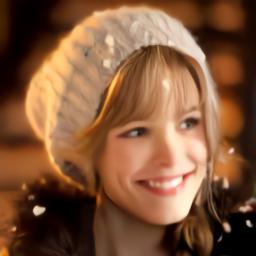}
            &
             \includegraphics[width=0.15\linewidth]{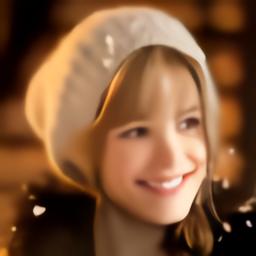}
            &
             \includegraphics[width=0.15\linewidth]{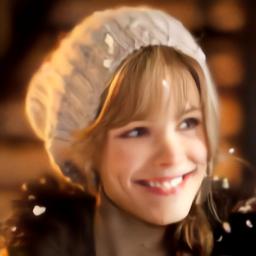}
             \vspace{-0.1cm} \\

	\toprule[1.5pt]
        \end{tabular}%\vspace{-0.3cm}
        \caption{\textbf{Denoise and dejpeg modulation examples to the image with noise and jpeg compression}. Our TA+TSNet-m modulates diverse imagery effects with respect to the given restoration tasks. It generates less over-smoothed textures. The values of task vector denote restoration levels of (deblur, denoise, dejpeg), respectively.}\vspace{-0.0cm}
        \label{fig:drastic_denoise_dejpeg}
    \end{center}
	\vspace{-0.2cm}
\end{figure*}

This section analyzes the image modulation performances in Figure~\ref{fig:downgrade} with examples.
Recall, CResMD incurs two problems in controlling imagery effects which are artifact in generated images and uneven modulation across task vectors.
Figure~\ref{fig:synthetic_deblur}~and~\ref{fig:synthetic_deblur_dejpeg} show that CResMD produces output images with undesired and visually unpleasing artifact when the degradation types are mismatched to the types of restoration tasks.
In the matched case, CResMD can successfully modulate the images, as demonstrated in Figure~\ref{fig:synthetic_all}.
Figure~\ref{fig:real_deblur},~\ref{fig:real_denoise},~and~\ref{fig:real_all} presents modulation scenarios with real-word image with unknown degradation, in which modulations with various task vectors are required to find the visually pleasing images.
Figures demonstrate that CResMD generates severely destructive artifacts (especially in Figure~\ref{fig:real_deblur}) in process whereas TA+TSNet-m modulates the image with various task vectors without disrupting the image.
Figure~~\ref{fig:drastic_blur} and \ref{fig:drastic_denoise_dejpeg} exemplify the other problem with CResMD of the uneven modulation across task levels.
In both figures, CResMD produces images with negligible changes at lower task vectors.
On the other hand, images modulated by CResMD at task vectors near the optimal task vectors exhibit drastic changes.
In contrast to CResMD, TA+TSNet-m generates various images as task vector changes, demonstrating more even modulation across task vectors.

\clearpage
% !TEX root = supple.tex

%%%%%%%%%%%%%%%%%%%%%%%%%%%%%%%%%%%%%%%%%%%%%%%%%% BTS

\clearpage

\subsection{Implementation details}
\label{sec:implementation_details}

The super network CResMD consists of 32 enhanced residual blocks~\cite{bee2017enhanced} which have a ReLU activation layer between two convolution layers with 64 filters of the kernel size 3$\times$3.
The first convolution layer with a stride of 2 downscales the input images and the last upsampling module consists of PixelShuffle layer, two convolution layers, and a ReLU activation layer.
Global skip connection adds the input image to the output of the upscaling module.
A task vector scales the residual feature map in the location of 32 local connections and 1 global connection by a 1$\times$1 convolution layer with channel-wise multiplication.

Our channel selection module determines the input and output channels of all convolution layers in CResMD except the input channels of the first convolution layer and the PixelShuffle layer and the output channels of the last convolution layer.
The architecture controller consists of 3 fully-connected layers with 2 ReLU activation layers and determines the channels in task-specific architectures with respect to the task vector.
Task-agnostic architecture removes the residual scaling module in CResMD blocks and its channels are determined by auxiliary parameters.
TSNet sets the hyperparameters $\alpha$, $\gamma$, $M$, $\lambda_1$, and $\lambda_2$ as $0.9$, $0.9$, $1$, $5\times e^{-11}$, and 0, respectively.
TA+TSNet changes the value of $M$ to the size of mini-batch and the value of $\lambda_2$ to $1 \times e^{-2}$.
All models adopt the mini-batch size of 64 with 64$\times$64 sub-images, the initial learning rate of $1\times10^{-4}$, and Adam optimizer.
Each model is trained in $1\times 10^6$ iterations with a learning rate decay at the half of training.

We adopt DIV2K~\cite{agustsson2017ntire} and CBSD68~\cite{MartinFTM01} as the train and test datasets.
DIV2K consists of clean 800 2K training images and 100 2K validation images while CBSD68 consists of clean 68 HVGA test images.
Following the degradation setting in CResMD~\cite{jingwen2020interactive}, we conduct 3 types of degradation ($D=3$), Gaussian blur, Gaussian noise, and JPEG compression.
Each degradation is applied to clean images sequentially.
The Gaussian blur kernel width denotes $r \in [0,4]$ and its kernel size is 21 $\times$ 21.
The Gaussian noise covariance denotes $\sigma \in [0,50]$ and the JPEG compression quality factor denotes $q \in [100,10]$.
The training dataset consists of the degradation levels with stride of 0.1, 1, and 2 for $r$, $\sigma$, and $q$, respectively, with an additional level of non-compressed images to the compressed image quality factor.
To ease the burden of evaluation, an image in the validation and test datasets has the combination of $r \in \{0,2,4\}$, $\sigma \in \{0,25,50\}$, and $q \in \{None,60,10\}$ except for the starting point $(r,\sigma,q) = (0,0,None)$ since the PSNR score reaches an infinity.

\subsection{Evaluation metric}
\label{sec:evaluation_metric}
\paragraph{Network computation cost.}
We measure the computation costs of neural networks with FLOPs and latency.
FLOPs is a classical device-agnostic metric and exponentially increases by image resolution.
Since latency is device-dependant, we measure latency on CPU with single-core (Latency-CPU-S), CPU with multi-core (Latency-CPU-M), and GPU (Latency-GPU).
We use Intel i7-5960X CPU which has 16 cores and GeForce RTX 2080 Ti GPU.
The reported computation costs, referred to as $\mathcal{R}_{total}(f,\boldsymbol{x})$, are average scores to generate $M$ imagery effects.
We set $M=27$ (unless otherwise mentioned) of which tasks include all degradation of test dataset with the starting point.

\vspace{-0.3cm}
\paragraph{Image quality.}
We adopts widely used 4 image quality measures, PSNR, SSIM, LPIPS~\cite{zhang2018perceptual}, and NIQE~\cite{mittal2013niqe}.
PSNR, SSIM, and LPIPS measure the image quality by comparing the restored images to the original clean images while NIQE evaluates the restored image quality without referring to an original image.
PSNR reports the mean squared error in log scale.
SSIM measures the structured similarity while LPIPS scores the perceptual similarity using neural networks.
NIQE evaluates the naturalness of images by predetermined images of natural scenes.
PSNR and SSIM score the better the higher while LPIPS and NIQE score the better the lower.
However, measuring image quality during modulating imagery effects has not been studied thoroughly.
Thus, we visualize extensive qualitative results and video figures in both the main manuscript and the supplementary document.
The video figures are best viewed using Adobe Reader.

\end{document}